\documentclass[journal]{IEEEtran}
\usepackage{cite}
\ifCLASSINFOpdf
\else
\fi
\usepackage{graphics}                        % for pdf, bitmapped graphics files
\usepackage{graphicx}
\graphicspath{ {./images/} }
\usepackage{booktabs}
\usepackage[flushleft]{threeparttable}
\usepackage{subcaption}
\usepackage{epsfig}                          % for postscript graphics files
\usepackage{mathptmx}                        % assumes new font selection scheme installed
\usepackage{times}                           % assumes new font selection scheme installed
\usepackage{algorithm,algpseudocode}
\usepackage{amsmath}                         % assumes amsmath package installed
\usepackage{amssymb}                         % assumes amsmath package installed
\usepackage{multicol,multirow,blindtext}
\usepackage{booktabs,amsfonts,dcolumn}
\newcolumntype{d}[1]{D..{#1}}

\usepackage{balance}
\usepackage{array}
\usepackage{hyperref}
\usepackage{esvect}
\usepackage{cuted}
\usepackage[dvipsnames]{xcolor}
\usepackage{tabularray}
\usepackage{fixmath}
\usepackage{comment}

\usepackage{xcolor,colortbl}

\definecolor{Gray}{gray}{0.85}
\definecolor{LightCyan}{rgb}{0.88,1,1}
\newcolumntype{a}{>{\columncolor{Gray}}c}

\makeatletter
\newcommand\notsotinyone{\@setfontsize\notsotinyone{6.4}{7}}
\newcommand\notsotinytwo{\@setfontsize\notsotinytwo{5.55}{7}}
\newcommand\notsotinythree{\@setfontsize\notsotinythree{5.73}{7}}
\newcommand\notsotinyfour{\@setfontsize\notsotinyfour{8.87}{10}}
\makeatother

\begin{document}
%
% paper title
% Titles are generally capitalized except for words such as a, an, and, as,
% at, but, by, for, in, nor, of, on, or, the, to and up, which are usually
% not capitalized unless they are the first or last word of the title.
% Linebreaks \\ can be used within to get better formatting as desired.
% Do not put math or special symbols in the title.
\title{Choosing the Correct Generalized Inverse for the Numerical Solution of the Inverse Kinematics of Incommensurate Robotic Manipulators}
%
%
% author names and IEEE memberships
% note positions of commas and nonbreaking spaces ( ~ ) LaTeX will not break
% a structure at a ~ so this keeps an author's name from being broken across
% two lines.
% use \thanks{} to gain access to the first footnote area
% a separate \thanks must be used for each paragraph as LaTeX2e's \thanks
% was not built to handle multiple paragraphs
%

\author{Jacket~Demby's,~\IEEEmembership{Student Member,~IEEE,}
        Jeffrey~Uhlmann,~\IEEEmembership{Member,~IEEE,}
        and~Guilherme~N.~DeSouza,~\IEEEmembership{Member,~IEEE}% <-this % stops a space
\thanks{All authors are with the Department of Electrical Engineering and Computer Science (EECS), University of Missouri-Columbia, Columbia, Missouri, 65201.}%
\thanks{Jacket Demby's and Guilherme N. DeSouza are with the Vision-Guided and Intelligent Robotics (ViGIR) Laboratory. (email: udembys@mail.missouri.edu; uhlmannj@missouri.edu; desouzag@missouri.edu)}%
%\thanks{Manuscript received April 19, 2005; revised August 26, 2015.}
}

\maketitle
%\noindent\textcolor{red}{[TRANSACTION DRAFT FOR REVIEW]}

% As a general rule, do not put math, special symbols or citations
% in the abstract or keywords.
\begin{abstract}
Numerical methods for Inverse Kinematics (IK) employ iterative, linear approximations of the IK until the end-effector is brought from its initial pose to the desired final pose. These methods require the computation of the Jacobian of the Forward Kinematics (FK) and its inverse in the linear approximation of the IK. Despite all the successful implementations reported in the literature, Jacobian-based IK methods  can still fail to preserve certain useful properties if an improper matrix inverse, e.g. Moore-Penrose (MP), is employed for incommensurate robotic systems. In this paper, we propose a systematic, robust and accurate numerical solution for the IK problem using the Mixed (MX) Generalized Inverse (GI) applied to any type of Jacobians (e.g., analytical, numerical or geometric) derived for any commensurate and incommensurate robot. This approach is robust to whether the system is under-determined (less than 6 DoF) or over-determined (more than 6 DoF). We investigate six robotics manipulators with various Degrees of Freedom (DoF) to demonstrate that commonly used GI's fail to guarantee the same system behaviors when the units are varied for incommensurate robotics manipulators. In addition, we evaluate the proposed methodology as a global IK solver and compare against well-known IK methods for redundant manipulators. Based on the experimental results, we conclude that the right choice of GI is crucial in preserving certain properties of the system (i.e. unit-consistency).
\end{abstract}

% Note that keywords are not normally used for peerreview papers.

% For peer review papers, you can put extra information on the cover
% page as needed:
% \ifCLASSOPTIONpeerreview
% \begin{center} \bfseries EDICS Category: 3-BBND \end{center}
% \fi
%
% For peerreview papers, this IEEEtran command inserts a page break and
% creates the second title. It will be ignored for other modes.
\IEEEpeerreviewmaketitle

\section{Introduction}
% The very first letter is a 2 line initial drop letter followed
% by the rest of the first word in caps.
% 
% form to use if the first word consists of a single letter:
% \IEEEPARstart{A}{demo} file is ....
% 
% form to use if you need the single drop letter followed by
% normal text (unknown if ever used by the IEEE):
% \IEEEPARstart{A}{}demo file is ....
% 
% Some journals put the first two words in caps:
% \IEEEPARstart{T}{his demo} file is ....
% 
% Here we have the typical use of a "T" for an initial drop letter
% and "HIS" in caps to complete the first word.
\IEEEPARstart{I}{ncommensurate} robotic manipulators refer to robotic systems having a combination of prismatic (linear) and revolute (rotational) joints. Such manipulators combine variables expressed in different units: i.e. pose vectors $\vec{D}$ with a combination of units of distance (meters, centimeters, etc.) and orientations (radians and degrees), with joint vector $\vec{Q} = [Q_1, Q_2, .., Q_n]$ where $Q_i = d_i$ is a prismatic joint, $Q_i = \theta_i$ is a revolute joint, and $n$ the number of DoF \cite{schwartz2002noncommensurate, schwartz2003non}. The joint $\vec{Q}$ and pose $\vec{D}$ vectors are linearly related by equation $\frac{\partial D}{\partial t} = J.\frac{\partial Q}{\partial t} \iff \frac{\partial Q}{\partial t} = J^{\widetilde{-1}}.\frac{\partial D}{\partial t}$, where $J$ is the Jacobian matrix (e.g., analytical \cite{sciavicco2001modelling}, geometric \cite{sciavicco2001modelling}, numerical \cite{farzan2013dh} or elementary transform sequence \cite{haviland2020systematic, corke2021not}), and $J^{\widetilde{-1}}$ any Generalized Inverse (GI) that can be employed to find the inverse of this matrix.

In the context of articulated robotic manipulators, the FK is a highly non-linear mapping from the joint space to the pose space of the end-effector. While in most useful cases these functions are neither injective (one-to-one) nor surjective (onto), depending on the robot configuration – i.e. the sequence of prismatic and revolute joints, and the number of Degrees of Freedom (DoF) – the associated IK problem may be practically or even theoretically impossible to be solved analytically  \cite{aristidou2009inverse, colome2014closed, colome2012redundant, craig2009introduction}. Therefore, in the past decades, several approximate methods have been developed for many instances of robots. These approximate methods can be divided into two distinct categories: data-driven and numerical approaches. In this paper, we will focus on numerical approaches \cite{aristidou2009inverse, aristidou2018inverse, colome2012redundant, colome2014closed}, but the reader should report to \cite{demby2020use, dembys2019solving} for other considerations and implementations of data-driven methods.
Despite many successful implementations, numerical approaches for IK may fail if an improper matrix inverse is employed. In fact, numerical IK methods often resort to inverting a Jacobian matrix at each iteration of the process, and whenever this Jacobian becomes singular – either under-determined (less than 6 DoF) or over-determined (more than 6 DoF) – its inverse cannot be uniquely defined. So, typically a left or right pseudo-inverse, or more generically, the Moore-Penrose (MP) GI  \cite{penrose1955generalized} is often employed without regard to whether the same GI can guarantee the required properties of the system. In particular, the MP is by far the most widely employed GI for robotics-related IK applications even though it often fails to preserve critical properties, e.g., in the case of incommensurate manipulators that require consistency with respect to the change of units of certain joint variables  \cite{uhlmann2018generalized, zhang2019applying, zhang2020examining}. In other words, an IK solution may exhibit unrecognized sensitivity to the choice of units when clearly the behavior of the system should not be affected by such choices \cite{schwartz2002noncommensurate, schwartz2003non, isbell1999mars, lloyd1999metric}. The problem can be understood by recognizing that the MP inverse provides rotational consistency, but not unit consistency \cite{uhlmann2018generalized, zhang2019applying, zhang2020examining, uhlmannRGA}. Therefore, an alternative GI, namely the Unit-Consistent (UC), represents the appropriate choice of GI in such cases \cite{uhlmann2018generalized, zhang2020examining}. On the other hand, while the UC inverse is consistent with respect to arbitrary changes of units, it does not provide consistency with respect to rotation of the reference frame or any frame prior to the frame where the unit change occurred -- according to the D-H representation of the robot \cite{hartenberg1955kinematic, niku2010introduction, craig2009introduction}. Thus, most IK problems will involve components with differing consistency requirements. When that is the case,  one can use a Mixed (MX) GI \cite{uhlmann2018generalized, zhang2019applying, zhang2020examining} to selectively provide consistency with respect to arbitrary changes of both units and rotations of the coordinate system or prior frames.

The main contributions of this paper are: (1) bringing to light the need to investigate the robustness of commonly used GI's \cite{colome2014closed, uhlmann2018generalized} towards unit changes when computing inverse matrices in many robotic control problems, in particular the Jacobian for numerical IK methods; and (2) a systematic way for selecting the correct GI by inspection of the  Denavit-Hartenberg (D-H) representation of the robot, while handling static (under/over-determined) and dynamic singularities of the Jacobian for arbitrary attenuation parameters $\alpha$ (aka gain) controlling smoothness and rate of convergence, without slowing down the numerical IK solver \cite{colome2012redundant, colome2014closed} and guaranteeing unit-consistency requirements for incommensurate robotic manipulators.

\vspace{-5mm}
\section{Background and Related Work}\label{sec:background}
\vspace{-1mm}
As mentioned earlier, the reader should report to \cite{demby2020use, dembys2019solving} for more details on: data-driven approaches using: 1) Artificial Neural Networks (ANN), both in task-driven \cite{choi1992inverse, jha2014neural, csiszar2017solving, srisuk2017inverse, polyzos2019solving} and task-independent cases \cite{dembys2019solving, daya2010applying, zhou2018inverse, el2017comparative}; 2) Soft-computing methods – MLP, ANFIS, and GA  \cite{el2017comparative}; 3) Quantum behaved Particle Swarm Optimization (QPSO)  \cite{dereli2020meta}; etc. In this work, we focus on well accepted numerical approaches for accurately approximating the IK for task-independent workspaces \cite{aristidou2009inverse, aristidou2018inverse}. Unfortunately, these approaches have been shown to fail in preserving unit-consistencies in the case of incommensurate manipulators \cite{zhang2019applying, zhang2020examining, uhlmann2018generalized}. 
Robotic systems with incommensurate units are very common and refer to systems having different types of units \cite{schwartz2002noncommensurate, schwartz2003non}. In such a system, a pose vector may combine positions with units of distance, and orientations with units of angle, while a joint vector may also combine revolute joints with units of angle and prismatic joints with units of distance. For such systems to be considered stable, an arbitrary change in a specific unit should not affect the behavior of the whole system \cite{isbell1999mars, lloyd1999metric, schwartz2002noncommensurate, schwartz2003non}. Numerical IK approaches have frequently defined the MP inverse as the inverse for singular Jacobian matrices. However, the MP inverse does not always guarantee the stability of the robotic systems of interest as it may produce inconsistent or erroneous results and lead researchers astray in the presence of incommensurate systems \cite{klein1983review, schwartz1993weighted, schwartz1995algebraic, uhlmann2018generalized, zhang2019applying, zhang2020examining}. In fact, Schwartz et al. \cite{schwartz1993weighted, schwartz1999algebraic, schwartz2002noncommensurate, schwartz2003non, mansard2009unified} conducted several investigations on the effects of the units involved in the control of incommensurate robotic systems for which control algorithms use eigenvectors, eigenvalues or singular values of the Jacobian matrix. Those investigations concluded that the use of Singular Value Decomposition (SVD) in Jacobian based kinematics methods yields erroneous or arbitrary solutions. One of the possibilities found in the literature, to circumvent the inconsistencies of incommensurate systems is the use of user-defined units-adjusting weights \cite{doty1993theory, doty1995robot}. However, the choice of units-adjusting weights just adds another level of arbitrariness between systems and applications, as mentioned in \cite{schwartz2002noncommensurate}. In that sense, Uhlmann \cite{uhlmann2018generalized} developed two generalized inverses UC and MX inverses applicable to incommensurate robotic systems, while Zhang and Uhlmann \cite{zhang2019applying} used an incommensurate 3DoF manipulator to show that the MP inverse was affecting the end-effector trajectories and making the system unstable when the units were being varied. They used the UC inverse and succeeded in achieving a more reliable control of the manipulator end-effector. Similarly, Zhang and Uhlmann \cite{zhang2020examining} examined the UC and MX inverses with an incommensurate robotic system made of a rover and a robotic arm. They showed that the MP inverse failed to preserve consistencies with respect to changes of units while the UC inverse failed to preserve consistencies with respect to changes in rotation of the coordinate frame. Interestingly, a combination of these two GI's in the MX inverse based on the block matrix inverse definition was able to provide a reliable behavior for the system with respect to rotations and units consistencies. 

\vspace{-4mm}
\subsection{Analytical, geometric and numerical Jacobians} \label{sec:jacobians}
The Jacobian matrix describes the relationship between the joint velocities and the corresponding end-effector velocities. When, the end-effector can be expressed with a minimal representation in the operational space of the manipulator (e.g., manipulators with a small number of DoF), this matrix is obtained through differentiation of the Forward Kinematics (FK) function with respect to the joint variables. When that is the case, this matrix is termed as the \textit{Analytical Jacobian} ($J_{A}$) \cite{sciavicco2001modelling}. Let $\vec{D}$ and $\vec{Q}$, be respectively the pose and joint vectors, $J_{A}$ can be expressed by: 
\begin{equation}
    J_{A} = \frac{\partial \vec{D}}{\partial \vec{Q}}
\end{equation}

\noindent Through, the DH methodology, one can easisly find $\vec{D}$ from the last column of the total transformation matrix. As $n$ the number of DoF increases, $J_{A}$ becomes difficult to derive and is often replaced by the \textit{Geometric Jacobian} ($J_{G}$) \cite{sciavicco2001modelling} which is easier to compute. Let $J_{P_i}$ and $J_{O_i}$ where $i=1,...,n$, be respectively the position and orientation (3x1) vector contributions in the jacobian matrix. $J_{G}$, can be defined by:
\begin{equation}
    J_{G} = \begin{bmatrix}
                \vec{J}_{P_1} & \cdots & \vec{J}_{P_n}\\
                \vec{J}_{O_1} & \cdots & \vec{J}_{O_n}
            \end{bmatrix}
\end{equation}

\begin{equation}
        \vec{J}_{P_i} =
    \begin{cases}
      \vec{z}_{i-1} & \text{for a prismatic joint}\\
      \vec{z}_{i-1}*(\vec{p}-\vec{p}_{i-1}) & \text{for a revolute joints}\\
    \end{cases}       
\end{equation}

\begin{equation}
        \vec{J}_{O_i} =
    \begin{cases}
      \vec{0} & \text{for a prismatic joint}\\
      \vec{z}_{i-1} & \text{for a revolute joints}\\
    \end{cases}       
\end{equation}

\noindent where $\mathbold{z}_{i-1}$ is given by the third column of the rotation matrix $R^{0}_{i-1}$, $\mathbold{p}$ is given by the first three elements of the fourth column of the transformation matrix $T^{0}_{n}$, $\mathbold{p_{i-1}}$ is given by the first three elements of the fourth column of the transformation matrix $T^{0}_{i-1}$. Furthermore, as explained, in \cite{farzan2013dh, farzan2014parallel, aristidou2018inverse}, the \textit{Numerical Jacobian} ($J_{N}$) is also employed in IK solvers. By causing small arbitrarily and individual perturbations to each joint variable, each column $J_{N_{c}}$ of $J_{N}$ can be retrieved by:

\begin{equation}
    J_{N_{c}} = \frac{\partial \mathbold{D}}{\partial \mathbold{Q_{c}}} = \mathbold{D}_{t} - f(\mathbold{Q}_{t} + \begin{bmatrix}
                \vdots\\
                0\\
                0.01\\
                0\\
                \vdots
            \end{bmatrix})
\end{equation}

\noindent where the subscript $c$ indicates the selected column of $J_{N}$ and $t$ the IK update step.
As mentioned in \cite{sciavicco2001modelling}, these three Jacobians may be different and yield completely different results. In this work, we investigated all the above-mentioned Jacobians to make sure the change of units is inherent or not to all or some of them.

\vspace{-3mm}
\subsection{Generalized Inverses} \label{sec:generalized-inverses}
\subsubsection{Moore-Penrose (MP) Inverse}\label{sec:MP-inverse} MP is a uniquely determined GI that can be defined based on the Singular Value Decomposition (SVD) \cite{penrose1955generalized, ben1963contributions, golub1971singular, albert1972regression} of the matrix to invert. Considering an arbitrary matrix $A$ whose SVD is given by equation $A = USV^{*}$, where $U$ and $V$ are unit/orthonormal matrices containing the singular vectors, $S$ is a diagonal matrix containing the singular values and $V^{*}$ is the conjugate transpose of $V$; the MP Generalized Inverse $A^{-P}$ is defined by $A^{-P} = (USV^{*})^{-P} = VS^{\widetilde{-1}}U^{*}$, where $U^{*}$ is the conjugate transpose of the matrix $U$, and $S^{\widetilde{-1}}$ is the inverse of the diagonal matrix $S$. The property established by this equation implies that the MP inverse is applicable to problems defined in a certain Euclidean space for which the behavior of the system should be invariant with respect to arbitrary rotations of the coordinate frame. Unfortunately, the MP inverse has been shown to fail, as it does not always provide reliable and stable control in the case of incommensurate manipulators \cite{zhang2019applying} when units are varied -- it is hence said that the MP inverse does not satisfy unit consistency.

\subsubsection{Unit-Consistent (UC) Inverse}\label{sec:UC-inverse}  UC  is a GI defined with the property of invariance with respect to the choice of units used on different state variables, e.g, miles, kilometers, meters, centimeters, etc. It can be expressed based on the Unit-Invariant Singular Valued Decomposition (UI-SVD)\cite{uhlmann2018generalized}. Hence, the UI-SVD of $A$ is given by $A = DA^{'}E  = D(USV^{*})E$, where $D$ and $E$ are diagonal scale matrices resulting from the scaling decomposition \cite{rothblum1992scalings} of $A$, that takes into account the scale due to the change of units; and $A^{'}$ is a matrix satisfying row and column product constraints. From this equation, the formula of the UC inverse can be expressed by $A^{-U} = \bigg(D(USV^{*})E\bigg)^{-U} = E^{-U}VS^{-U}U^{*}D^{-U} = E^{-1}VS^{-U}U^{*}D^{-1}$. While the UC inverse is consistent with respect to arbitrary changes of units, it may also fail to provide consistent inverses in the presence of rotation of the reference frame or any frame defined in the D-H representation of the robot that appears earlier than the frame where the unit change occurred. In those cases, we need to systematically provide both rotation and unit consistencies through another GI (i.e. the MX inverse \cite{uhlmann2018generalized}).

\subsubsection{Mixed (MX) Inverse}\label{sec:MX-inverse}  MX is an inverse that selectively provides invariance with respect to arbitrary changes of units as well as with respect to rotations \cite{uhlmann2018generalized}. The MX inverse is derived using the concept of block matrix inverse, where variables requiring unit consistency are block-partitioned in the top left of $A$, and the variables requiring rotation consistency are block-partitioned in the bottom right of $A$. This partitioning is expressed as:
$A = \begin{bmatrix}
A_W & A_X\\
A_Y & A_Z
\end{bmatrix}$, where $A_W$ is the block of variables requiring unit-consistency, $A_Z$ is the block of variables requiring rotation consistency, $A_X$ and $A_Y$ are blocks of variables requiring rotation and unit consistencies. Once, the variables of $A$ have been partitioned, the block-matrix inverse is applied to $A$ to compute its MX inverse:
\begin{equation} \label{eq7}
{\footnotesize{}A^{-M}=\left[\begin{array}{cc}
(A_W-A_XA_Z^{-P}A_Y)^{-U} & -A_W^{-U}A_X(A_Z-A_YA_W^{-U}A_X)^{-P}\\
-A_Z^{-P}A_Y(A_W-A_XA_Z^{-P}A_Y)^{-U} & (A_Z-A_YA_W^{-U}A_X)^{-P}
\end{array}\right]}{\footnotesize}
\end{equation}

\subsubsection{Other Inverses}\label{sec:other-inverses}  However, they are other inverse Jacobian methods used in the literature as summarized in Table \ref{tab:review-IK-methods}. The reader should report to \cite{penrose1955generalized, chan1988general, chiaverini1991achieving, chiaverini1994review, buss2005selectively, sugihara2011solvability, colome2012redundant, uhlmann2018generalized} for more details about these methods.

\begin{table}[t] %!htb
\footnotesize 
    \caption{\footnotesize Investigated inverse Jacobian methods}
      \centering
         \begin{tabular}{|c | c | c | c|} 
         \hline
         Method & Abbreviation & Reference & Year\\  
         \hline\hline
         Moore-Penrose & MP & \cite{penrose1955generalized} & 1955\\ 
         \hline
         Error Damping & ED & \cite{chan1988general} & 1988\\
         \hline
         Filtered Jacobian & JF & \cite{chiaverini1991achieving} & 1991\\
         \hline
         Damped Jacobian & JD & \cite{chiaverini1994review} & 1994\\
         \hline
         Selective Damping & SD & \cite{buss2005selectively} & 2005\\
         \hline
         Improve Error Damping & IED & \cite{sugihara2011solvability} & 2011\\
         \hline
         Singular Value Filtering & SVF & \cite{colome2012redundant} & 2012\\
         \hline
         Unit-Consistent & UC & \cite{uhlmann2018generalized} & 2018\\
         \hline
         Mixed & MX & \cite{uhlmann2018generalized} & 2018\\
         \hline
        \end{tabular}
    \label{tab:review-IK-methods}
    \vspace{-5mm}
\end{table}

In general, robotics applications that require matrix inversion of the Jacobian will employ the two-sided inverse whenever the same Jacobian matrix is square and non-singular. However, when that is not the case -- i.e. for temporarily singular configurations of the robot and in the cases of over ($>$6DoF) or under ($<$6DoF) determined system, as illustrated later -- left/right pseudo-inverses and the MP inverse are typically a substitution for the two-sided inverse. As explained before, this can represent a problem for systems requiring properties such as unit and/or rotation invariance \cite{schwartz2003non, uhlmann2018generalized}. Moreover, even if no particular need of using UC is identified for the corresponding (i.e. 6DoF) robot, providing an implementation using real inverse is not a solution as robots undergo singular configurations during a trajectory towards its goal. In this paper, the three GI's, described in the subsections \ref{sec:MP-inverse}, \ref{sec:UC-inverse} and \ref{sec:MX-inverse}, are exploited to create a systematic solution for the IK computation of any arbitrary incommensurate serial robot using a Jacobian-based IK solver such as in \cite{farzan2013dh, colome2012redundant, colome2014closed}. The keyword here is “systematically”: a task that, so far, had been done manually by human inspection, and its automation is one of the main contributions of this work. The proposed inverse solution is also compared to other well accepted inverse Jacobian methods mentioned in subsection \ref{sec:other-inverses}.

\begin{algorithm}[!ht]
\caption{Jacobian-based IK Solver using all the GI's}\label{alg:euclid}
\begin{algorithmic}[1] 
\Procedure{IK}{$\mathbold{Q}_{t_0}$, $\mathbold{D}_{final}$, $DH$, $\epsilon_{r}$, $J_{type}$, $invJ_{method}$}
\State $\textit{Assign } A_{w}, A_{x}, A_{y}, A_{z} \textit{ matrices according to DH}$ 
\State $\mathbold{D}_{t} \Leftarrow \mathbold{D}_{t_0} = f_{DH}({\mathbold{Q_{t_0}}})$ 
	\While{$\|\mathbold{D_t} - \mathbold{D}_{final}\|>\epsilon_{r}$} 
        \State $J_{t} = getJacobian(\mathbold{D}_{t}, \mathbold{Q}_{t}, J_{type})$
        \State $J^{\widetilde{-1}}_t = getInverseJacobian(\mathbold{J}, invJ_{method})$
        \State $\Delta\mathbold{Q_{t}} = J^{\widetilde{-1}}_t*\alpha_{t}(\mathbold{D}_{final} - \mathbold{D_t})$
		\State $\mathbold{Q}_{t+1} = \mathbold{Q_{t}} + \Delta\mathbold{Q_{t}}$ 
		\State $\mathbold{D}_{t} \Leftarrow \mathbold{D}_{t+1} =  f_{DH}(\mathbold{Q_{t+1}})$ 
	\EndWhile\label{euclidendwhile} 
\State \textbf{return} $\mathbold{Q}_{final}\Leftarrow  \mathbold{Q}_{t+1} $ 
\EndProcedure 
\end{algorithmic}
\label{alg1}
\end{algorithm}

\vspace{-4mm}
\section{Proposed Approach} \label{sec:proposed-algorithm}
 All the main steps followed in this approach can be found in the pseudo-code  presented in Algorithm \ref{alg1}. This algorithm, which is based on the Jacobian-based IK solver framework, mainly takes the initial joints configuration $\vec{Q}_{t_0}$, the final desired pose $\vec{X}_{final}$, the $DH$ parameters, the desired pose error $\epsilon_{r}$, the Jacobian type $J_{type}$, and the inverse Jacobian method $invJ_{method}$ as input parameters to compute the IK solution. In this paper, we focus on three important aspects of this algorithm embedded in line 7 and expressed in the motion control equation $\Delta\overrightarrow{Q_t}=J_{t}^{\widetilde{-1}}*\alpha_{t}(\overrightarrow{D}_{final}-\overrightarrow{D}_{t})$: the attenuation parameter $\alpha_{t}$, and the inverse Jacobian $J_{t}^{\widetilde{-1}}$ which itself depends on the Jacobian $J_{t}$. The attenuation parameter $\alpha_{t}$ is selected between $\left[0,1\right]$ to smooth the path of the end-effector and control the convergence of the algorithm. The inverse Jacobian ($J_{t}^{\widetilde{-1}}$) denotes the use of a GI, as mentioned in Table \ref{tab:review-IK-methods}, to find the inverse Jacobian matrix. Most of the time, $J_{t}^{\widetilde{-1}}$ is calculated using the MP ($J_{t}^{-P}$) inverse, hence $J_{t}^{\widetilde{-1}} \Leftarrow J_{t}^{-P}$. As previously explained in Section \ref{sec:background}, the MP inverse cannot always guarantee reliable IK solutions in the presence of incommensurate systems. So here, we investigate the use of the MX ($J_{t}^{-M}$) inverse to achieve more reliable and stable IK solutions through the use of the D-H methodology. In the D-H, a sequence of coordinate frames is attached to each of the robot joints. Each coordinate frame moves with respect to the previous coordinate frames, but it is stationary with respect to motions of the following frames. Since every movement is either a rotation or a translation with respect to the $Z-axis$ and there is no movement with respect to the other 2 axes, we can define a systematic way to assign joint variables to the blocks $A_{W}$, $A_{X}$, $A_{Y}$ and $A_{Z}$ for the computation of the MX inverse based on the sequence of prismatic and revolute joints according to the D-H table as depicted in line 2 of algorithm \ref{alg1}.

We formulate the following rule of thumb for the use of the MX inverse: all revolute joints appearing before a prismatic joint of interest whose $Z-axis$  are not parallel need to be included in $A_W$. That is because if the $Z-axis$ of a revolute joint prior to a prismatic joint are not parallel, a rotation caused by the revolute joint will affect the prismatic joint, which will violate unit-consistency unless they are placed in the $A_W$ block partitioning where they can be handled by the UC inverse. On the other hand, if the two $Z-axis$ are parallel, the revolute joint will not affect the prismatic joint, and hence they need to be placed in $A_Z$, so that they can be handled by the MP inverse. This rule of thumb needs to be checked continuously as dynamic configurations of the robot can lead to temporary alignment of the $Z-axis$ even when they are not explicit in the D-H table.

\begin{table}[!ht] %!htb
\footnotesize 
\centering
    \caption{\footnotesize D-H Parameters of the serial robots used in the experiments. The angles $\theta$ and $\alpha$ are expressed in degrees. The variables $d$ and $a$ are shown in millimeters ($mm$) – but they were changed in the implementation to match the different choices of units.}
    \begin{subtable}{.46\linewidth}
      \centering
         \begin{tabular}{|c | c | c | c | c|} 
         \hline
         $i$ & $\theta$ & $d$ & $a$ & $\alpha$  \\  
         \hline\hline
         1 & $\theta_1$ & 0 & 1000 & 0\\ 
         \hline
         2 & $\theta_2$ & 0 & 1100 & 90 \\
         \hline
         3 & 0 & $d_3$ & 0 & 0 \\
         \hline
        \end{tabular}
        \medskip
        \caption{\footnotesize 3 DoF Planar arm}
    \end{subtable}
    \medskip
    \begin{subtable}{.46\linewidth}
      \centering
         \begin{tabular}{|c | c | c | c | c|} 
         \hline
         $i$ & $\theta$ & $d$ & $a$ & $\alpha$  \\  
         \hline\hline
         1 & $\theta_1$ & 400 & 250 & 0\\ 
         \hline
         2 & $\theta_2$ & 0 & 150 & 180 \\
         \hline
         3 & 0 & $d_3$ & 0 & 0 \\
         \hline
         4 & $\theta_4$ & 150 & 0 & 0 \\
         \hline
        \end{tabular}
        \medskip
        \caption{\footnotesize 4 DoF Scara arm}
    \end{subtable} 
    \medskip
    \begin{subtable}{.5\linewidth}
      \centering
         \begin{tabular}{|c | c | c | c | c|} 
         \hline
         $i$ & $\theta$ & $d$ & $a$ & $\alpha$  \\  
         \hline\hline
         1 & $\theta_1$ & 0 & 0 & -90\\ 
         \hline
         2 & $\theta_2$ & 140 & 0 & 90 \\
         \hline
         3 & 0 & $d_3$ & 0 & 0 \\
         \hline
         4 & $\theta_4$ & 0 & 0 & -90 \\
         \hline
         5 & $\theta_5$ & 0 & 0 & 90 \\
         \hline
        \end{tabular}
        \medskip
        \caption{\footnotesize 5 DoF (Modified) Stanford arm}
    \end{subtable}%
    \medskip
    \begin{subtable}{.5\linewidth}
      \centering
         \begin{tabular}{|c | c | c | c | c|} 
         \hline
         $i$ & $\theta$ & $d$ & $a$ & $\alpha$  \\  
         \hline\hline
         1 & $\theta_1$ & 0 & 0 & -90\\ 
         \hline
         2 & $\theta_2$ & 140 & 0 & 90 \\
         \hline
         3 & 0 & $d_3$ & 0 & 0 \\
         \hline
         4 & $\theta_4$ & 0 & 0 & -90 \\
         \hline
         5 & $\theta_5$ & 0 & 0 & 90 \\
         \hline
         6 & $\theta_6$ & 8.5 & 0 & 0 \\
         \hline
        \end{tabular}
        \medskip
        \caption{\footnotesize 6 DoF Stanford arm}
    \end{subtable} 
    \medskip
    \begin{subtable}{.49\linewidth}
      \centering
         \begin{tabular}{|c | c | c | c | c|} 
         \hline
         $i$ & $\theta$ & $d$ & $a$ & $\alpha$  \\  
         \hline\hline
         1 & $\theta_1$ & 0 & 0 & 90\\ 
         \hline
         2 & $\theta_2$ & 0 & 250 & 90 \\
         \hline
         3 & 0 & $d_3$ & 0 & 0 \\
         \hline
         4 & $\theta_4$ & 0 & 0 & 90 \\
         \hline
         5 & $\theta_5$ & 140 & 0 & 90 \\
         \hline
         6 & $\theta_6$ & 0 & 0 & 90 \\
         \hline
         7 & $\theta_7$ & 0 & 0 & 0 \\
         \hline
        \end{tabular}
        \medskip
        \caption{\footnotesize 7 DoF GP66+1 arm}
    \end{subtable}
    \medskip
    \begin{subtable}{.49\linewidth}
      \centering
         \begin{tabular}{|c | c | c | c | c|} 
         \hline
         $i$ & $\theta$ & $d$ & $a$ & $\alpha$  \\  
         \hline\hline
         1 & $\theta_1$ & 0 & 0 & -90\\ 
         \hline
         2 & $\theta_2$ & 0 & 0 & 90 \\
         \hline
         3 & $\theta_3$ & 550 & 45 & -90 \\
         \hline
         4 & $\theta_4$ & 0 & -45 & 90 \\
         \hline
         5 & $\theta_5$ & 300 & 0 & -90 \\
         \hline
         6 & $\theta_6$ & 0 & 0 & 90 \\
         \hline
         7 & $\theta_7$ & 60 & 0 & 0 \\
         \hline
        \end{tabular}
        \medskip
        \caption{\footnotesize 7 DoF WAM arm}
    \end{subtable}
    \label{tab:DH-parameters}
    \vspace{-12mm}
\end{table}

\vspace{-3mm}
\section{Experimental Results and Discussion} \label{sec:experimental-results}
In this section, we present and discuss the experiments that were performed when using the rule of thumb introduced in section \ref{sec:generalized-inverses}. For these experiments, the numerical IK algorithm presented in section \ref{sec:proposed-algorithm} was applied to various configurations and motions of six manipulators (five incommensurate and one commensurate): 1) a 3DoF planar arm; 2) a 4DoF SCARA (Selective Compliance Articulated Robot Arm)  arm; 3) a 5DoF (modified) Stanford  arm; 4) a 6DoF Stanford  arm; 5) a 7DoF GP66+1 arm, and 6) a 7DoF WAM arm. Their D-H parameters are presented in Table \ref{tab:DH-parameters}. These arm robots were chosen to illustrate various cases of the application of the MX inverse in the computation of the inverse Jacobian while correctly/incorrectly selecting the variables to be included in each block partition. Correct selections will achieve unit-invariant behaviors of the robot end-effector, while incorrect ones will lead to inappropriate use of MP and UC inverses inside the MX inverse, which will lead to uncontrollable situations (oscillations) when the units are varied from $m$ to $mm$.

\begin{table*}[hbt!]
\footnotesize 
\centering
\caption{\footnotesize Example behavior of the Geometric Jacobian ($J_{G}$) and Inverse Jacobian ($J^{\widetilde{-1}}$) of the 3DoF (2RP) manipulator estimated at the joint configuration $\mathbold{Q} = [\theta_1 = 30^{o}, \theta_2=30^{o}, d_3=-0.7m]$. The prismatic joint $d_3$ is varied from $m$ (meter), $dm$ (decimeter), $cm$ (centimeter), $mm$ (millimeter).} 
\begin{threeparttable}
         \begin{tabular}{|c|c|c|c|c|} 
         %\begin{tblr}{
         %       colspec = {|c|c|c|c|c|},
         %       row{9} = {gray9},
         %       row{10} = {gray9},
                %column{3} = {teal7},
                %cell{2}{3} = {yellow7},
         %     }
         \hline
         \textbf{Method} & $m$ & $dm$ & $cm$ & $mm$\\   
         \hline\hline
         %\multicolumn{5}{|c|}{\textbf{Geometric Jacobian}}\\  
         %\hline
         $J_{G}$ 
          & 
          ${\scriptsize{} \begin{bmatrix}
            -1.8026 &  -1.3026  &  0.866\\
             0.8098 &  -0.0562  & -0.500\\
            \end{bmatrix}{\scriptsize}}$
          & 
          ${\scriptsize{} \begin{bmatrix}
              -18.026 & -13.026 &   0.866 \\
                8.098 &  -0.562 &  -0.500 \\
            \end{bmatrix}{\scriptsize}}$
          & 
          ${\scriptsize{} \begin{bmatrix}
             -180.26 & -130.26 &   0.866 \\
               80.98 &  -5.62 &  -0.500 \\
            \end{bmatrix}{\scriptsize}}$
          & 
          ${\scriptsize{} \begin{bmatrix}
             -1802.6 & -1302.6 &   0.866 \\
               809.8 &  -56.2 &  -0.500 \\
            \end{bmatrix}{\scriptsize}}$ \\
         \hline
         MP 
          & 
        ${\scriptsize{} \begin{bmatrix}
               -0.08838467  &  0.71399628 \\
               -0.68903295  & -1.44117808 \\
               -0.06567735  & -0.68156105 \\
            \end{bmatrix}{\scriptsize}}$
          & 
        ${\scriptsize{} \begin{bmatrix}
               -0.00491752  &  0.11208890 \\
               -0.07002356  & -0.15574331 \\
               -0.00091353  & -0.00948012 \\
            \end{bmatrix}{\scriptsize}}$
          & 
        ${\scriptsize{} \begin{bmatrix}
               -0.00048627 &   0.01126570 \\
               -0.00700392 &  -0.01559056 \\
               -0.00000917 &  -0.00009517 \\
            \end{bmatrix}{\scriptsize}}$
          & 
        ${\scriptsize{} \begin{bmatrix}
                   -0.00004862 &   0.00112662 \\
                   -0.00070039 &  -0.00155907 \\
                   -0.00000009 &  -0.00000095 \\
            \end{bmatrix}{\scriptsize}}$ \\
         \hline
         ED 
          & 
        ${\scriptsize{} \begin{bmatrix}
                   -0.14349153 &   0.51554428 \\
                   -0.47992060 &  -0.87327526 \\
                    0.00385736 &  -0.46317921 \\
            \end{bmatrix}{\scriptsize}}$
          & 
        ${\scriptsize{} \begin{bmatrix}
                   -0.00493348 &   0.11203080 \\
                   -0.06998827 &  -0.15563505 \\
                   -0.00091195 &  -0.00947473 \\
            \end{bmatrix}{\scriptsize}}$
          & 
        ${\scriptsize{} \begin{bmatrix}
                   -0.00048627 &   0.01126570 \\
                   -0.00700391 &  -0.01559055 \\
                   -0.00000917 &  -0.00009517 \\
            \end{bmatrix}{\scriptsize}}$
          & 
        ${\scriptsize{} \begin{bmatrix}
                   -0.00004862 &   0.00112662 \\
                   -0.00070039 &  -0.00155907 \\
                   -0.00000009 &  -0.00000095 \\
            \end{bmatrix}{\scriptsize}}$ \\ 
         \hline
         FJ 
          & 
        ${\scriptsize{} \begin{bmatrix}
                   -0.11303971 &   0.64082068 \\
                   -0.62873610 &  -1.26221842 \\
                   -0.04032062 &  -0.60630287 \\
            \end{bmatrix}{\scriptsize}}$
          & 
        ${\scriptsize{} \begin{bmatrix}
                   -0.00496554 &   0.11192648 \\
                   -0.06994696 &  -0.15548427 \\
                   -0.00090930 &  -0.00946581 \\
            \end{bmatrix}{\scriptsize}}$
          & 
        ${\scriptsize{} \begin{bmatrix}
                   -0.00048632 &   0.01126554 \\
                   -0.00700384 &  -0.01559030 \\
                   -0.00000917 &  -0.00009517 \\
            \end{bmatrix}{\scriptsize}}$
          & 
        ${\scriptsize{} \begin{bmatrix}
               -0.00004862 &   0.00112662 \\
               -0.00070039 &  -0.00155907 \\
               -0.00000009 &  -0.00000095 \\
            \end{bmatrix}{\scriptsize}}$ \\
         \hline
         DJ 
          & 
        ${\scriptsize{} \begin{bmatrix}
                   -0.09000226 &   0.70880860 \\
                   -0.68471840 &  -1.42861179 \\
                   -0.06395170 &  -0.67624663 \\
            \end{bmatrix}{\scriptsize}}$
          & 
        ${\scriptsize{} \begin{bmatrix}
               -0.00492036  &  0.11207869 \\
               -0.07001866  & -0.15572713 \\
               -0.00091327  & -0.00947922 \\
            \end{bmatrix}{\scriptsize}}$
          & 
        ${\scriptsize{} \begin{bmatrix}
               -0.00048628 &   0.01126569 \\
               -0.00700391 &  -0.01559054 \\
               -0.00000917 &  -0.00009517 \\
            \end{bmatrix}{\scriptsize}}$
          & 
        ${\scriptsize{} \begin{bmatrix}
               -0.00004862 &   0.00112662 \\
               -0.00070039 &  -0.00155907 \\
               -0.00000009 &  -0.00000095 \\
            \end{bmatrix}{\scriptsize}}$ \\
         \hline
         IED 
          & 
        ${\scriptsize{} \begin{bmatrix}
                   -0.14596205  &  0.50656137 \\
                   -0.47328790  & -0.85391331 \\
                    0.00650038  & -0.45439157 \\
            \end{bmatrix}{\scriptsize}}$
          & 
        ${\scriptsize{} \begin{bmatrix}
                   -0.00494484  &  0.11199001 \\
                   -0.06996872  & -0.15557042 \\
                   -0.00091093  & -0.00947114 \\
            \end{bmatrix}{\scriptsize}}$
          & 
        ${\scriptsize{} \begin{bmatrix}
                   -0.00048628  &  0.01126566 \\
                   -0.00700389  & -0.01559048 \\
                   -0.00000917  & -0.00009517 \\
            \end{bmatrix}{\scriptsize}}$
          & 
        ${\scriptsize{} \begin{bmatrix}
                   -0.00004862 &   0.001126627 \\
                   -0.00070039 &  -0.001559072 \\
                   -0.00000009 &  -0.000000951 \\
            \end{bmatrix}{\scriptsize}}$ \\ 
         \hline
         SVF 
          &        
        ${\scriptsize{} \begin{bmatrix}
               -0.08929372 &   0.71106573 \\
               -0.68659419 &  -1.43408370 \\
               -0.06470513 &  -0.67855966 \\
            \end{bmatrix}{\scriptsize}}$
          & 
        ${\scriptsize{} \begin{bmatrix}
               -0.00491904 &   0.11208361 \\
               -0.07002105 &  -0.15573490 \\
               -0.00091339 &  -0.00947965 \\
            \end{bmatrix}{\scriptsize}}$
          & 
        ${\scriptsize{} \begin{bmatrix}
               -0.00048627 &   0.01126570 \\
               -0.00700392 &  -0.01559056 \\
               -0.00000917 &  -0.00009517 \\
            \end{bmatrix}{\scriptsize}}$
          & 
        ${\scriptsize{} \begin{bmatrix}
               -0.00004862 &   0.00112662 \\
               -0.00070039 &  -0.00155907 \\
               -0.00000009 &  -0.00000095 \\
            \end{bmatrix}{\scriptsize}}$ \\
         \hline
         \cellcolor{gray9}UC 
         &
        \cellcolor{gray9}${\scriptsize{} \begin{bmatrix}
                   -0.02734497 &   0.49301544 \\
                   -0.70647286 &  -1.37804070 \\
                    0.0351443 &  -1.04656388 \\
            \end{bmatrix}{\scriptsize}}$
          & 
        \cellcolor{gray9}${\scriptsize{} \begin{bmatrix}
                   -0.00273449 &   0.04930154 \\
                   -0.07064728 &  -0.13780407 \\
                    0.03514434 &  -1.04656388 \\
            \end{bmatrix}{\scriptsize}}$
          & 
        \cellcolor{gray9}${\scriptsize{} \begin{bmatrix}
                   -0.00027344 &   0.00493015 \\
                   -0.00706472 &  -0.01378040 \\
                    0.03514434 &  -1.04656388 \\
            \end{bmatrix}{\scriptsize}}$
          & 
        \cellcolor{gray9}${\scriptsize{} \begin{bmatrix}
                   -0.00002734 &   0.00049301 \\
                   -0.00070647 &  -0.00137804 \\
                    0.03514434 &  -1.04656388 \\
            \end{bmatrix}{\scriptsize}}$ \\
         \hline
         \cellcolor{gray9}MX 
          & 
        \cellcolor{gray9}${\scriptsize{} \begin{bmatrix}
                   -0.02734497 &   0.49301544 \\
                   -0.70647286 &  -1.37804070 \\
                    0.0351443 &  -1.04656388 \\
            \end{bmatrix}{\scriptsize}}$
          & 
        \cellcolor{gray9}${\scriptsize{} \begin{bmatrix}
                   -0.00273449 &   0.04930154 \\
                   -0.07064728 &  -0.13780407 \\
                    0.03514434 &  -1.04656388 \\
            \end{bmatrix}{\scriptsize}}$
          & 
        \cellcolor{gray9}${\scriptsize{} \begin{bmatrix}
                   -0.00027344 &   0.00493015 \\
                   -0.00706472 &  -0.01378040 \\
                    0.03514434 &  -1.04656388 \\
            \end{bmatrix}{\scriptsize}}$
          & 
        \cellcolor{gray9}${\scriptsize{} \begin{bmatrix}
                   -0.00002734 &   0.00049301 \\
                   -0.00070647 &  -0.00137804 \\
                    0.03514434 &  -1.04656388 \\
            \end{bmatrix}{\scriptsize}}$ \\
         \hline
        \end{tabular}
        %\end{tblr}
    \end{threeparttable}
    \label{tab:inverses-3DoF}
    \vspace{-4.5mm}
\end{table*}

\vspace{-4mm}
\subsection{Sanity test}
In this experiment, we investigated the behavior of all the inverse methods described in section \ref{sec:generalized-inverses} for the 3DoF (2RP) robot. That is, the Geometric Jacobian $J_{G}$ was computed at the joint configuration  $\vec{Q} = [\theta_1 = 30^{o}, \theta_2=30^{o}, d_3=-0.7m]$ while the unit of the prismatic joint $d_3$ was varied from $m$ to $mm$. Table \ref{tab:inverses-3DoF} presents the computed inverse Jacobians $J^{\widetilde{-1}}$ with eight floating point numbers to emphasize their differences. This table clearly shows how only the highlighted UC and MX inverses are providing consistent results across all units. As explained in section \ref{sec:proposed-algorithm}, applying the proposed MX rule of thumb for this manipulator happens to reduce the MX inverse to the UC inverse. We made the same observation when replacing  $J_{G}$ with $J_{A}$ or $J_{N}$; hence, revealing that the unit-consistency issues are inherent to all Jacobian types commonly used in IK solvers.

%%%%%%%%%%%%%%%%%%%%%%%%%%%%%%%%%%%%%%%%%%%%%%%%%%%%%%%%%%%%%%%%%%%%%%%%%%%%%
%%%%%%%%%%%%%%%%% 4DoF
%%%%%%%%%%%%%%%%%%%%%%%%%%%%%%%%%%%%%%%%%%%%%%%%%%%%%%%%%%%%%%%%%%%%%%%%%%%%%

\begin{figure}[hbt!]
\centering

\begin{subfigure}{.23\textwidth}
  \centering
  % include first image
  \includegraphics[width=4.6cm]{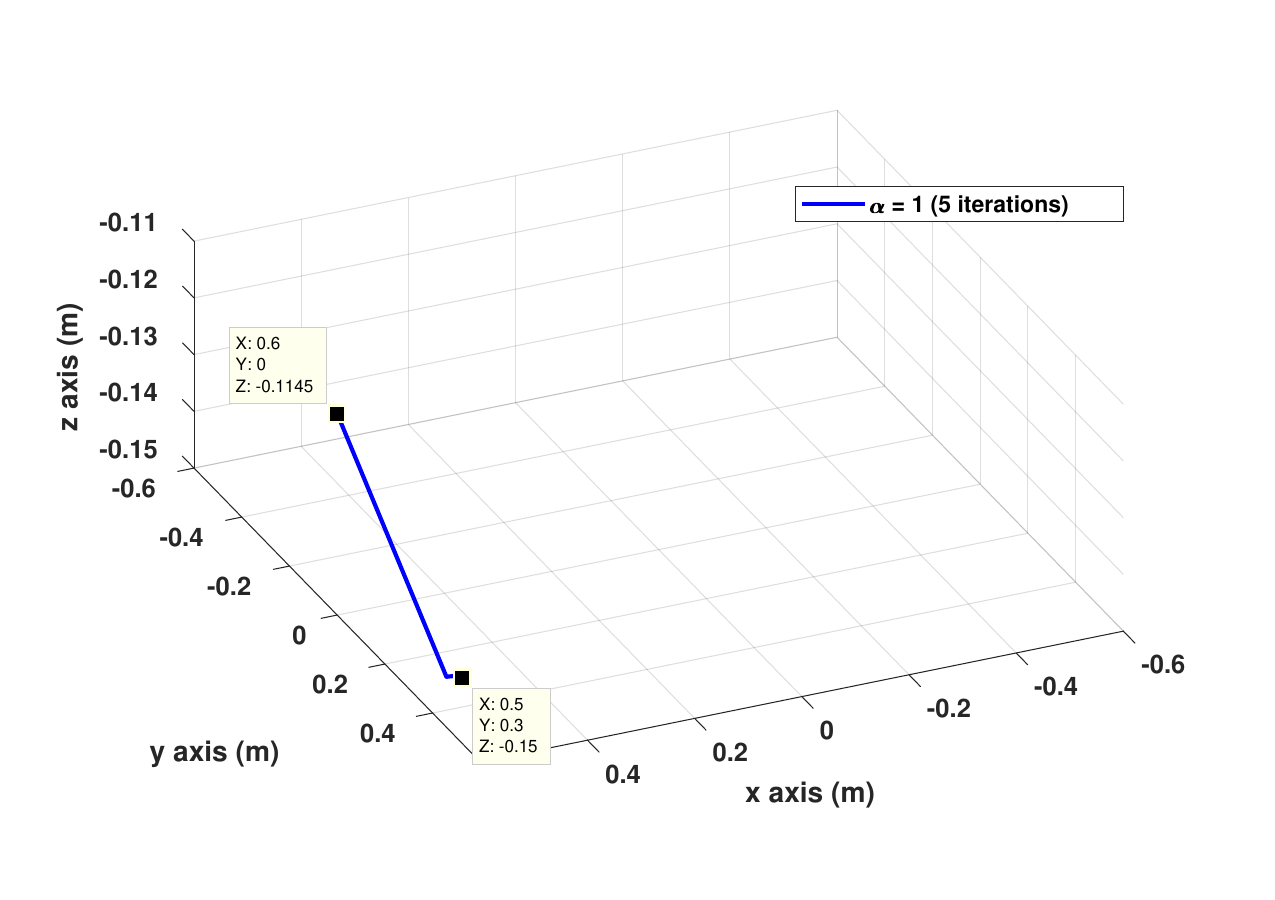}  
  \caption{\footnotesize MP / m / $\alpha=1$}
  \label{fig:4dof-mp-alpha-m}
\end{subfigure}
\begin{subfigure}{.23\textwidth}
  \centering
  % include third image
  \includegraphics[width=4.6cm]{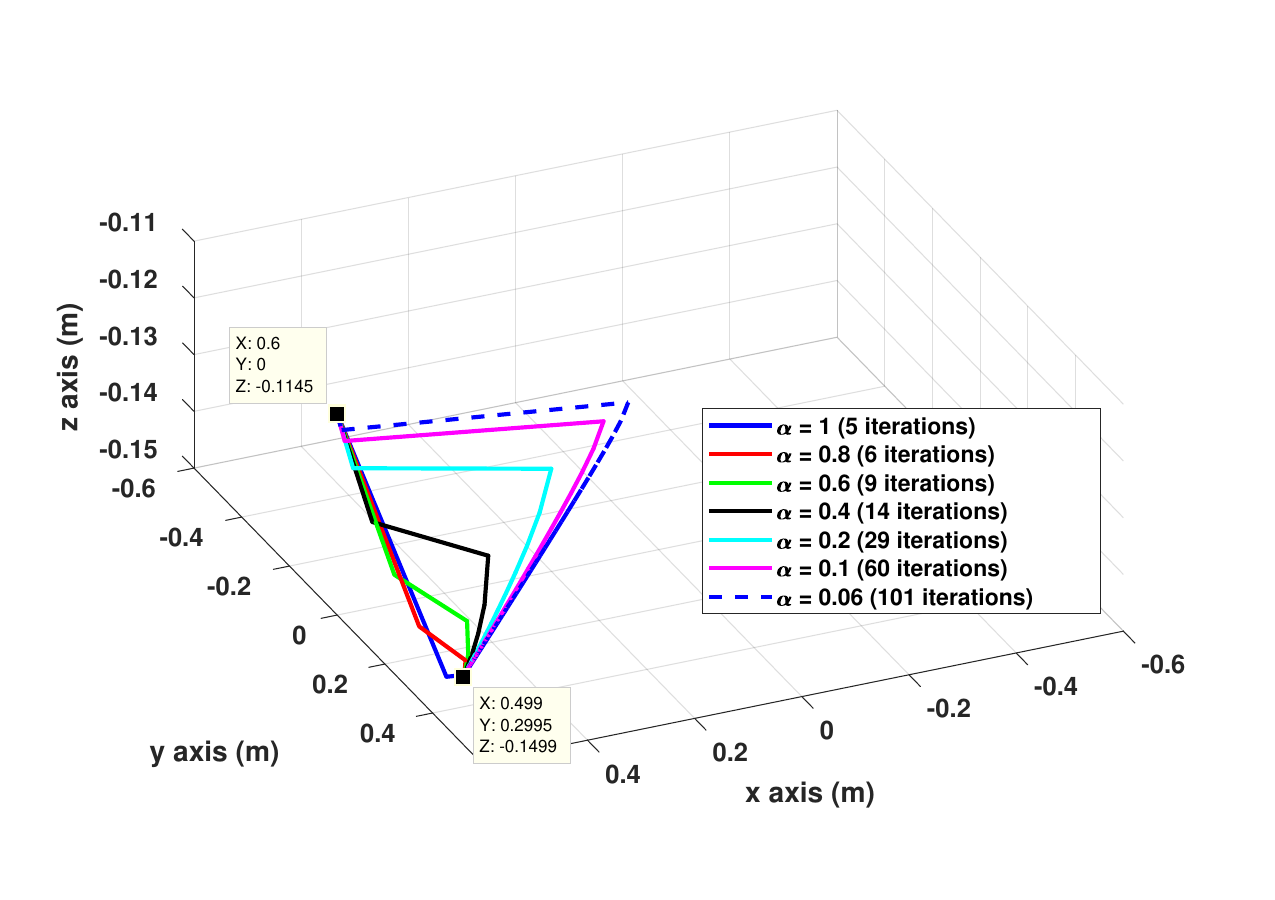}  
  \caption{\footnotesize MP / m / multiple $\alpha$}
  \label{fig:4dof-mp-alphas-m}
\end{subfigure}

\begin{subfigure}{.23\textwidth}
  \centering
  % include second image
  \includegraphics[width=4.6cm]{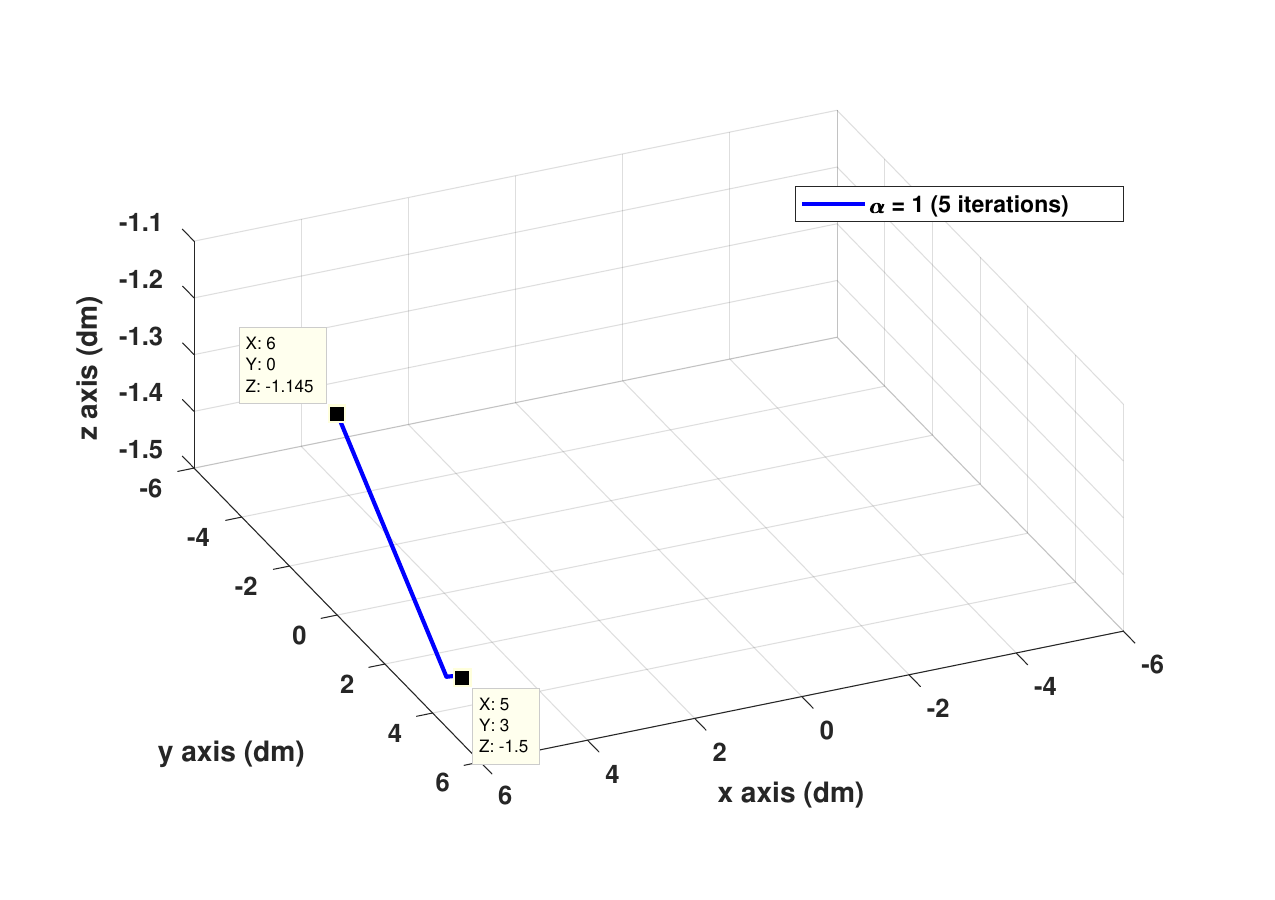}  
  \caption{\footnotesize MP / dm / $\alpha=1$}
  \label{fig:4dof-mp-alpha-dm}
\end{subfigure}
\begin{subfigure}{.23\textwidth}
  \centering
  % include fourth image
  \includegraphics[width=4.6cm]{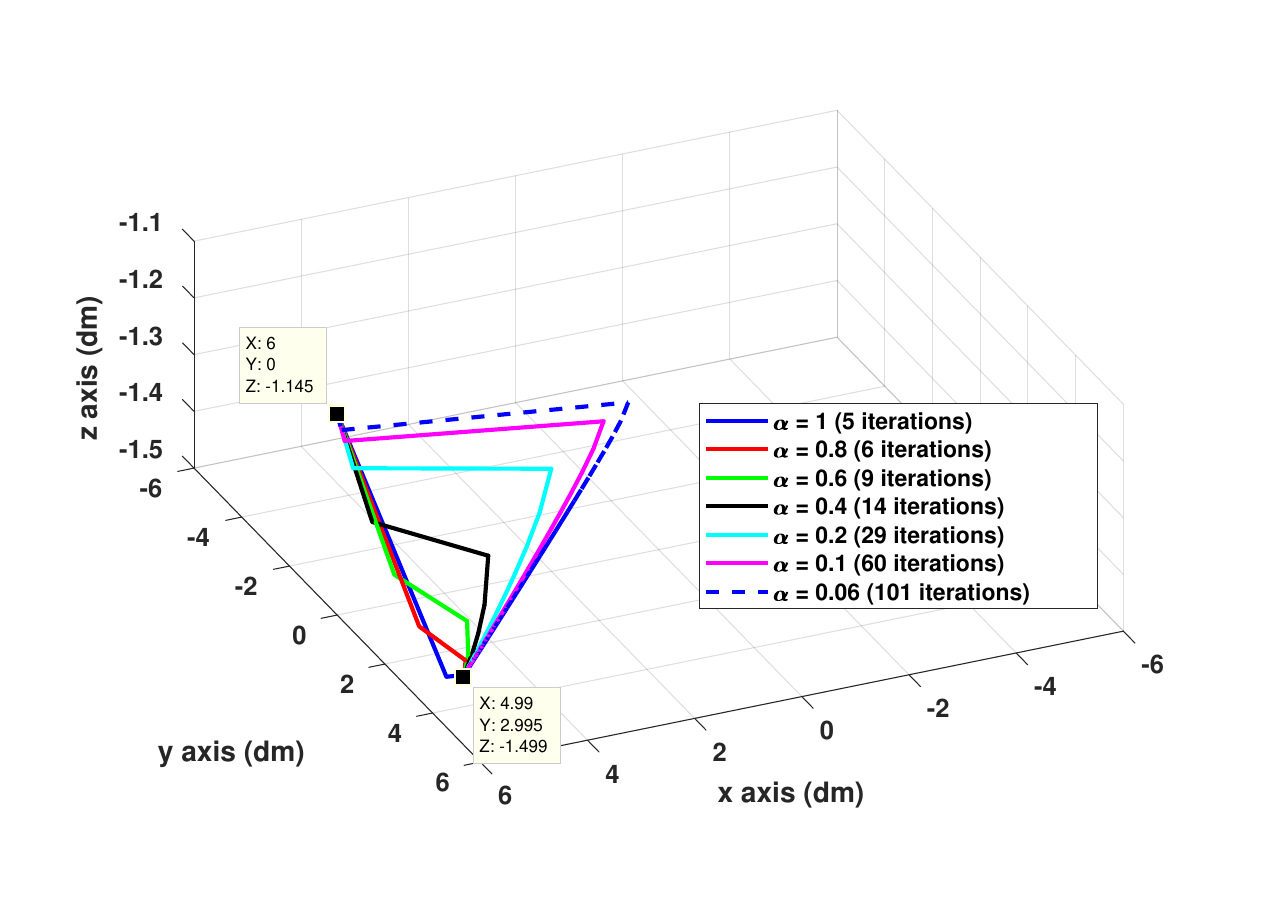}  
  \caption{\footnotesize MP / dm / multiple $\alpha$}
  \label{fig:4dof-mp-alphas-dm}
\end{subfigure}

\begin{subfigure}{.23\textwidth}
  \centering
  % include second image
  \includegraphics[width=4.6cm]{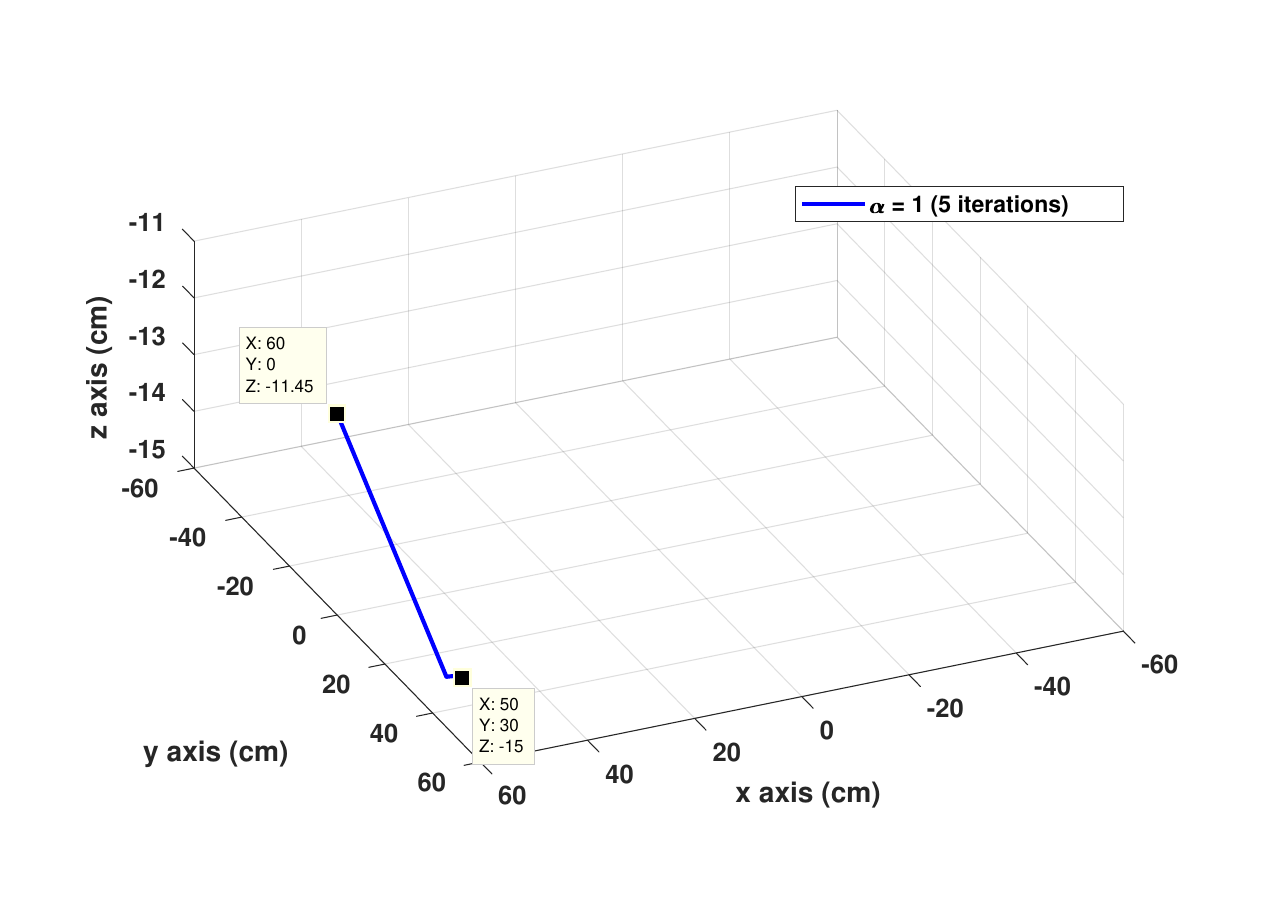}  
  \caption{\footnotesize MP / cm / $\alpha=1$}
  \label{fig:4dof-mp-alpha-cm}
\end{subfigure}
\begin{subfigure}{.23\textwidth}
  \centering
  % include fourth image
  \includegraphics[width=4.6cm]{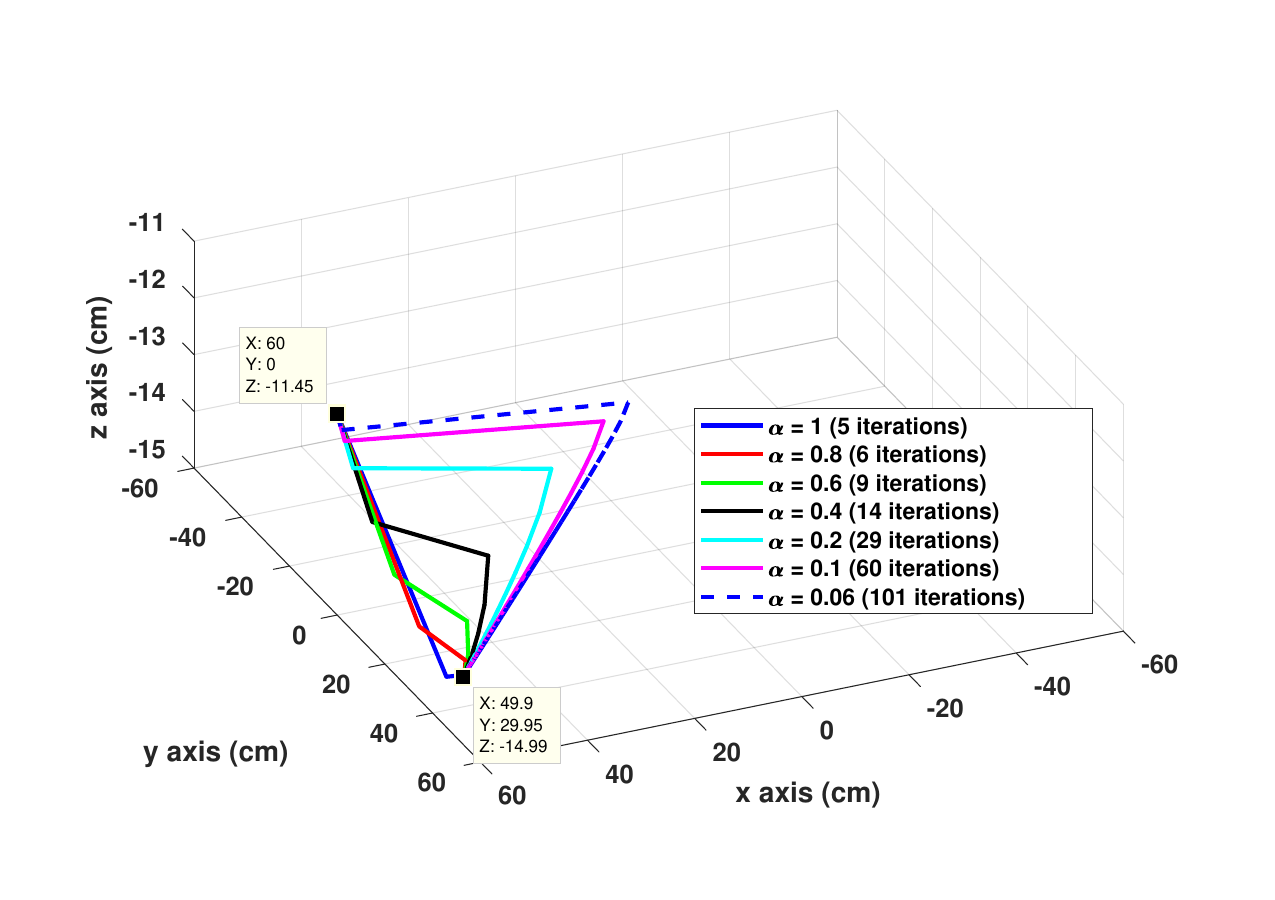}  
  \caption{\footnotesize MP / cm / multiple $\alpha$}
  \label{fig:4dof-mp-alphas-cm}
\end{subfigure}

\begin{subfigure}{.23\textwidth}
  \centering
  % include second image
  \includegraphics[width=4.6cm]{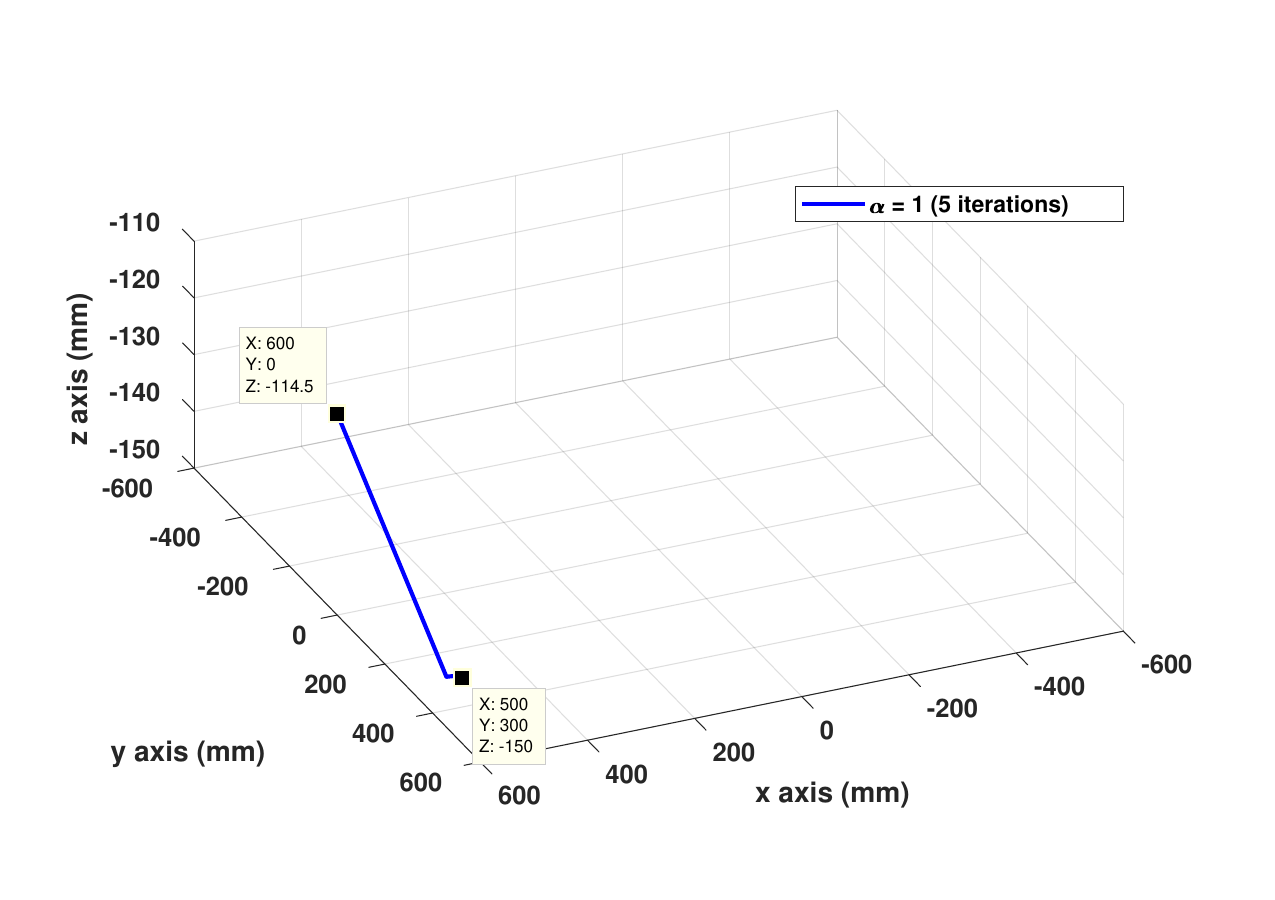}  
  \caption{\footnotesize MP / mm / $\alpha=1$}
  \label{fig:4dof-mp-alpha-mm}
\end{subfigure}
\begin{subfigure}{.23\textwidth}
  \centering
  % include fourth image
  \includegraphics[width=4.6cm]{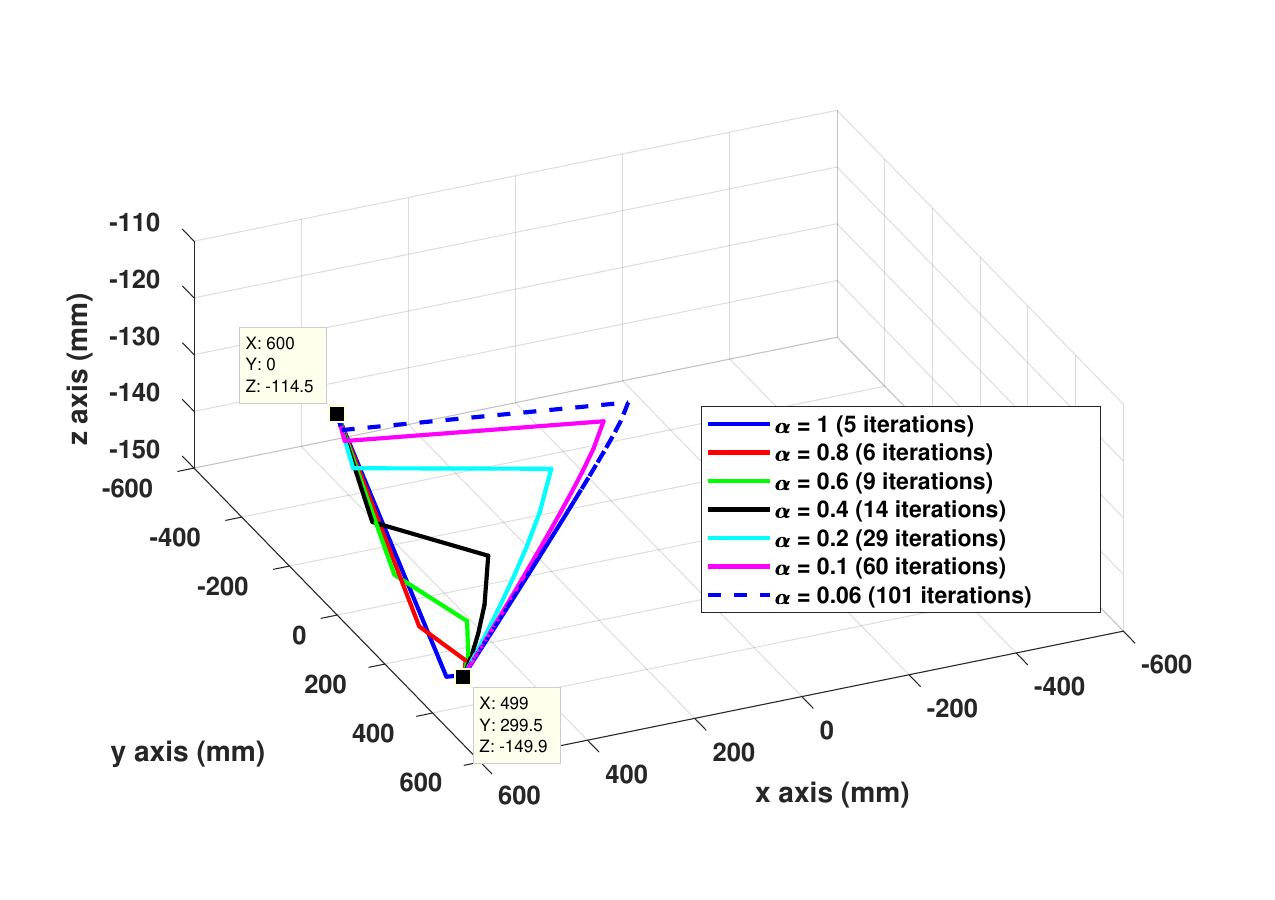}  
  \caption{\footnotesize MP / mm / multiple $\alpha$}
  \label{fig:4dof-mp-alphas-mm}
\end{subfigure}

\caption{\footnotesize Behavior of the trajectories of the end-effector of the 4DoF robot when varying the units while using the MP inverse with (\ref{fig:4dof-mp-alpha-m}),(\ref{fig:4dof-mp-alpha-dm}), (\ref{fig:4dof-mp-alpha-cm}), and (\ref{fig:4dof-mp-alpha-mm}) the attenuation parameter $\alpha = 1$ and (\ref{fig:4dof-mp-alphas-m}), (\ref{fig:4dof-mp-alphas-dm}), (\ref{fig:4dof-mp-alphas-cm}), and (\ref{fig:4dof-mp-alphas-mm})  multiple values of $\alpha$.}
\label{fig:trajectories-MP-4DoF-alpha(s)}
\vspace{-6mm}
\end{figure}

%\vspace{-1.5cm}
\begin{figure}
\centering
%%%%%%%%%%%%%%%%% UC Inverse - 4DoF alpha = 1
\begin{subfigure}{.23\textwidth}
  \centering
  % include first image
  \includegraphics[width=4.6cm]{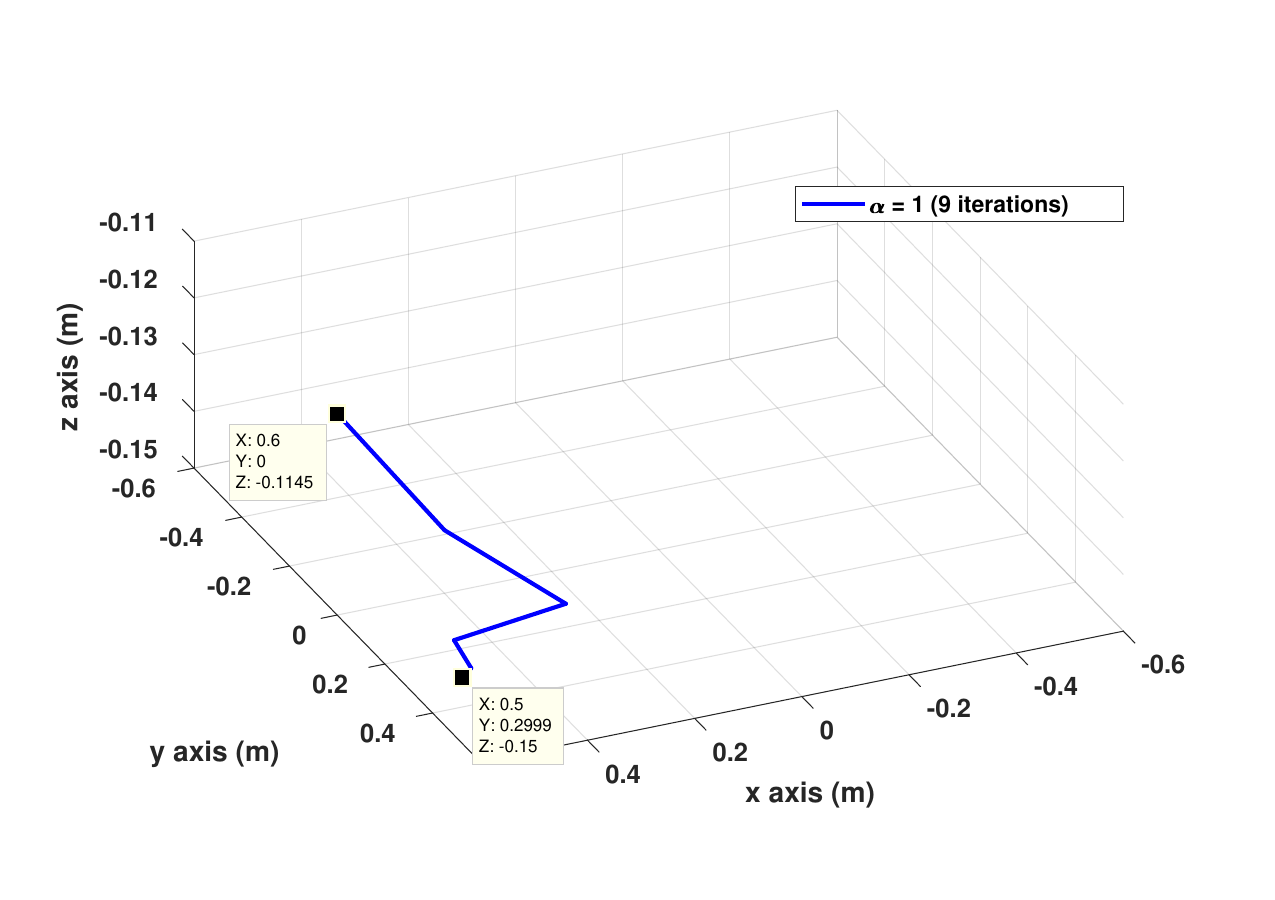}  
  \caption{UC / m / $\alpha=1$}
  \label{fig:4dof-uc-alpha-m}
\end{subfigure}
\begin{subfigure}{.24\textwidth}
  \centering
  % include third image
  \includegraphics[width=4.6cm]{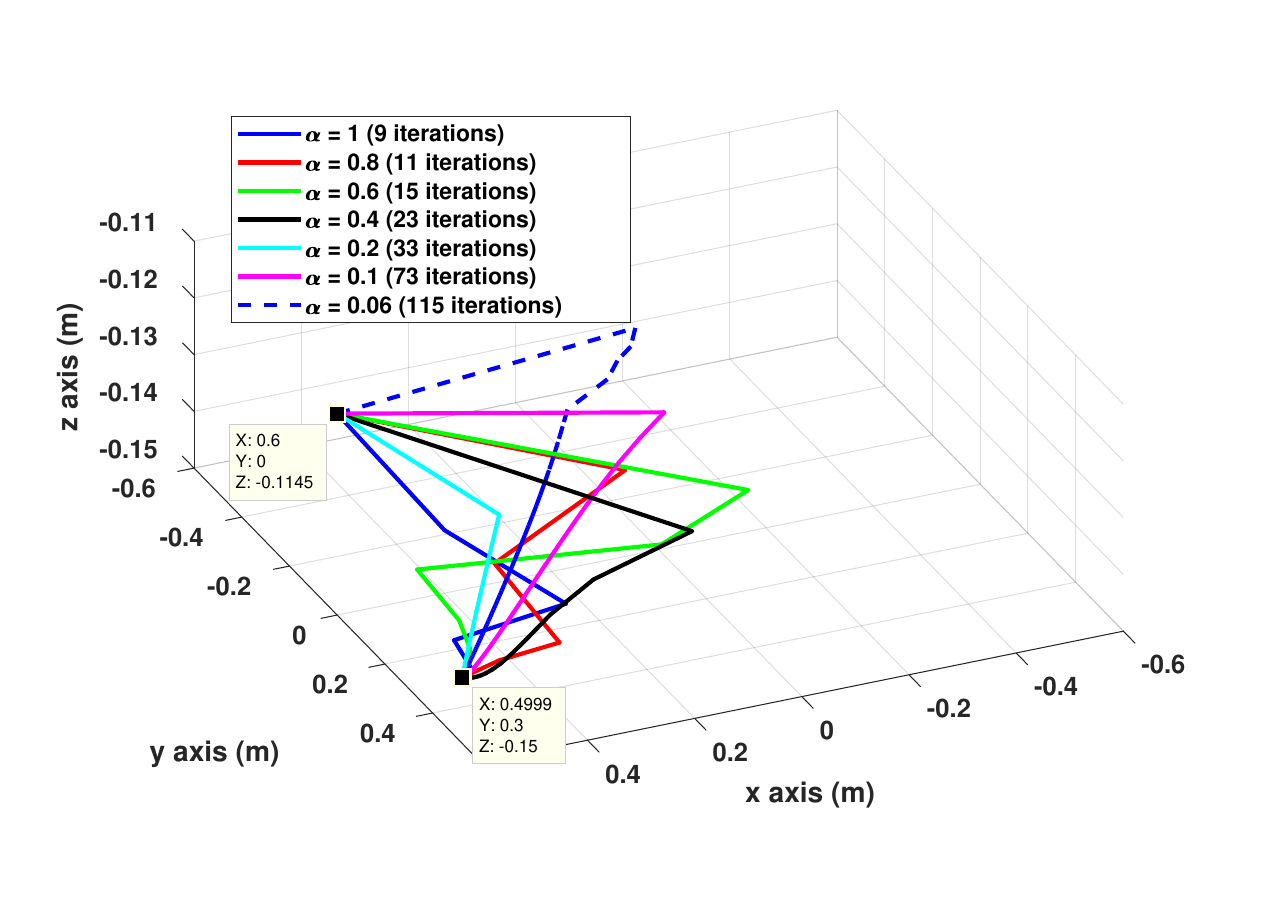}  
  \caption{\footnotesize UC / m / multiple $\alpha$}
  \label{fig:4dof-uc-alphas-m}
\end{subfigure}

\begin{subfigure}{.23\textwidth}
  \centering
  % include second image
  \includegraphics[width=4.6cm]{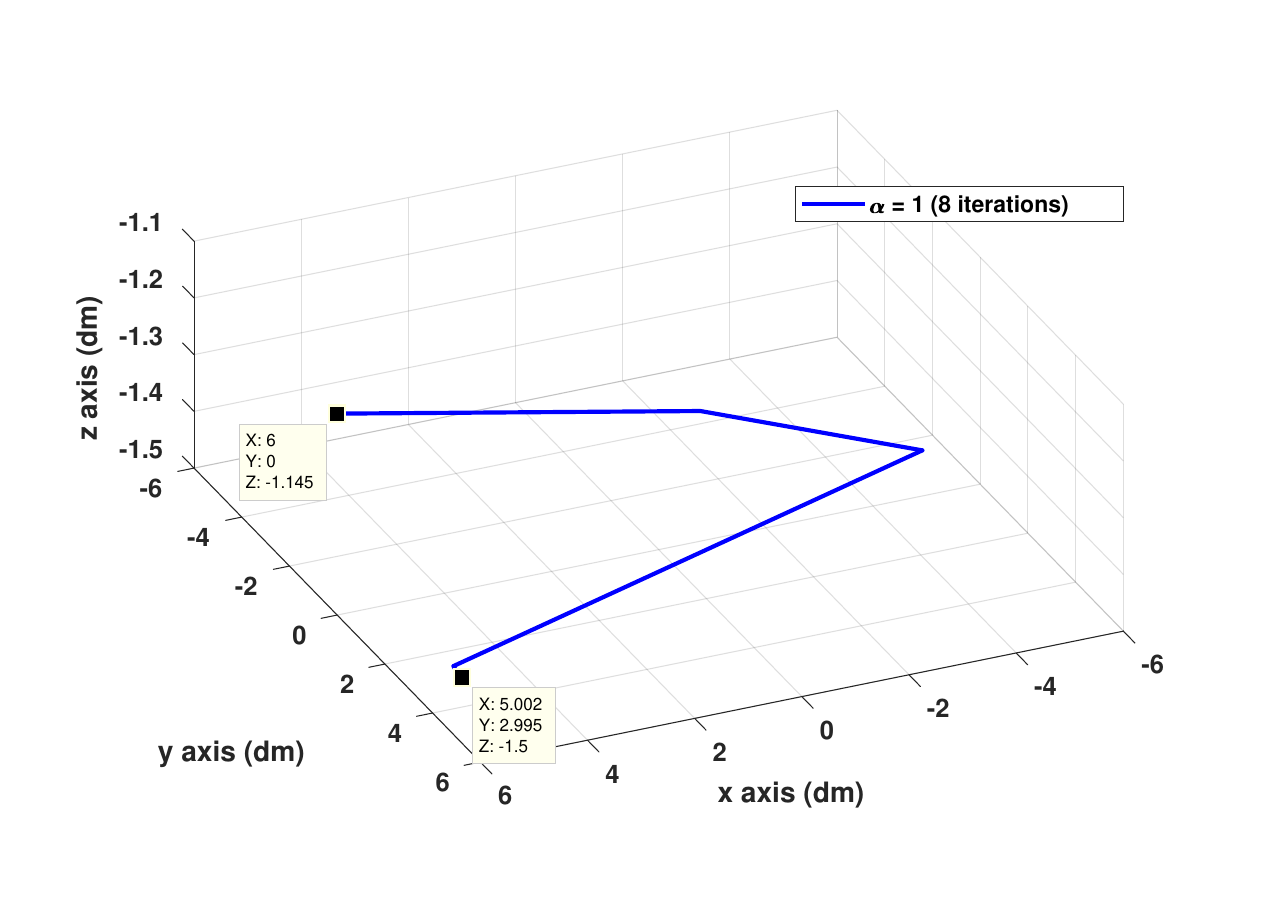}  
  \caption{\footnotesize UC / dm / $\alpha=1$}
  \label{fig:4dof-uc-alpha-cm}
\end{subfigure}
\begin{subfigure}{.23\textwidth}
  \centering
  % include fourth image
  \includegraphics[width=4.6cm]{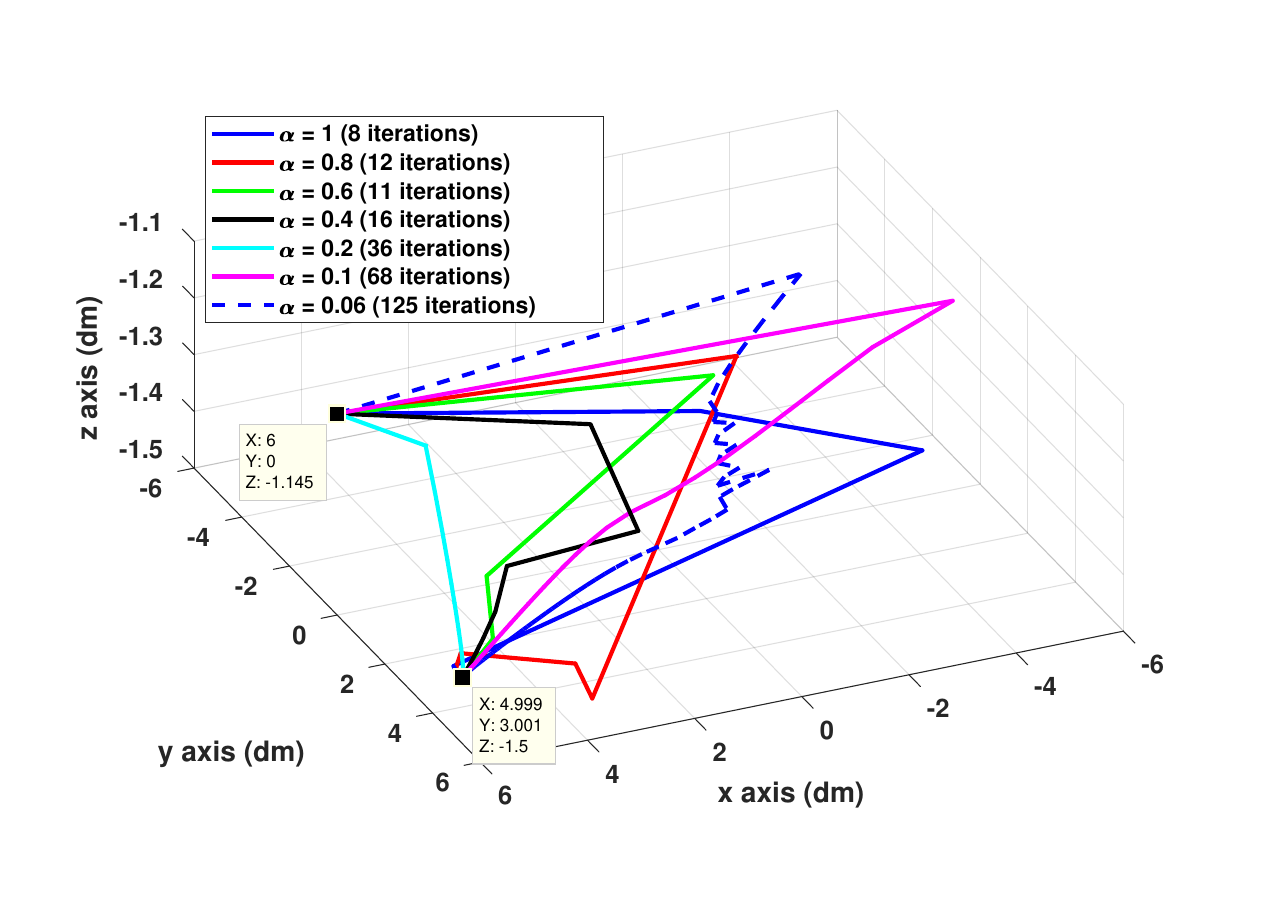}  
  \caption{\footnotesize UC / dm / multiple $\alpha$}
  \label{fig:4dof-uc-alphas-cm}
\end{subfigure}

\begin{subfigure}{.23\textwidth}
  \centering
  % include second image
  \includegraphics[width=4.6cm]{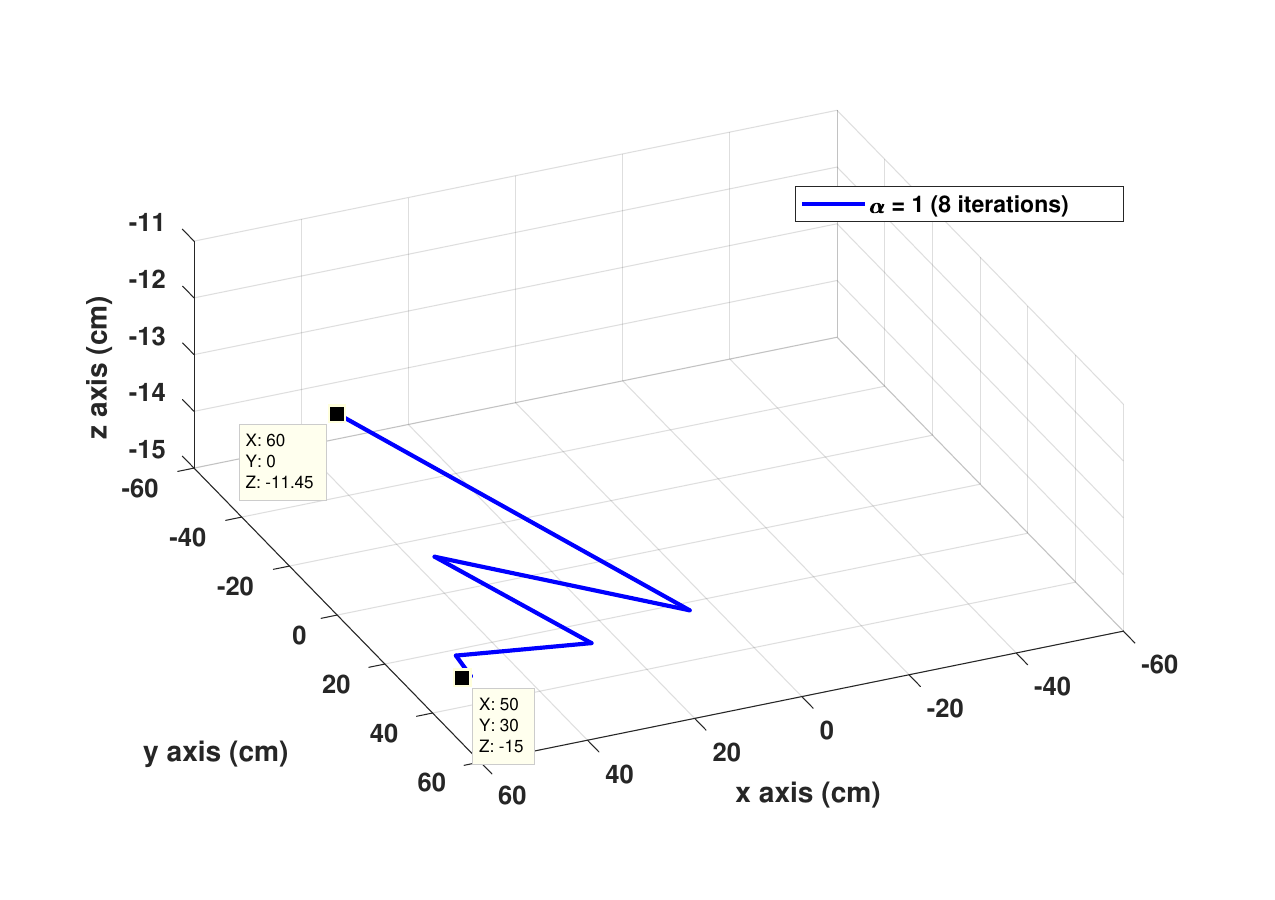}  
  \caption{\footnotesize UC / cm / $\alpha=1$}
  \label{fig:4dof-uc-alpha-dm}
\end{subfigure}
\begin{subfigure}{.23\textwidth}
  \centering
  % include fourth image
  \includegraphics[width=4.6cm]{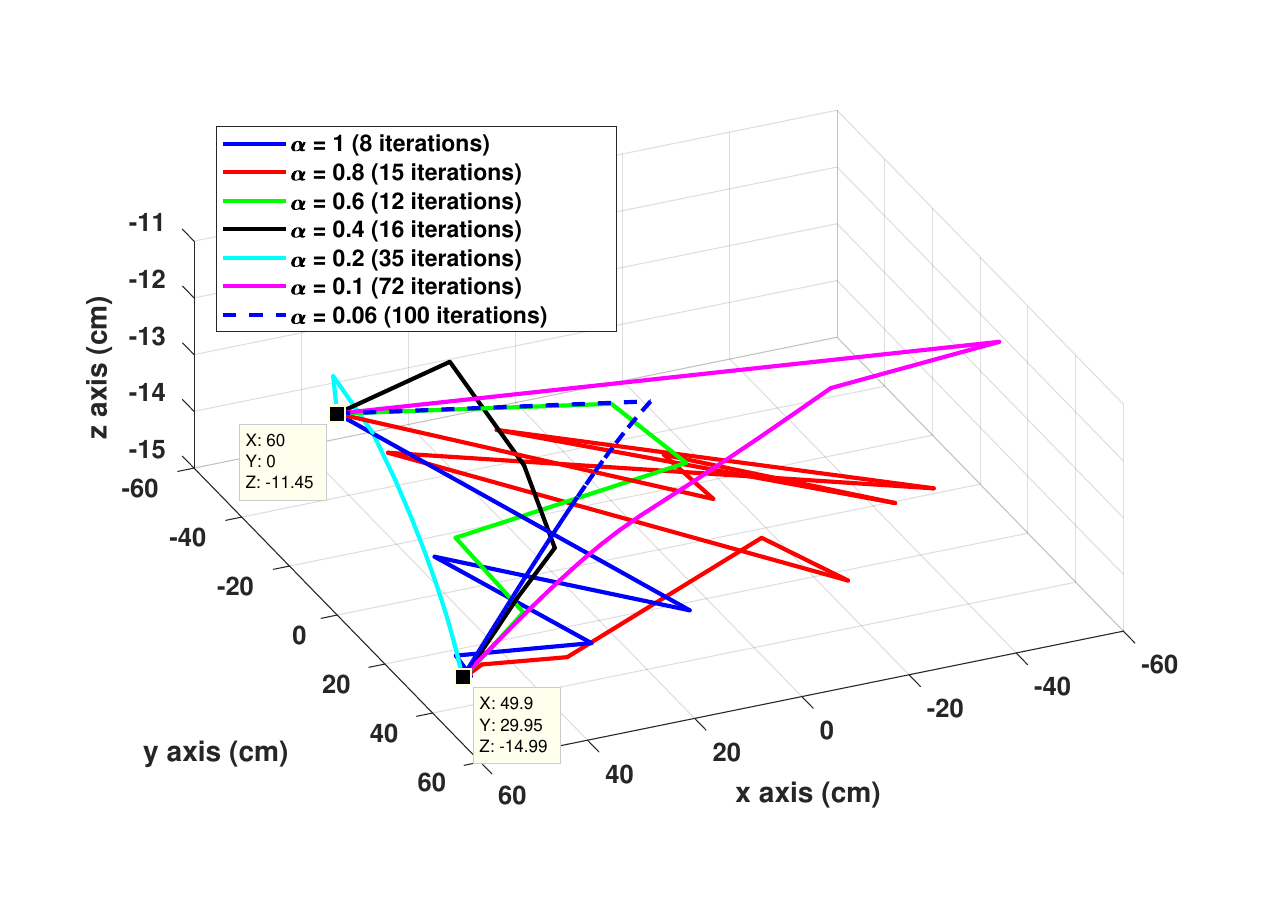}  
  \caption{\footnotesize UC / cm / multiple $\alpha$}
  \label{fig:4dof-uc-alphas-dm}
\end{subfigure}

\begin{subfigure}{.23\textwidth}
  \centering
  % include second image
  \includegraphics[width=4.6cm]{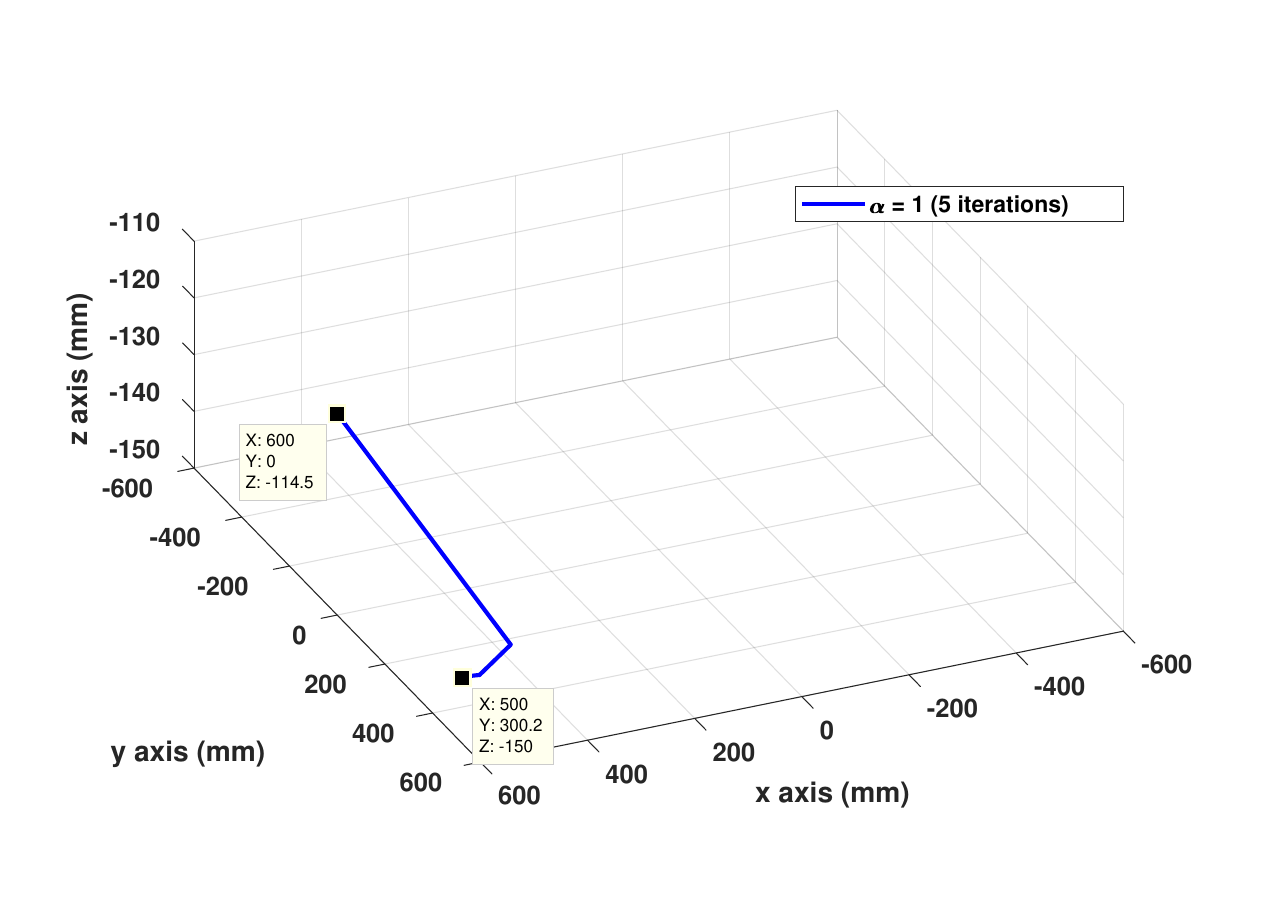}  
  \caption{\footnotesize UC / mm / $\alpha=1$}
  \label{fig:4dof-uc-alpha-mm}
\end{subfigure}
\begin{subfigure}{.23\textwidth}
  \centering
  % include fourth image
  \includegraphics[width=4.6cm]{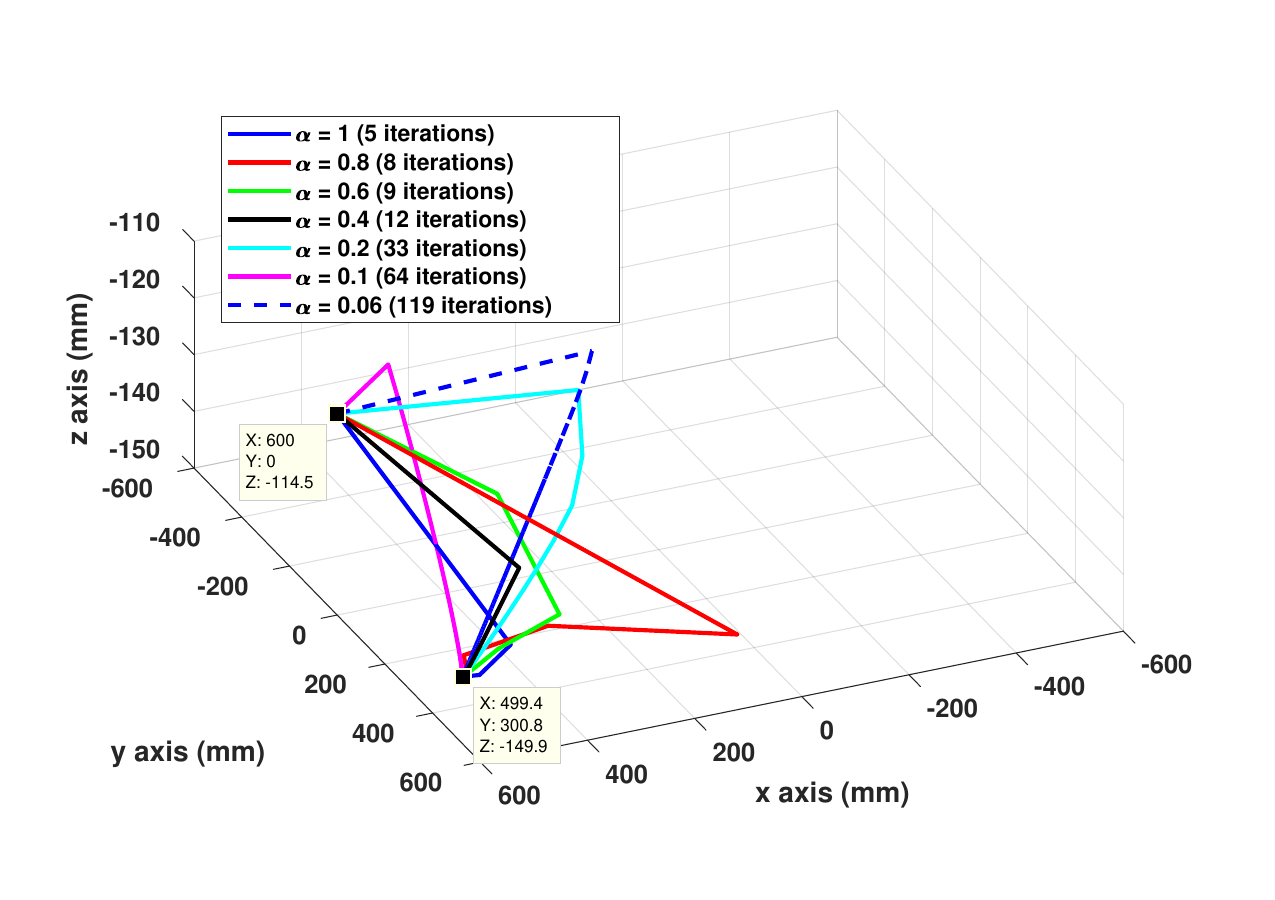}  
  \caption{\footnotesize UC / mm / multiple $\alpha$}
  \label{fig:4dof-uc-alphas-mm}
\end{subfigure}

\caption{\footnotesize Behavior of the trajectories of the end-effector of the 4DoF robot when varying the units while using the UC inverse with (\ref{fig:4dof-uc-alpha-m}), (\ref{fig:4dof-uc-alpha-dm}),(\ref{fig:4dof-uc-alpha-cm}), and (\ref{fig:4dof-uc-alpha-mm}) the attenuation parameter $\alpha = 1$ and (\ref{fig:4dof-uc-alphas-m}), (\ref{fig:4dof-uc-alphas-dm}), (\ref{fig:4dof-uc-alphas-cm}), and (\ref{fig:4dof-uc-alphas-mm}) multiple values of $\alpha$.}
\label{fig:trajectories-UC-4DoF-alpha(s)}
\vspace{-7mm}
\end{figure}

\vspace{-2.9mm}
\subsection{Cases in which the MP, UC or MX must be used}
We explored several motions for each of the incommensurate robots using the MP, UC and MX GI's; however, due to space limitations, we will only provide results for one motion in this paper.  We provide experimental results for an extensive list of motions for all manipulators \href{http://vigir.missouri.edu/~dembysj/publications/GI2023/index.html}{in our website}. \footnote{\label{note1}\href{http://vigir.missouri.edu/~dembysj/publications/GI2023/index.html}{http://vigir.missouri.edu/\~dembysj/publications/GI2023/index.html}}

\subsubsection{Case in which the MP must be used}
When the axis of movement of the joint variables are parallel to each other, our MX rule of thumb established that the rotational joints before the prismatic joint where the change of units occurs, and the prismatic joint need to be handled by the MP only. In fact, in such a configuration the change of units does not affect the end-effector path when using the MP inverse. This case is illustrated by the use of a 4DoF Scara manipulator whose configuration involves all the joints $Z-axes$ being parallel to each other.  So, for this 4DoF serial manipulator with a RRPR configuration, since the axis of translation of the linear joint is aligned (parallel) to the axes of rotation of the revolute joints, the linear joint is not affected by these rotations, and hence it does not need to be included in the $A_{W}$ block and any change of units will be adequately handled by the MP inverse. So, given the Jacobian $J$ for this 4DoF robot:

\begin{equation} \label{eq10}
{\small{}J= \begin{bmatrix}
A_W & A_X\\
A_Y & A_Z
\end{bmatrix}
= \begin{bmatrix}
\frac{\partial X}{\partial \theta_1} & \frac{\partial X}{\partial \theta_2} & \frac{\partial X}{\partial d_3} & \frac{\partial X}{\partial \theta_4}\\
\frac{\partial Y}{\partial \theta_1} & \frac{\partial Y}{\partial \theta_2} & \frac{\partial Y}{\partial d_3} & \frac{\partial Y}{\partial \theta_4}\\
\frac{\partial Z}{\partial \theta_1} & \frac{\partial Z}{\partial \theta_2} & \frac{\partial Z}{\partial d_3} & \frac{\partial Z}{\partial \theta_4}\\
\frac{\partial Ro}{\partial \theta_1} & \frac{\partial Ro}{\partial \theta_2} & \frac{\partial Ro}{\partial d_3} & \frac{\partial Ro}{\partial \theta_4}\\
\frac{\partial Pi}{\partial \theta_1} & \frac{\partial Pi}{\partial \theta_2} & \frac{\partial Pi}{\partial d_3} & \frac{\partial Pi}{\partial \theta_4}\\
\frac{\partial Ya}{\partial \theta_1} & \frac{\partial Ya}{\partial \theta_2} & \frac{\partial Ya}{\partial d_3} & \frac{\partial Ya}{\partial \theta_4}
\end{bmatrix}{\small}}
\end{equation}

\noindent and the fact that all the variables should be in the bottom-right block $A_Z$, so it is handled by the MP inverse, the block partitioning for this robot becomes: $A_W = [0]$  a 1x1 matrix of zeros; $A_X = [0 \;\; 0 \;\;  0 \;\;  0]$ a 1x4 matrix of zeros; $A_Y = [0 \;\; 0 \;\; 0 \;\; 0 \;\; 0 \;\; 0 \;\; 0]^{T}$ a 6x1 matrix of zeros; and $A_Z = J$  the entire 6x4 original Jacobian matrix in \ref{eq7}. Then, the MX inverse $J^{-M}$ can be computed by $J^{-M}= \begin{bmatrix}
0 & 0\\
0 & A_Z^{-P}
\end{bmatrix}$. This final $J^{-M}$ inverse Jacobian matrix is a 5x7 matrix with one row and one column of zeros. 
Even though, our ultimate goal is to always apply the MX inverse; we still provide the results obtained from the application of the MP and UC inverses. Figure \ref{fig:trajectories-MP-4DoF-alpha(s)} shows the paths of the end-effector when the units of the linear joints in the 4DoF robot are varied from $m$ to $mm$, still for the same motion. Because, the axes of rotation of the revolute joints are parallel to the axis of translation of the linear joint, the behavior of the end-effector is the same when the units are varied as presented in Sub-Figures \ref{fig:4dof-mp-alpha-m}, \ref{fig:4dof-mp-alpha-dm}, \ref{fig:4dof-mp-alpha-cm}, and \ref{fig:4dof-mp-alpha-mm} where the attenuation factor $\alpha = 1$. We can observe that, changes of units do not affect the paths of the end-effector and the system is handled consistently by simply using the MP inverse. Next, we studied the effects of the attenuation parameter $\alpha$ on the path followed by the end-effector under different units and still with the MP inverse. Sub-Figures \ref{fig:4dof-mp-alphas-m}, \ref{fig:4dof-mp-alphas-dm}, \ref{fig:4dof-mp-alphas-cm}, and \ref{fig:4dof-mp-alphas-mm} show these results, and as it can be observed, while $\alpha$ can be effectively used to make the path smoother, the paths followed by each choice of $\alpha$ are exactly the same despite the change of units.

We also applied the UC inverse to the same motion of the 4DoF robot. As before, Figure \ref{fig:trajectories-UC-4DoF-alpha(s)} shows the paths of the end-effector when the units are varied from $m$ to $mm$. Not surprisingly, the behavior of the robot is quite different and unpredictable when the units are varied as presented in Sub-Figures \ref{fig:4dof-uc-alpha-m}, \ref{fig:4dof-uc-alpha-dm}, \ref{fig:4dof-uc-alpha-cm}, and \ref{fig:4dof-uc-alpha-mm} for an attenuation factor $\alpha = 1$. That is, the UC inverse is mishandling variables that are not affected by the change of units of the linear joint. As it can be seen, a simple change of units causes the robot to follow quite different paths. We also studied the effects of the attenuation parameter $\alpha$ on the path followed by the end-effector under different units and still using the UC inverse. Sub-Figures \ref{fig:4dof-uc-alphas-m}, \ref{fig:4dof-uc-alphas-dm}, \ref{fig:4dof-uc-alphas-cm}, and \ref{fig:4dof-uc-alphas-mm} show these results. Now, also as expected, the attenuation parameter cannot correct the negative effect on the path followed by the robot since the the UC is unable to provide consistency when the units are changed.

The last GI used was the MX inverse with the same motion of the 4DoF robot. As we explained above, in this case, the MX inverse reduces to the MP inverse. Indeed, we observe the exact same results obtained for the MP and presented in Figure \ref{fig:trajectories-MP-4DoF-alpha(s)} -- and the reader can check this fact in the corresponding figures \href{http://vigir.missouri.edu/~dembysj/publications/GI2023/index.html}{in our website}.

Our website also presents a complete table that summarizes the overall performance of the proposed IK solver for $\alpha = 1$, by presenting the input parameters provided to the algorithm; the number of iterations; and final end-effector pose error with respect to the desired pose for the reported 4DoF motion while using all the three GI's MP, UC and MX along with all the units from $m$, to $dm$, to $cm$, and finally to $mm$.

\subsubsection{Case in which the UC must be used}
When the axis of the joint variables are not parallel to each other, the MX rule of thumb establishes that the rotational joints before the prismatic joint where the change of units occurs and the same prismatic joint need to be handled by the UC inverse as this will provide unit consistency. The remaining variables must be handled by the MP inverse. This case is illustrated by the use of a 3DoF Planar manipulator with a RRP configuration and the $Z-axes$ of the prismatic joint not being parallel to the others -- an inspection of the two early rotations will show that all variables are affected by them. So, given the Jacobian for this 3DoF robot:

\begin{equation} \label{eq12}
{\small{} J= \begin{bmatrix}
A_W & A_X\\
A_Y & A_Z
\end{bmatrix}
= \begin{bmatrix}
\frac{\partial X}{\partial \theta_1} & \frac{\partial X}{\partial \theta_2} & \frac{\partial X}{\partial d_3}\\
\frac{\partial Y}{\partial \theta_1} & \frac{\partial Y}{\partial \theta_2} & \frac{\partial Y}{\partial d_3}
\end{bmatrix}{\small}}
\end{equation}

\noindent and the fact that all the variables should be in the top-left block $A_W$, so it is handled by the UC inverse, the block partitioning for $J$ becomes: $A_W = J$  the entire 2x3 matrix; $A_X = [0 \;\; 0]^T$ is a 2x1 matrix of zeros, $A_Y = [0 \;\; 0 \;\; 0]$ is 1x3 matrix of zeros and $A_Z = [0]$ is a 1x1 matrix of zeros. Then, the MX inverse $J^{-M}$ reduces to a simple UC inverse as showed by $J^{-M}= \begin{bmatrix}
A_W^{-U} & 0\\
0 & 0
\end{bmatrix}$. The resulting $J^{-M}$ inverse Jacobian matrix is a 4x3 matrix with one row and one column of zeros. Once again, we provide the results obtained from applying the MP and UC inverses, before applying the MX inverse. 

%%%%%%%%%%%%%%%%%%%%%%%%%%%%%%%%%%%%%%%%%%%%%%%%%%%%%%%%%%%%%%%%%%%%%%%%%%%%%
%%%%%%%%%%%%%%%%% 3DoF
%%%%%%%%%%%%%%%%%%%%%%%%%%%%%%%%%%%%%%%%%%%%%%%%%%%%%%%%%%%%%%%%%%%%%%%%%%%%%

\begin{figure}[t!] %[htb!]
\centering

\begin{subfigure}{.23\textwidth}
  \centering
  % include first image
  \includegraphics[width=4.6cm]{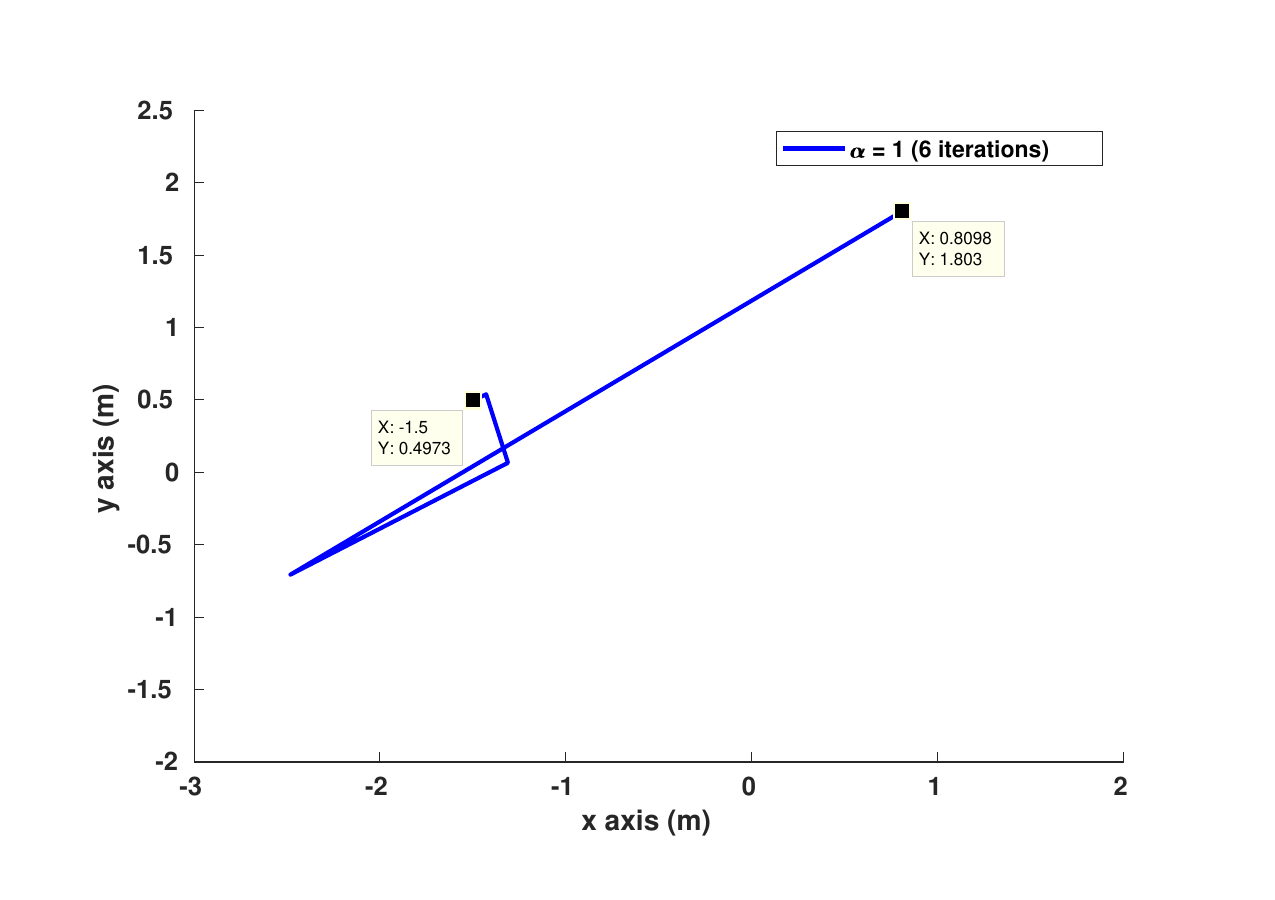}  
  \caption{\footnotesize MP / m / $\alpha=1$}
  \label{fig:3dof-mp-alpha-m}
\end{subfigure}
\begin{subfigure}{.23\textwidth}
  \centering
  % include third image
  \includegraphics[width=4.6cm]{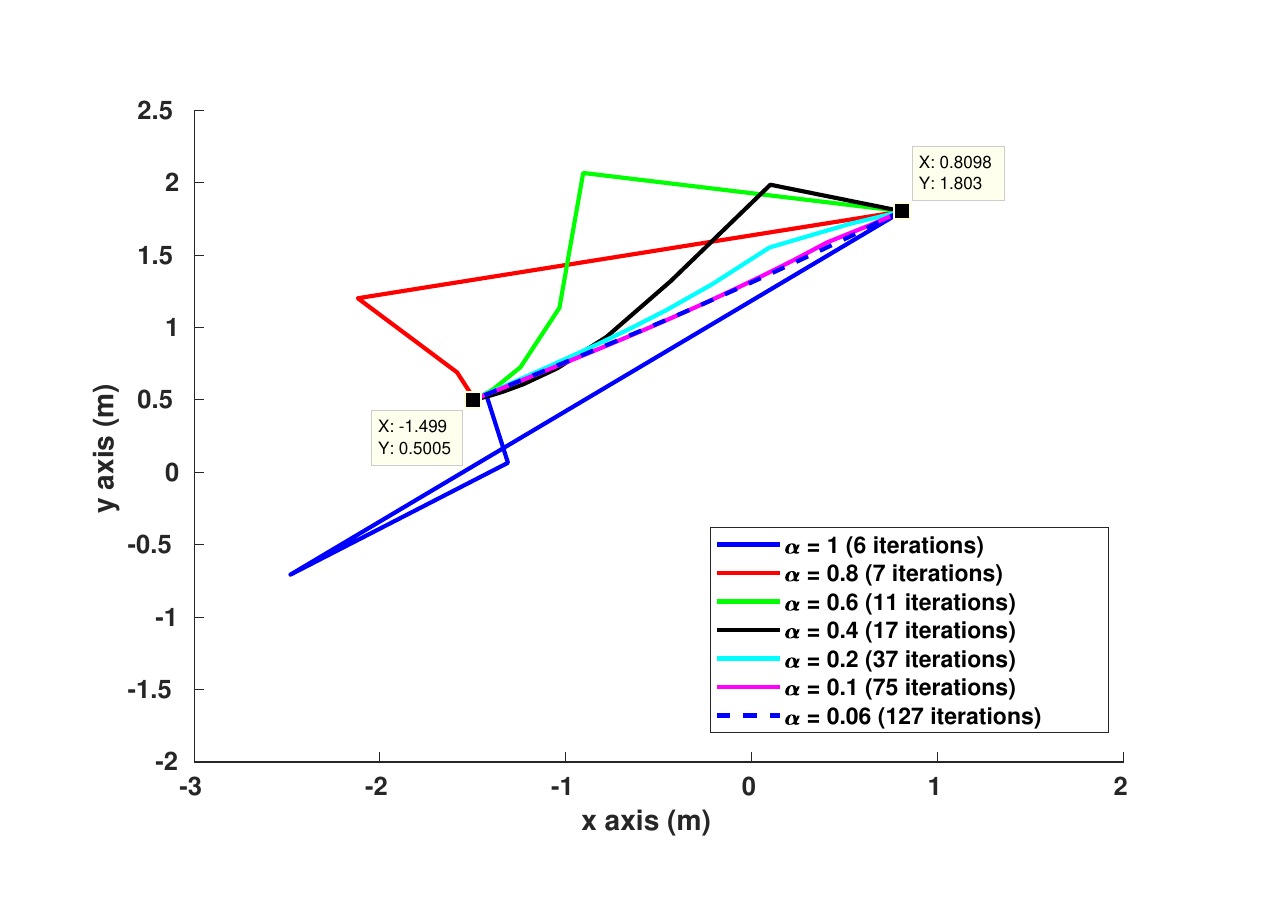}  
  \caption{\footnotesize MP / m / multiple $\alpha$}
  \label{fig:3dof-mp-alphas-m}
\end{subfigure}

\begin{subfigure}{.23\textwidth}
  \centering
  % include second image
  \includegraphics[width=4.6cm]{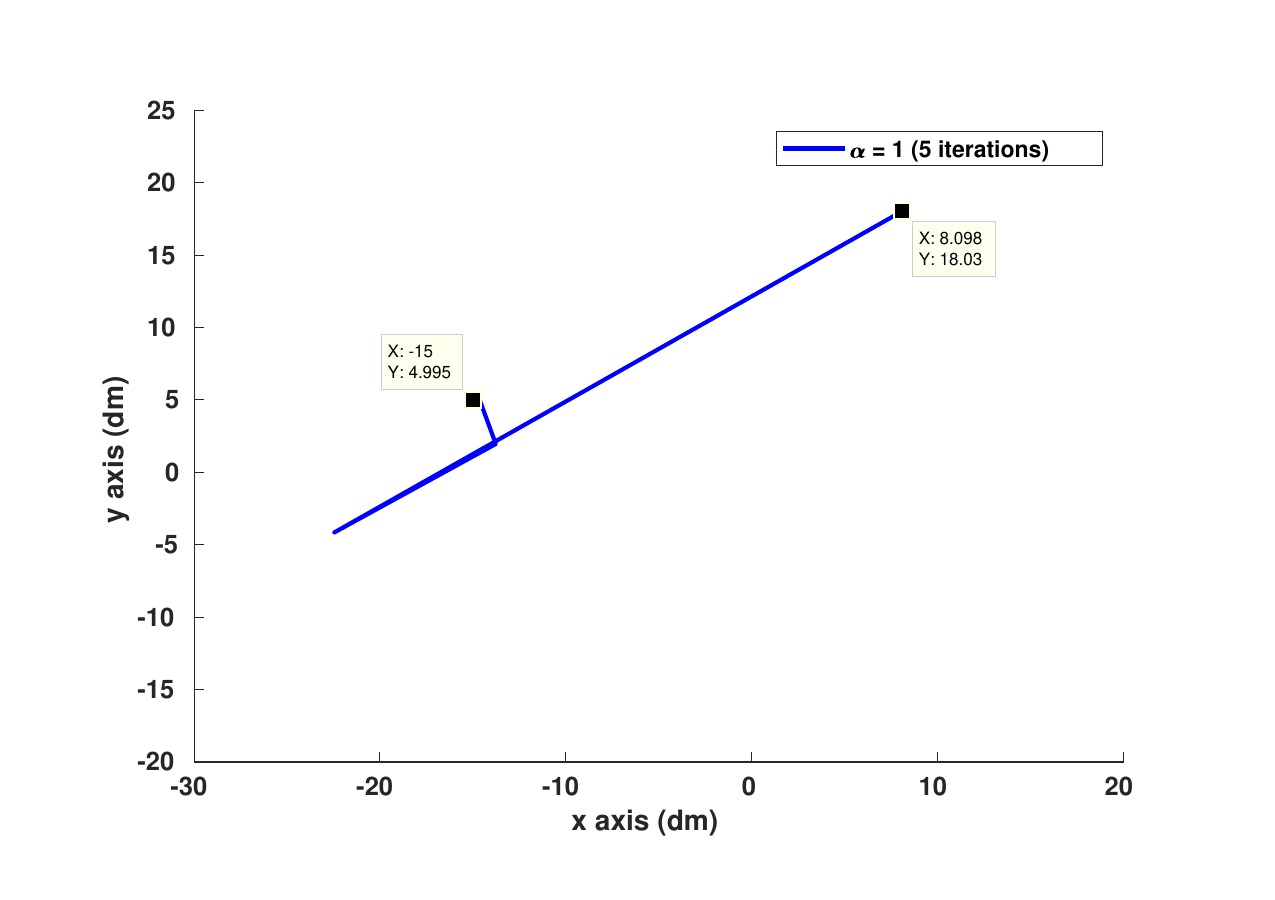}  
  \caption{\footnotesize MP / dm / $\alpha=1$}
  \label{fig:3dof-mp-alpha-dm}
\end{subfigure}
\begin{subfigure}{.23\textwidth}
  \centering
  % include fourth image
  \includegraphics[width=4.6cm]{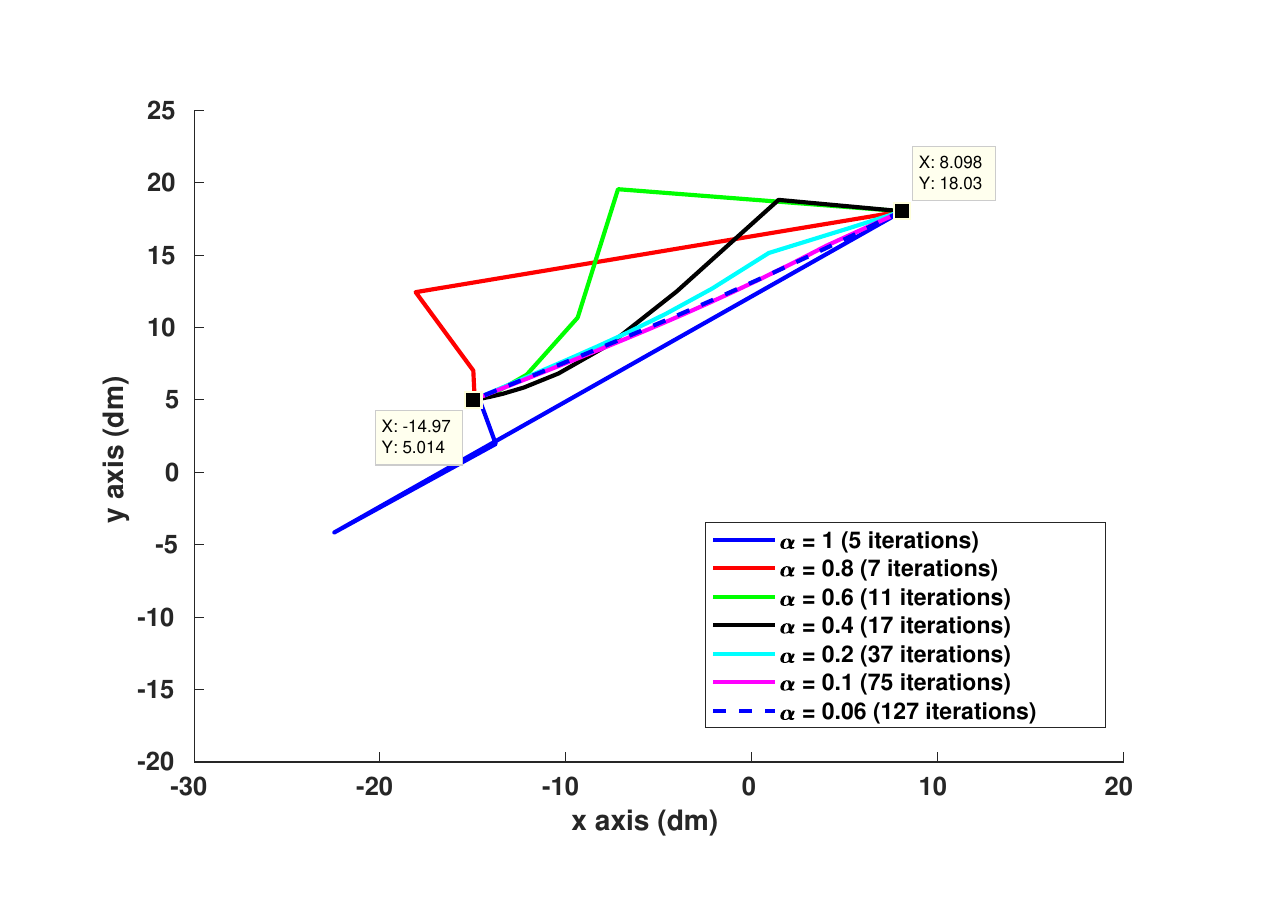}  
  \caption{\footnotesize MP / dm / multiple $\alpha$}
  \label{fig:3dof-mp-alphas-dm}
\end{subfigure}

\begin{subfigure}{.23\textwidth}
  \centering
  % include second image
  \includegraphics[width=4.6cm]{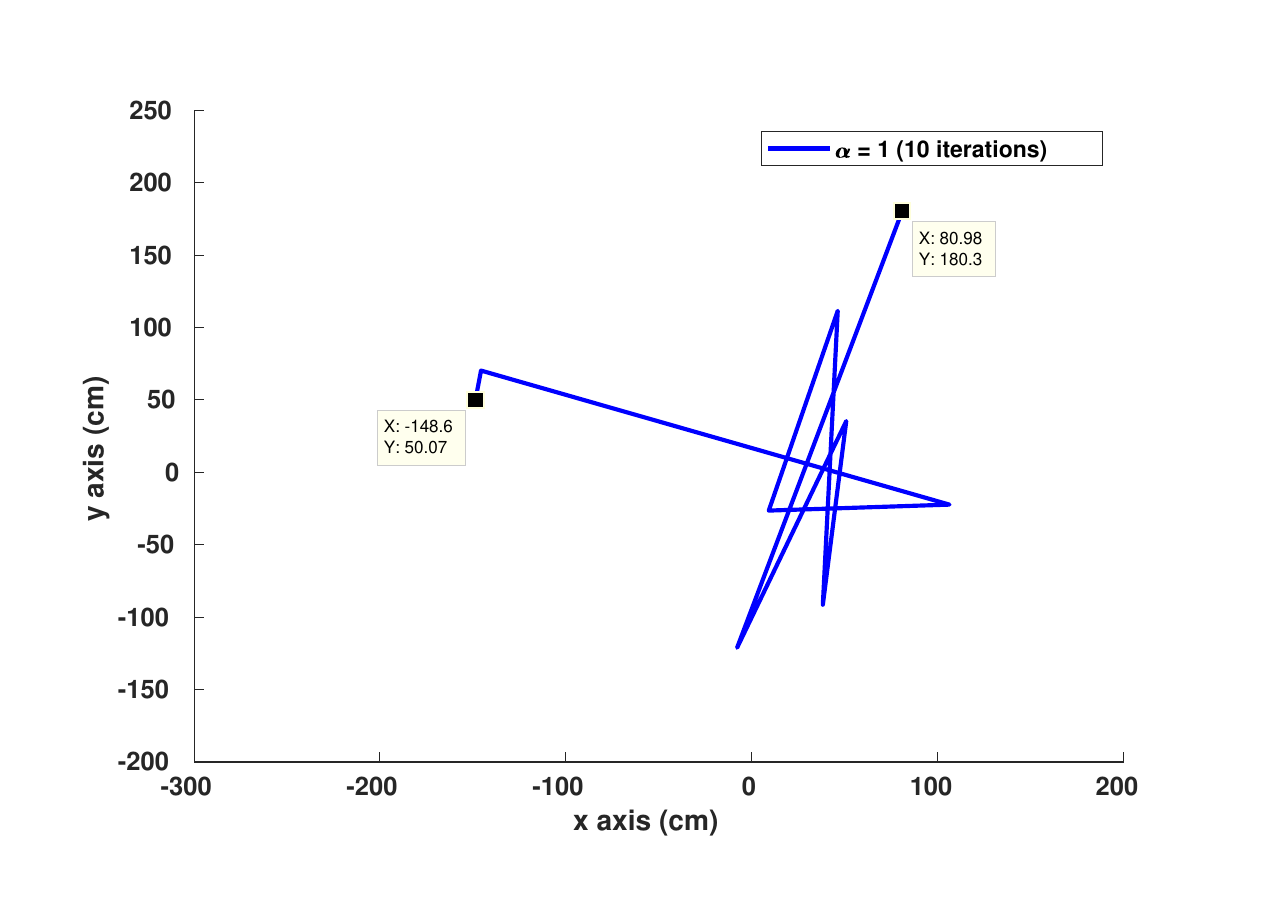}  
  \caption{\footnotesize MP / cm / $\alpha=1$}
  \label{fig:3dof-mp-alpha-cm}
\end{subfigure}
\begin{subfigure}{.23\textwidth}
  \centering
  % include fourth image
  \includegraphics[width=4.6cm]{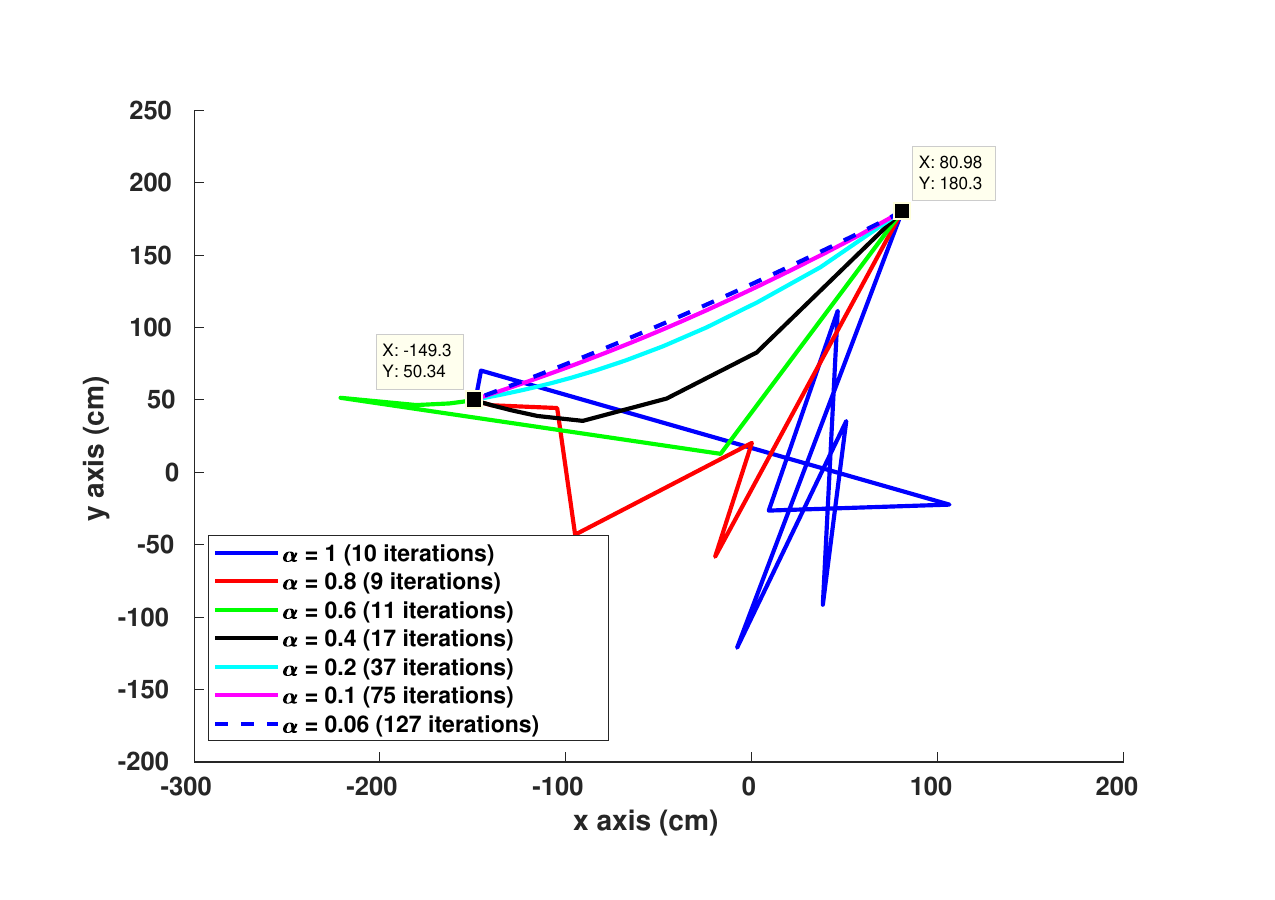}  
  \caption{\footnotesize MP / cm / multiple $\alpha$}
  \label{fig:3dof-mp-alphas-cm}
\end{subfigure}

\begin{subfigure}{.23\textwidth}
  \centering
  % include second image
  \includegraphics[width=4.6cm]{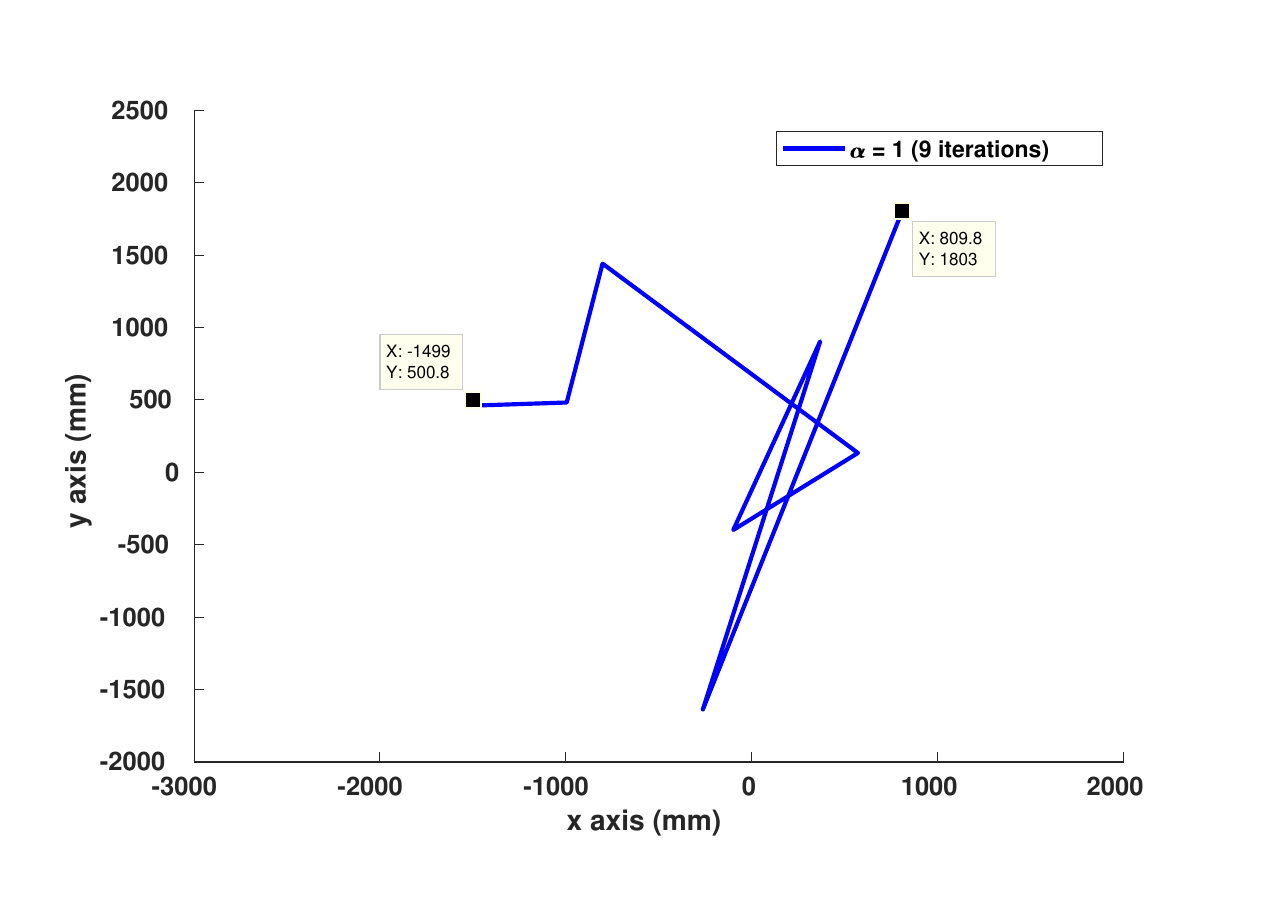}  
  \caption{\footnotesize MP / mm / $\alpha=1$}
  \label{fig:3dof-mp-alpha-mm}
\end{subfigure}
\begin{subfigure}{.23\textwidth}
  \centering
  % include fourth image
  \includegraphics[width=4.6cm]{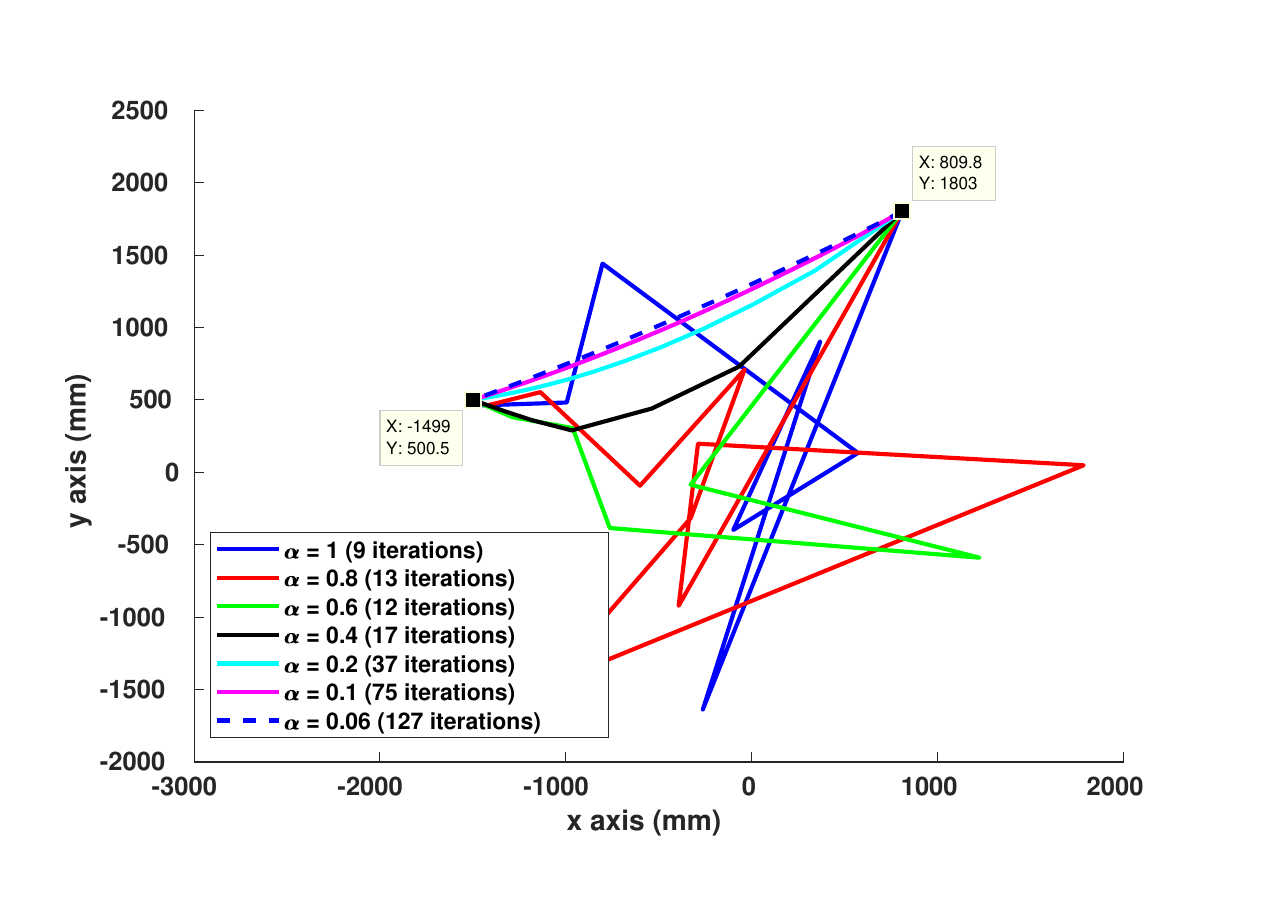}  
  \caption{\footnotesize MP / mm / multiple $\alpha$}
  \label{fig:3dof-mp-alphas-mm}
\end{subfigure}

\caption{\footnotesize Behavior of the trajectories of the end-effector of the 3DoF robot when varying the units while using the MP Inverse with (\ref{fig:3dof-mp-alpha-m}), (\ref{fig:3dof-mp-alpha-dm}), (\ref{fig:3dof-mp-alpha-cm}), and (\ref{fig:3dof-mp-alpha-mm}) the attenuation parameter $\alpha = 1$ and (\ref{fig:3dof-mp-alphas-m}), (\ref{fig:3dof-mp-alphas-dm}), (\ref{fig:3dof-mp-alphas-cm}), and (\ref{fig:3dof-mp-alphas-mm}) multiple values of $\alpha$.}
\label{fig:trajectories-MP-3DoF-alpha(s)}
\vspace{-7mm}
\end{figure}

\begin{figure}[t!]  %[htb!]
\centering
%%%%%%%%%%%%%%%%% UC Inverse - 3DoF alpha = 1
\begin{subfigure}{.23\textwidth}
  \centering
  % include first image
  \includegraphics[width=4.6cm]{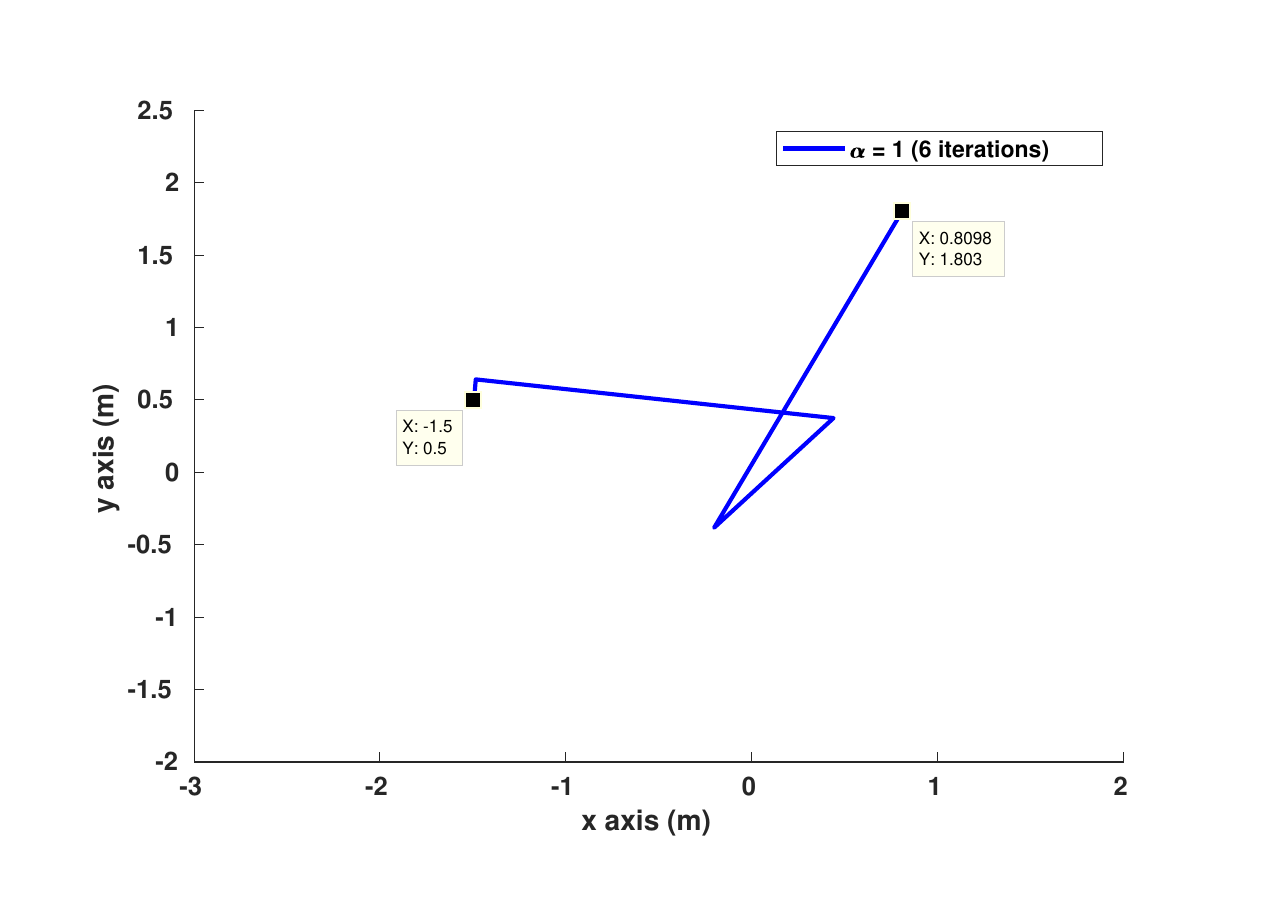}  
  \caption{\footnotesize UC / m / $\alpha=1$}
  \label{fig:3dof-uc-alpha-m}
\end{subfigure}
\begin{subfigure}{.23\textwidth}
  \centering
  % include third image
  \includegraphics[width=4.6cm]{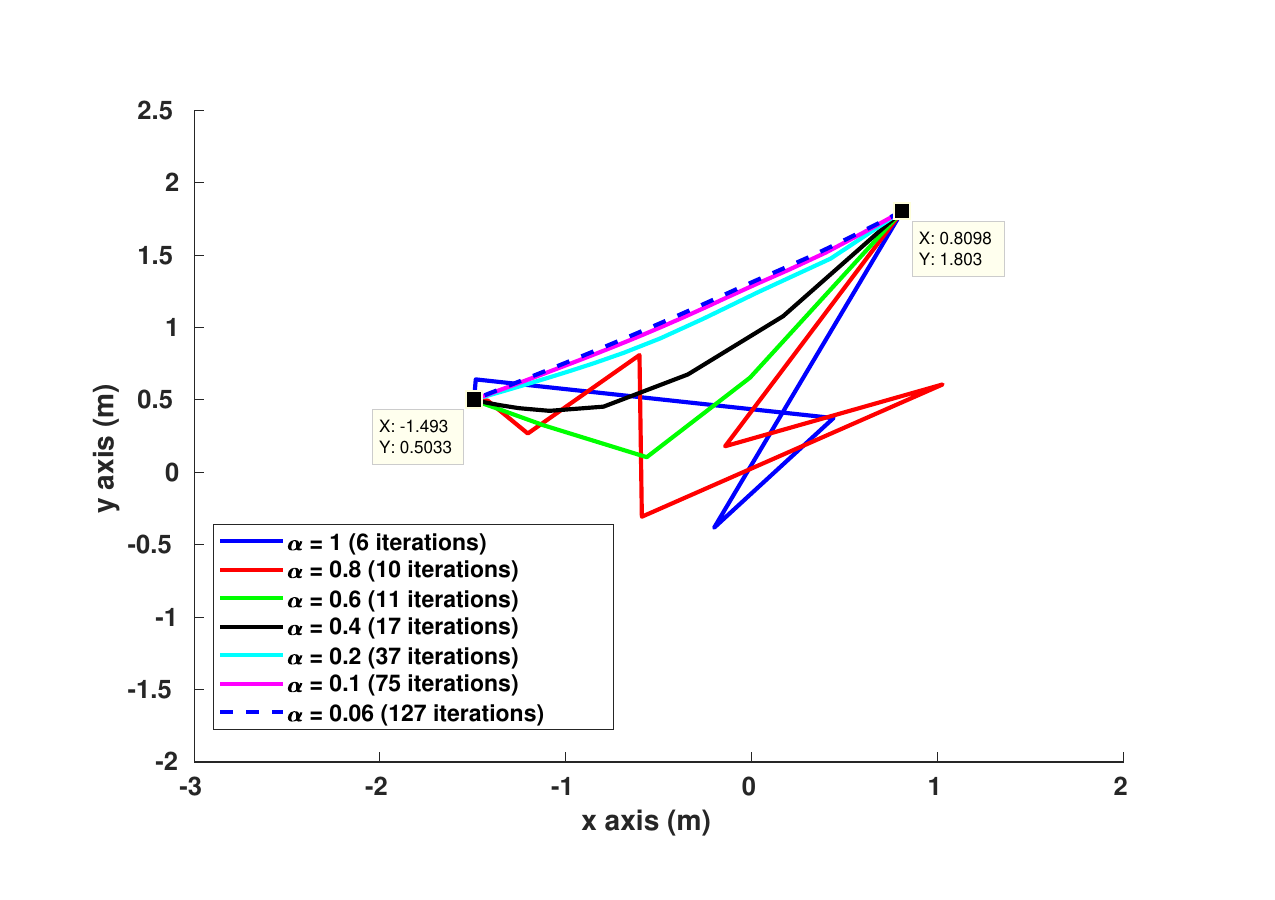}  
  \caption{\footnotesize UC / m / multiple $\alpha$}
  \label{fig:3dof-uc-alphas-m}
\end{subfigure}

\begin{subfigure}{.23\textwidth}
  \centering
  % include second image
  \includegraphics[width=4.6cm]{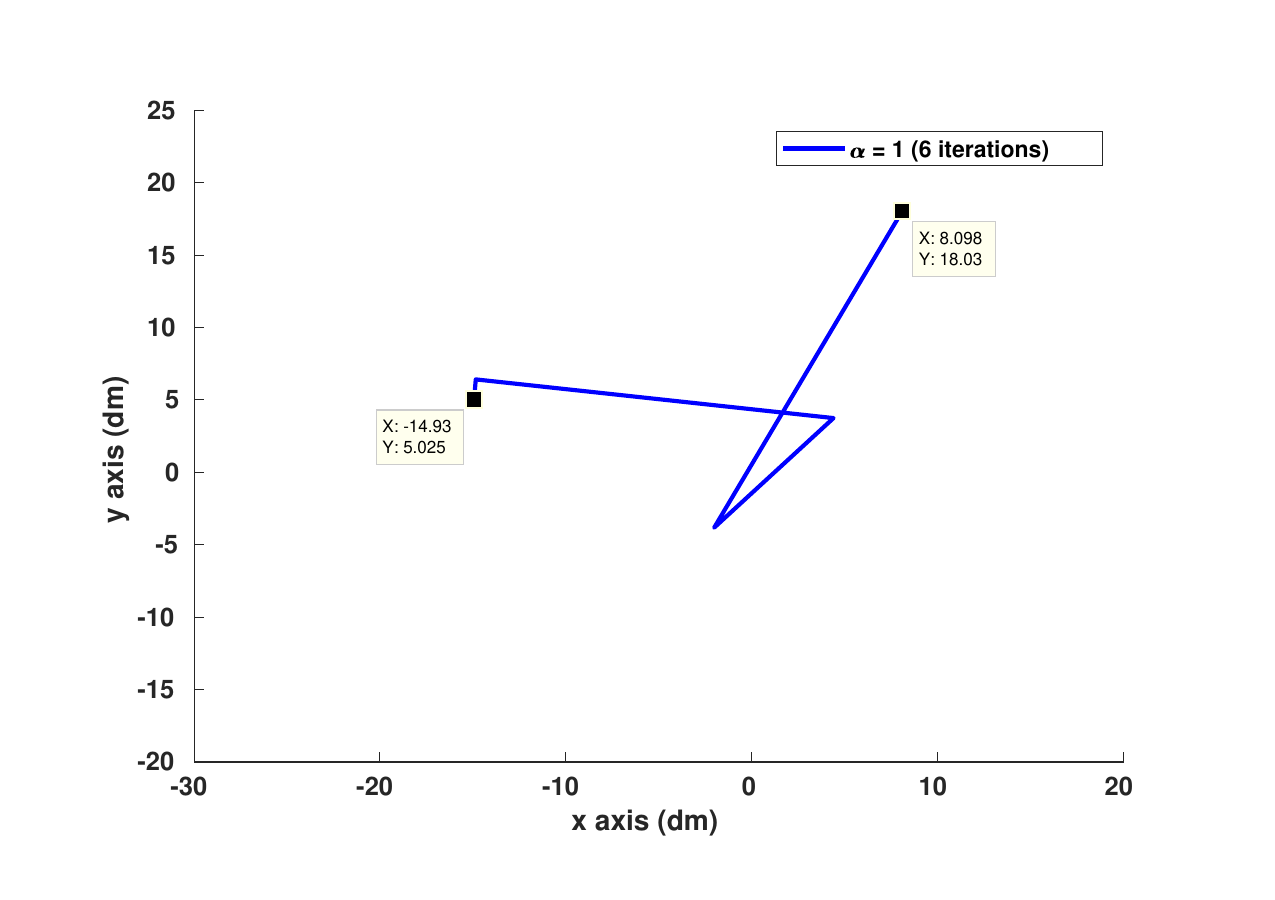}  
  \caption{\footnotesize UC / dm / $\alpha=1$}
  \label{fig:3dof-uc-alpha-dm}
\end{subfigure}
\begin{subfigure}{.23\textwidth}
  \centering
  % include fourth image
  \includegraphics[width=4.6cm]{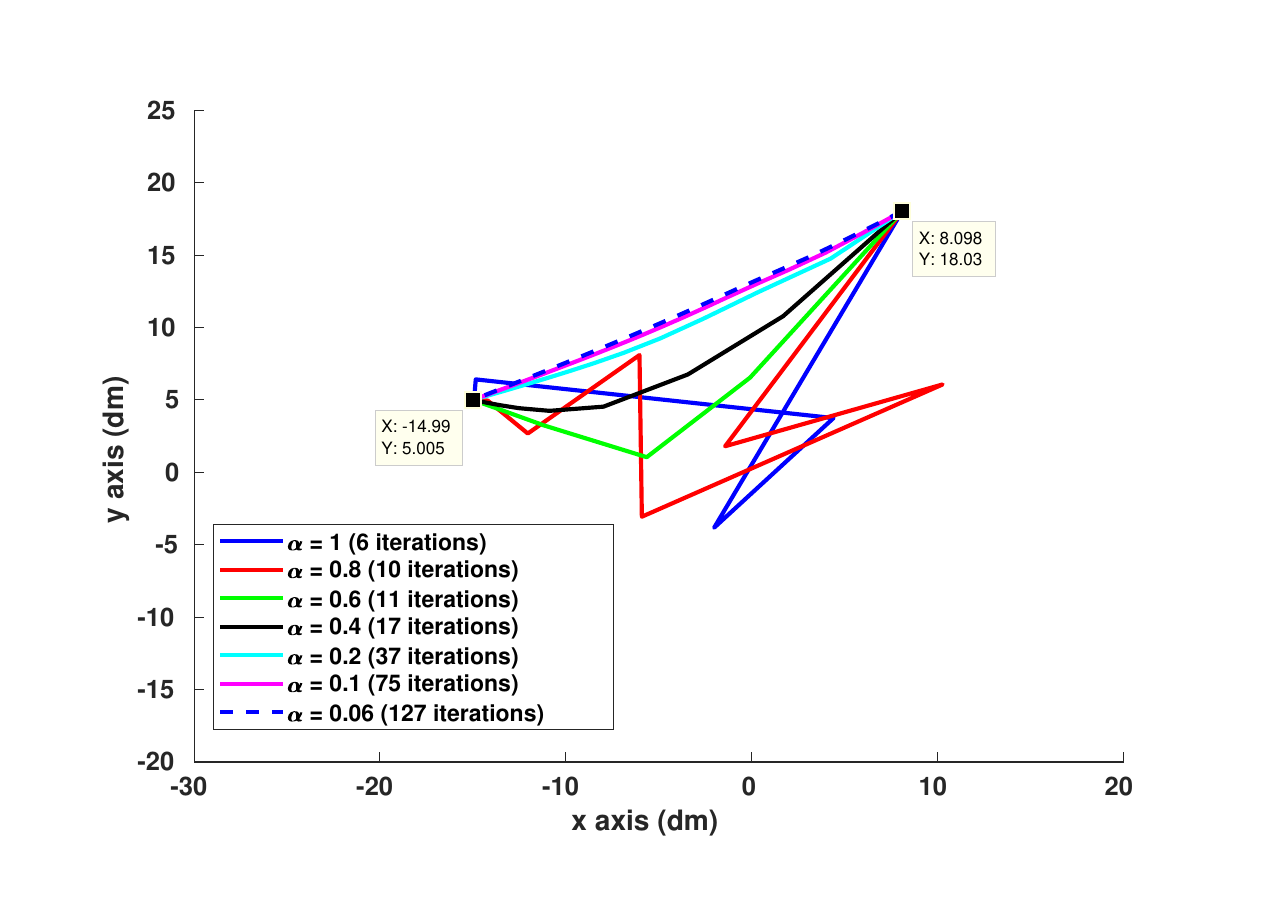}  
  \caption{\footnotesize UC / dm / multiple $\alpha$}
  \label{fig:3dof-uc-alphas-dm}
\end{subfigure}

\begin{subfigure}{.23\textwidth}
  \centering
  % include second image
  \includegraphics[width=4.6cm]{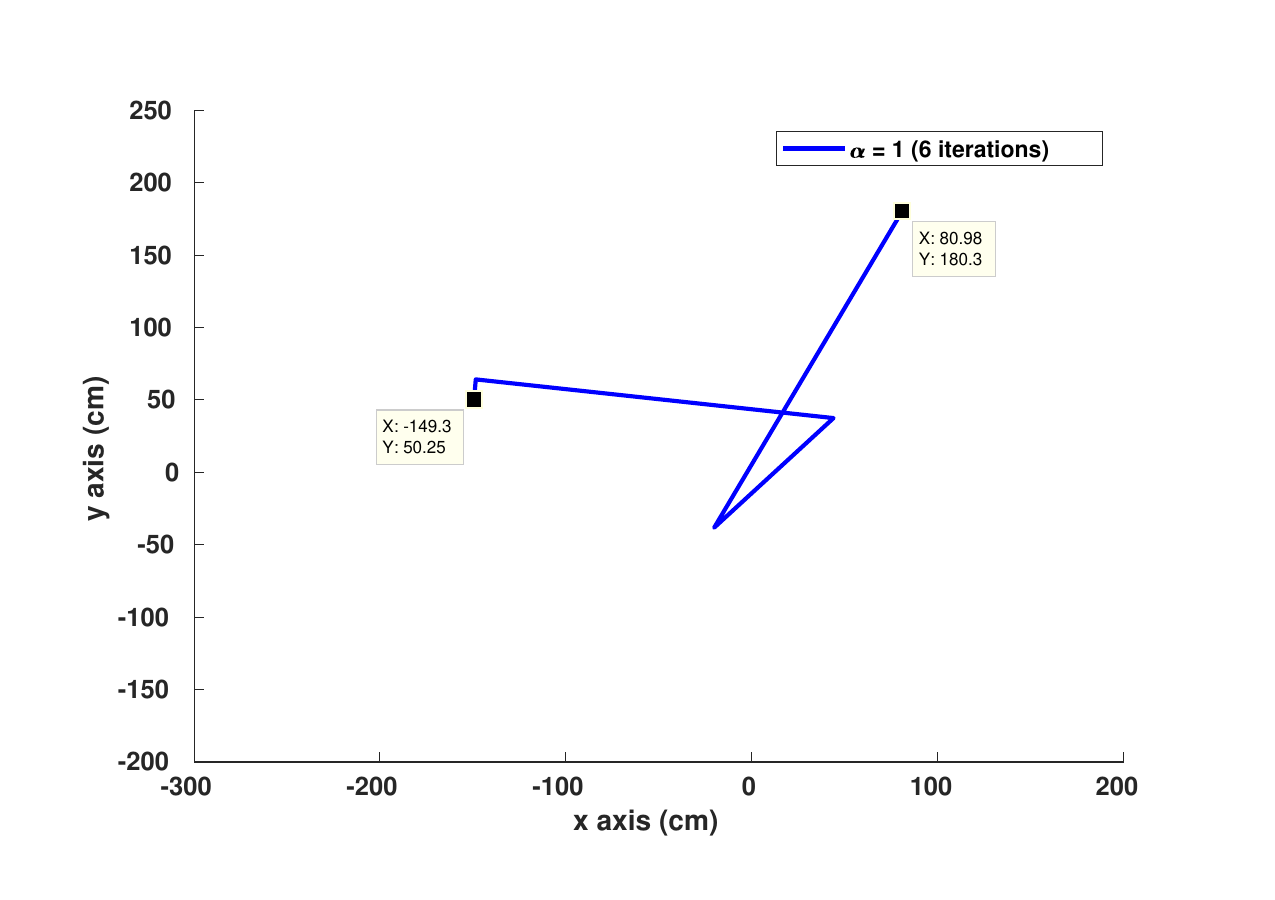}  
  \caption{\footnotesize UC / cm / $\alpha=1$}
  \label{fig:3dof-uc-alpha-cm}
\end{subfigure}
\begin{subfigure}{.23\textwidth}
  \centering
  % include fourth image
  \includegraphics[width=4.6cm]{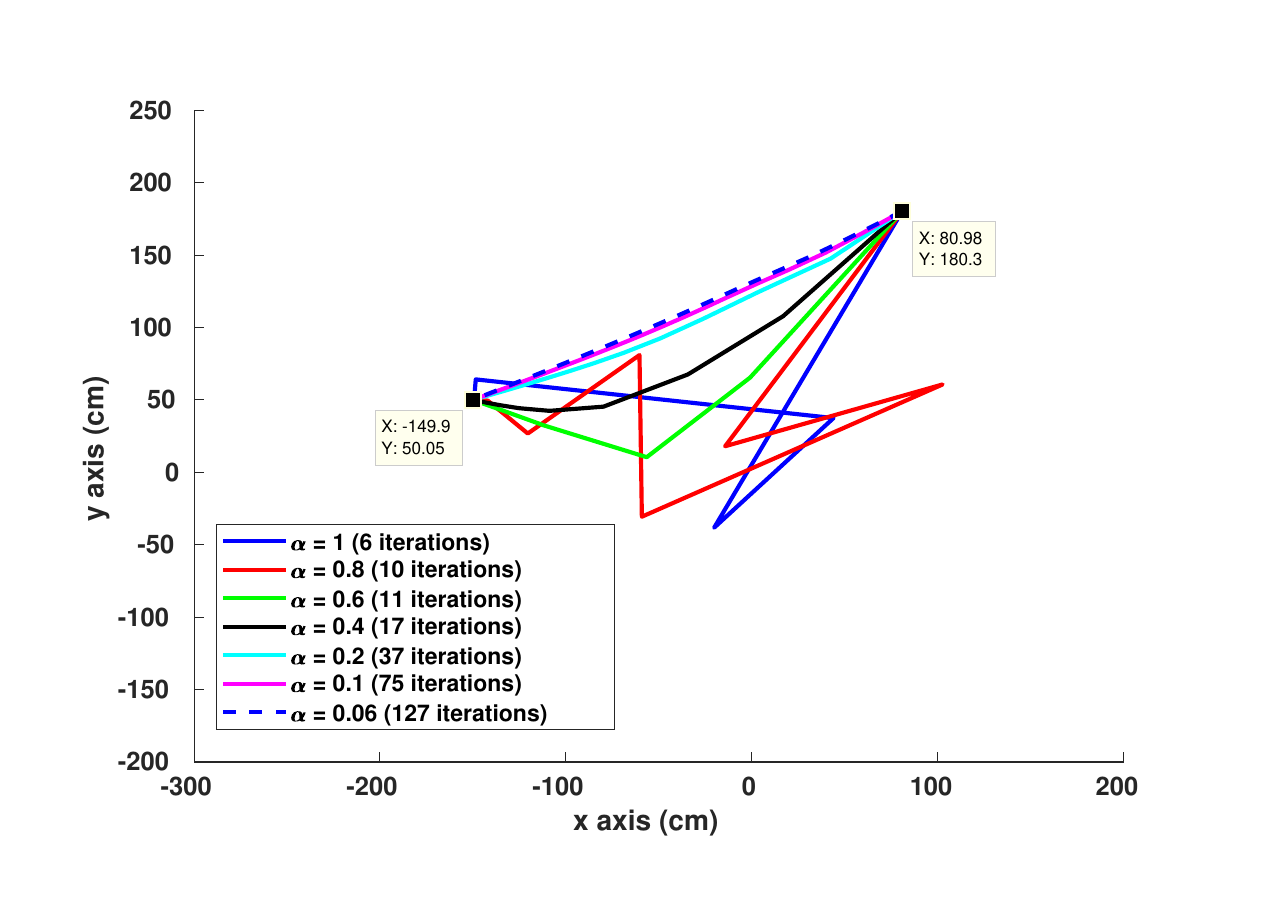}  
  \caption{\footnotesize UC / cm / multiple $\alpha$}
  \label{fig:3dof-uc-alphas-cm}
\end{subfigure}

\begin{subfigure}{.23\textwidth}
  \centering
  % include second image
  \includegraphics[width=4.6cm]{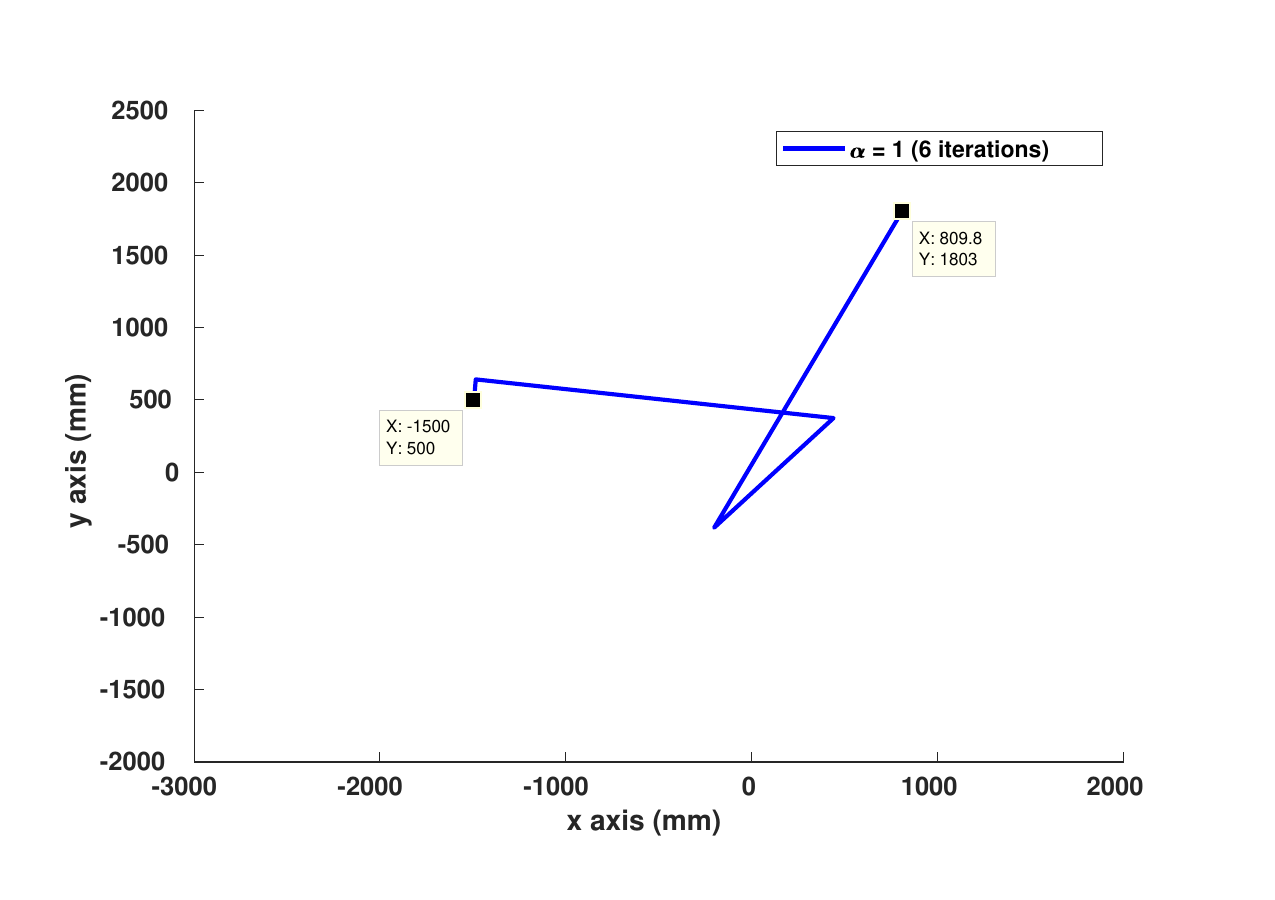}  
  \caption{\footnotesize UC / mm / $\alpha=1$}
  \label{fig:3dof-uc-alpha-mm}
\end{subfigure}
\begin{subfigure}{.23\textwidth}
  \centering
  % include fourth image
  \includegraphics[width=4.6cm]{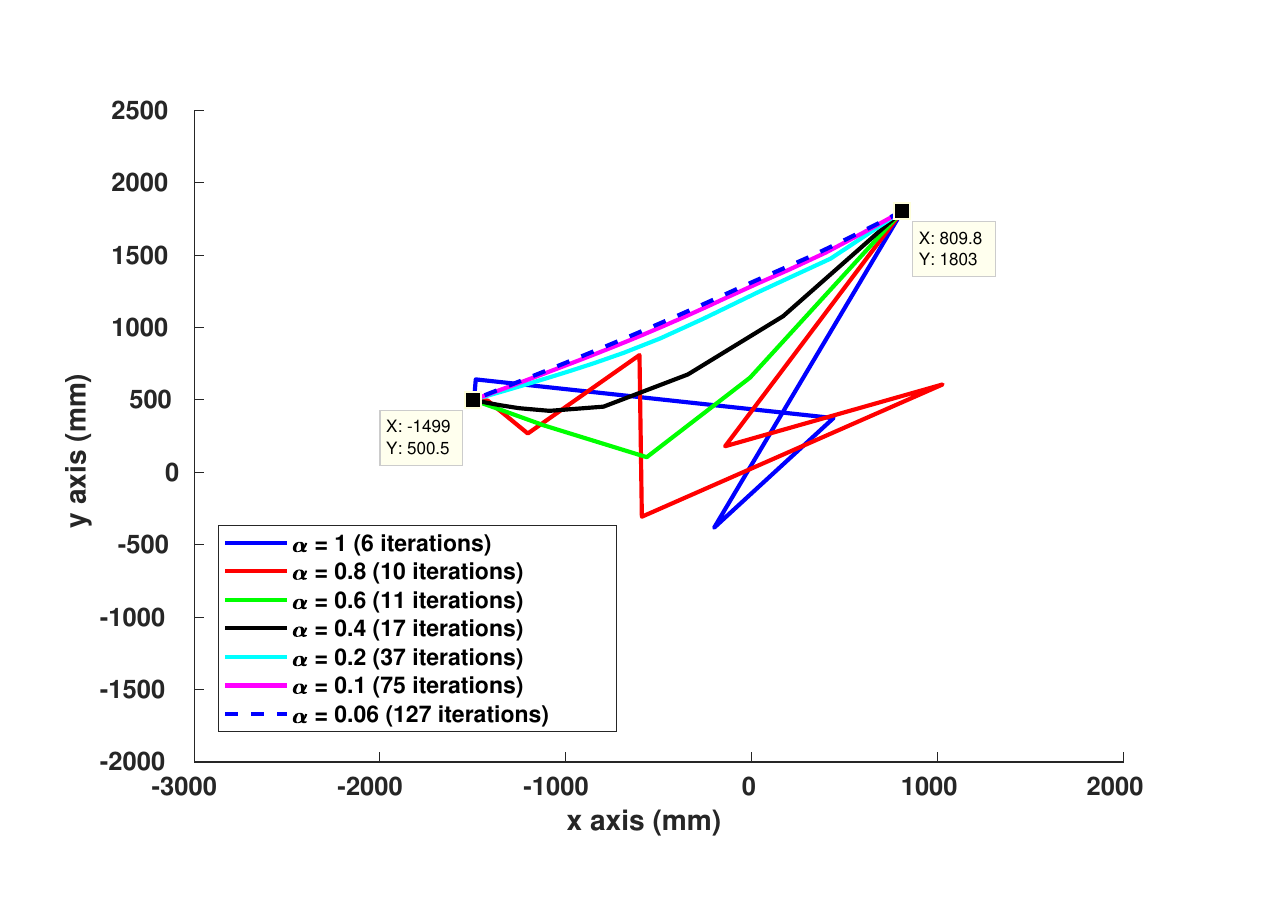}  
  \caption{\footnotesize UC / mm / multiple $\alpha$}
  \label{fig:3dof-uc-alphas-mm}
\end{subfigure}

\caption{\footnotesize Behavior of the trajectories of the end-effector of the 3DoF robot when varying the units while using the UC inverse with (\ref{fig:3dof-uc-alpha-m}), (\ref{fig:3dof-uc-alpha-dm}), (\ref{fig:3dof-uc-alpha-cm}), and (\ref{fig:3dof-uc-alpha-mm}) the attenuation parameter $\alpha = 1$ and (\ref{fig:3dof-uc-alphas-m}), (\ref{fig:3dof-uc-alphas-dm}), (\ref{fig:3dof-uc-alphas-cm}), and (\ref{fig:3dof-uc-alphas-mm})  multiple values of $\alpha$.}
\label{fig:trajectories-UC-3DoF-alpha(s)}
\vspace{-7mm}
\end{figure}

For the MP inverse, the results, presented in Figure \ref{fig:trajectories-MP-3DoF-alpha(s)} shows that the behavior (path) of the robot is quite different and unpredictable when the unit of the linear joints in the 3DoF are varied from $m$, to $mm$. That is, since the MP inverse does not guarantee unit consistency in this case, a simple change of units causes the robot to follow quite different paths for an attenuation parameter $\alpha = 1$. Also, when studying the effects of $\alpha$ on the path followed by the end-effector under different units and still with the MP inverse, the results show that the attenuation parameter $\alpha$ cannot remedy the effects of the unit changes. In other words, the simple use of a GI that does not guarantee unit consistency can cause from mild differences to severe oscillations in the trajectory of this particular robot end-effector.

As before, we also applied the UC inverse to this same motion of the 3DoF robot. As expected, in Figure \ref{fig:trajectories-UC-3DoF-alpha(s)}, the behavior of the robot is now exactly the same, no matter what units are used. That is, since the UC inverse do guarantee unit consistency, no matter what units one chooses ($m$, $dm$, $cm$, or $mm$), the robot still follows the exact same path. In this case, the attenuation parameter $\alpha$  can also reliably control the path followed by the end-effector under different units -- i.e. it can help smooth the path and it has no negative effect on the path followed by the robot.

Finally, the MX inverse was applied to the same motion of the 3DoF robot. As we explained at the beginning of this section, the MX inverse reduces to the UC inverse. So, we expected and we did observe the exact same results as for the UC inverse presented in Figure \ref{fig:trajectories-UC-3DoF-alpha(s)}. Once again, the reader can check this fact, as well as obtain more information about the input parameters provided to the algorithm; the number of iterations; and the final error in the position of the end-effector with respect to the desired position in the corresponding figures and tables  \href{http://vigir.missouri.edu/~dembysj/publications/GI2023/index.html}{in our website}. 

\subsubsection{Case of a System Involving a Mix of the Two Previous Cases} 

%%%%%%%%%%%%%%%%%%%%%%%%%%%%%%%%%%%%%%%%%%%%%%%%%%%%%%%%%%%%%%%%%%%%%%%%%%%%%
%%%%%%%%%%%%%%%%% 5DoF
%%%%%%%%%%%%%%%%%%%%%%%%%%%%%%%%%%%%%%%%%%%%%%%%%%%%%%%%%%%%%%%%%%%%%%%%%%%%%
\begin{figure}[t!] %[!htb]
\centering

\begin{subfigure}{.23\textwidth}
  \centering
  % include first image
  \includegraphics[width=4.6cm]{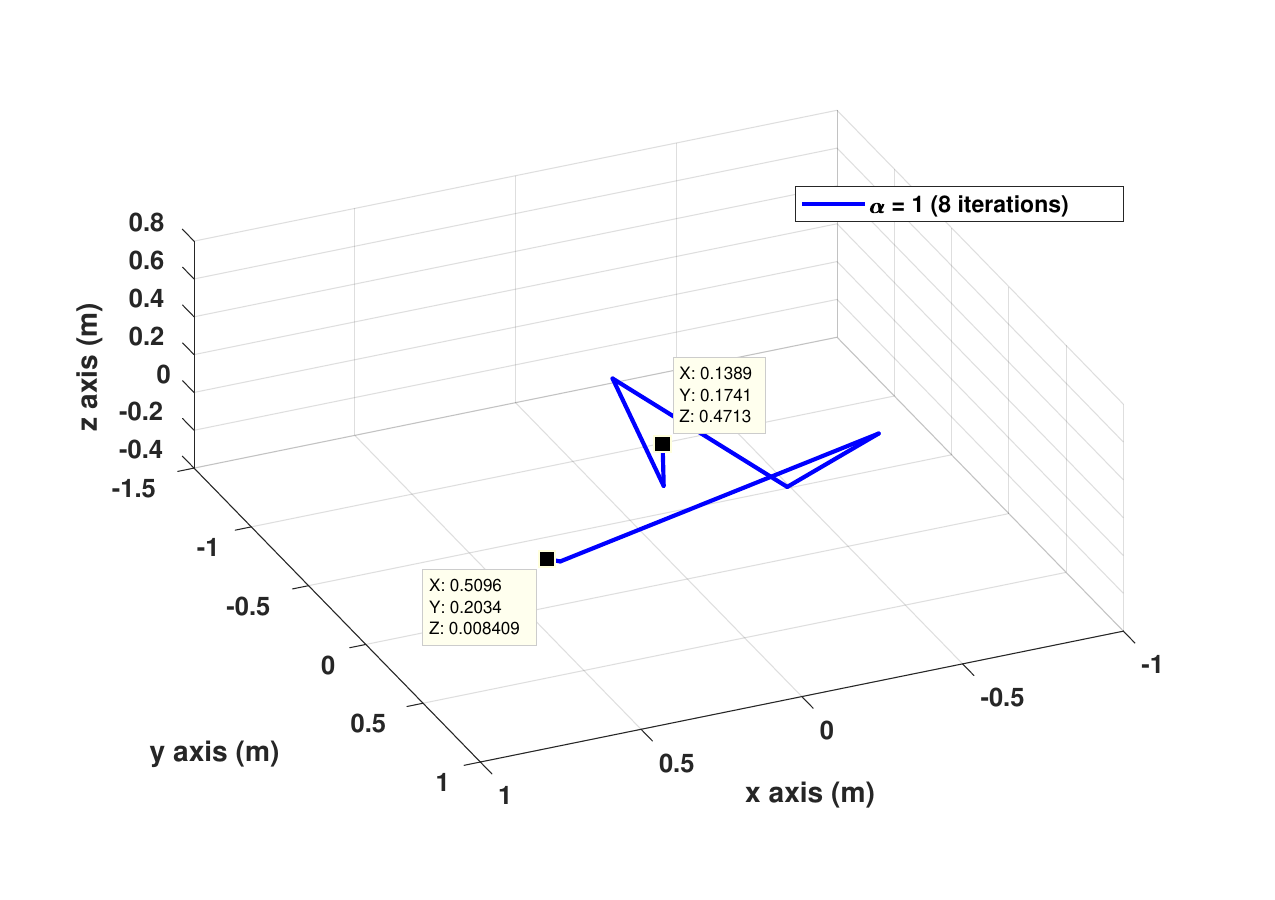}  
  \caption{\footnotesize MP / m / $\alpha=1$}
  \label{fig:5dof-mp-alpha-m}
\end{subfigure}
\begin{subfigure}{.23\textwidth}
  \centering
  % include third image
  \includegraphics[width=4.6cm]{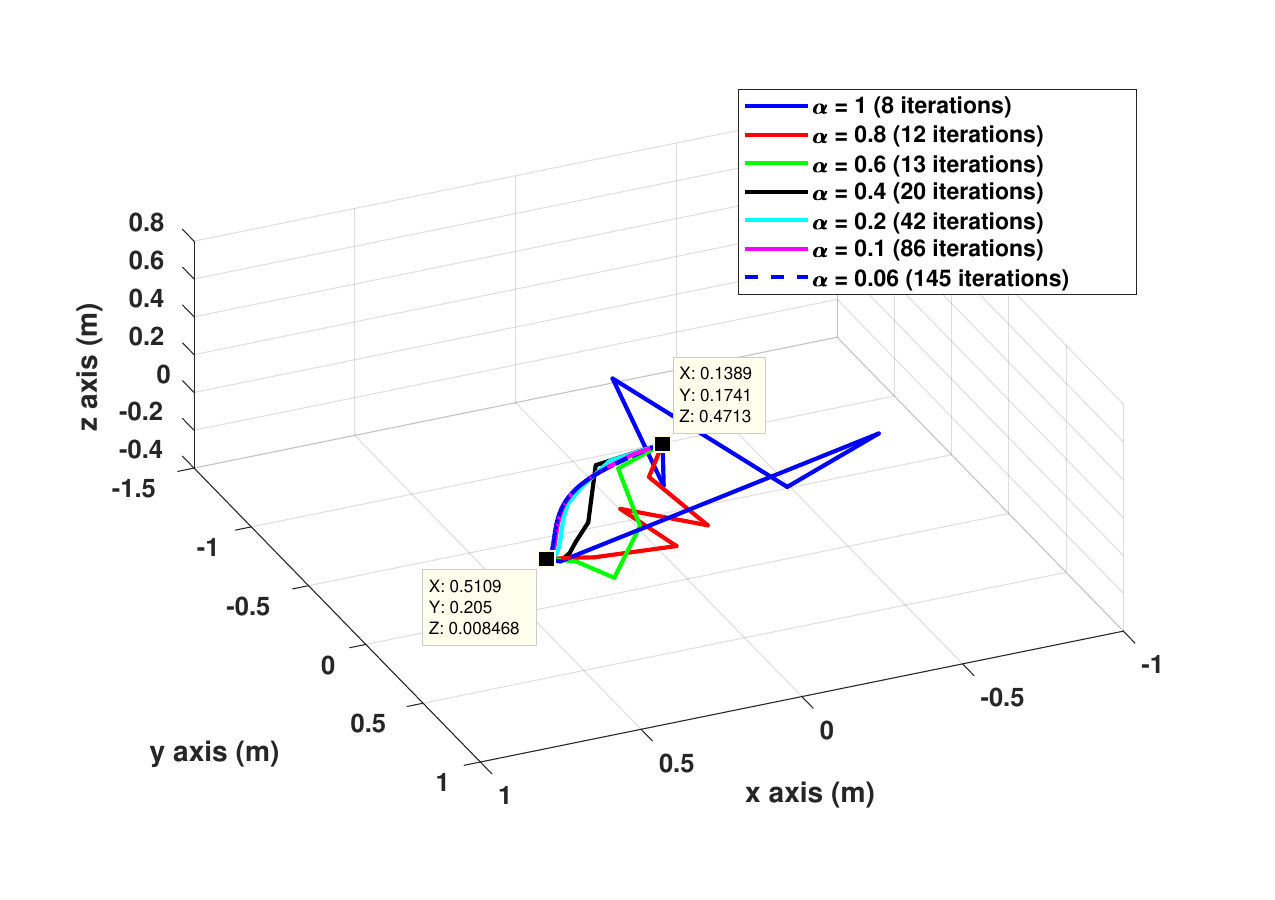}  
  \caption{\footnotesize MP / m / multiple $\alpha$}
  \label{fig:5dof-mp-alphas-m}
\end{subfigure}

\begin{subfigure}{.23\textwidth}
  \centering
  % include second image
  \includegraphics[width=4.6cm]{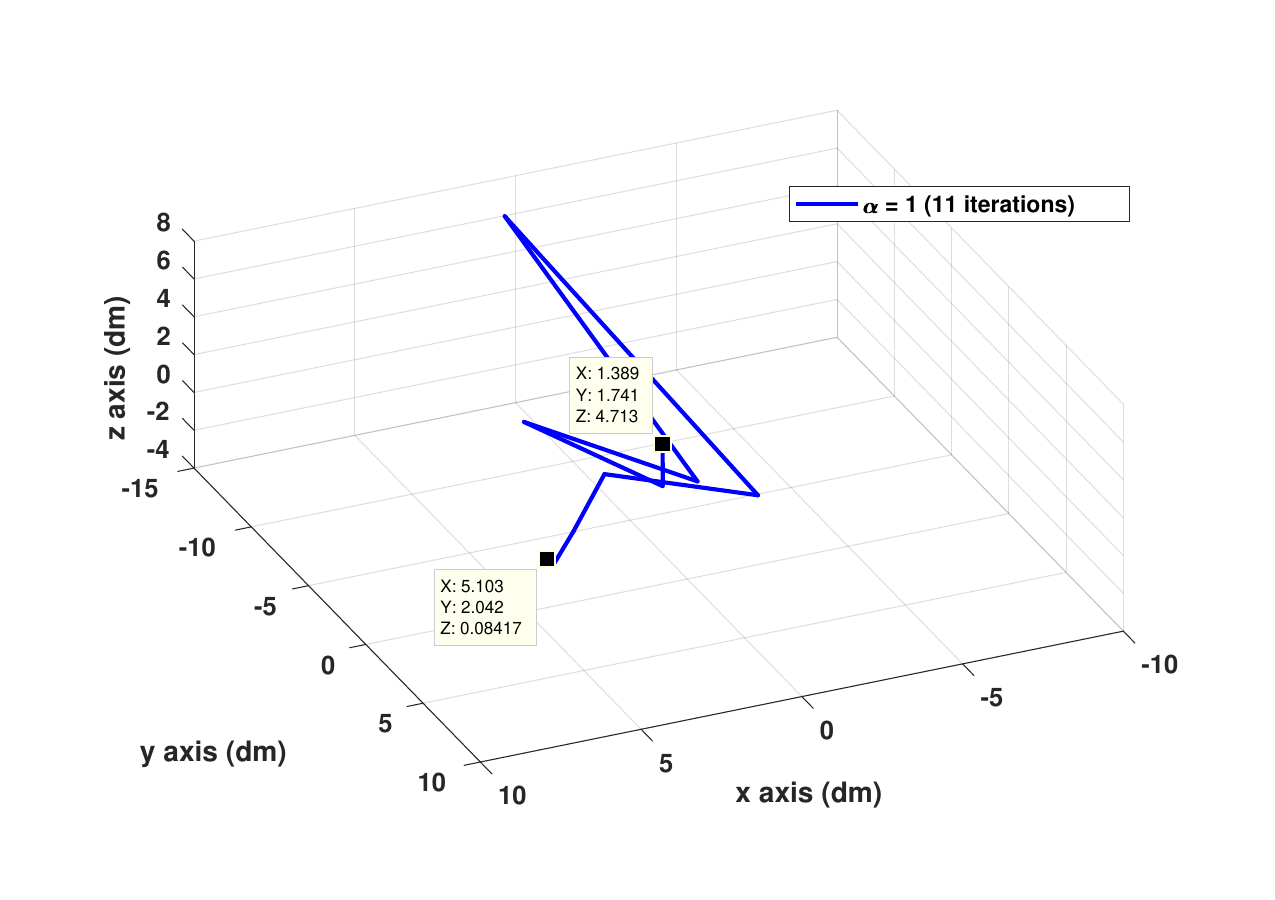}  
  \caption{\footnotesize MP / dm / $\alpha=1$}
  \label{fig:5dof-mp-alpha-dm}
\end{subfigure}
\begin{subfigure}{.23\textwidth}
  \centering
  % include fourth image
  \includegraphics[width=4.6cm]{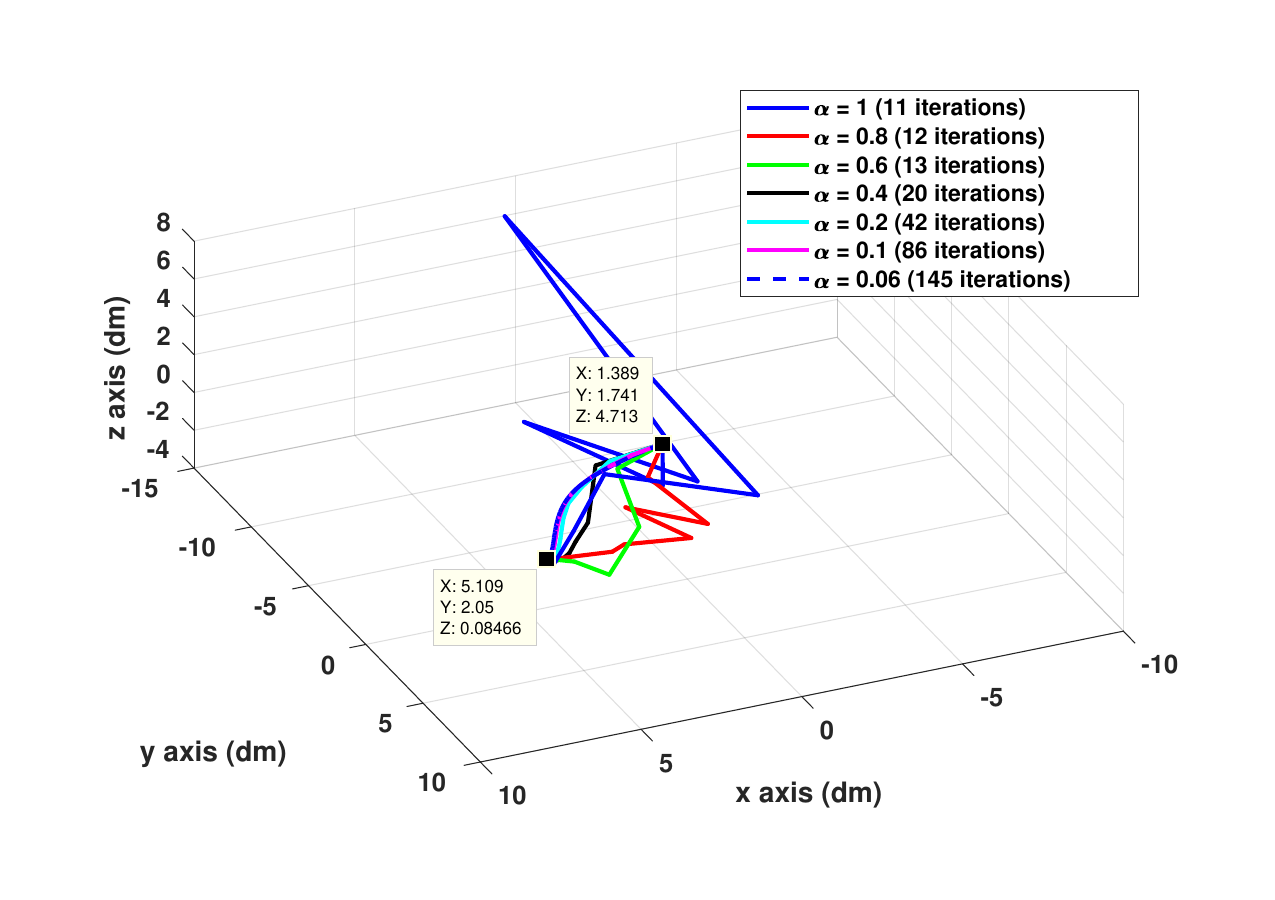}  
  \caption{\footnotesize MP / dm / multiple $\alpha$}
  \label{fig:5dof-mp-alphas-dm}
\end{subfigure}

\begin{subfigure}{.23\textwidth}
  \centering
  % include second image
  \includegraphics[width=4.6cm]{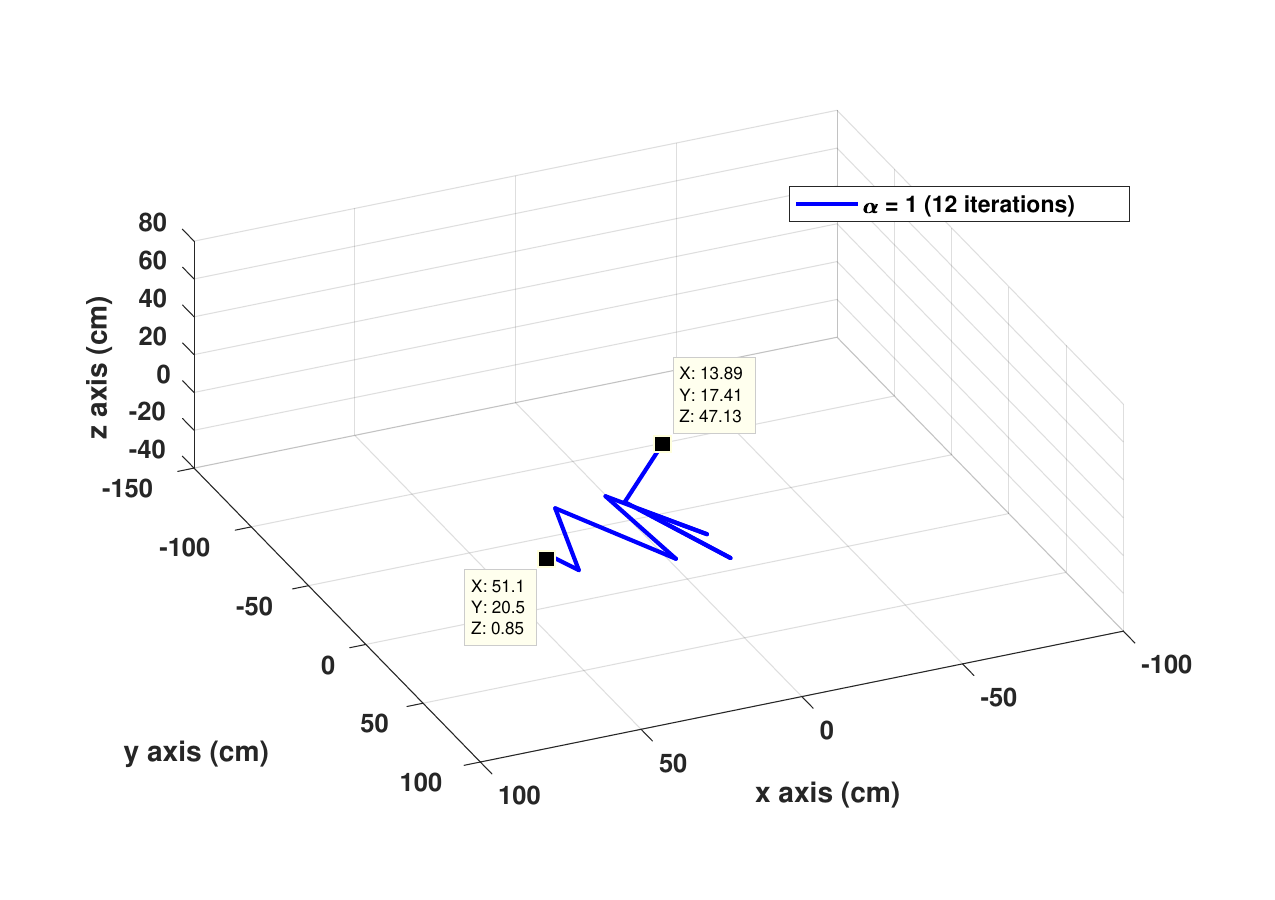}  
  \caption{\footnotesize MP / cm /  $\alpha=1$}
  \label{fig:5dof-mp-alpha-cm}
\end{subfigure}
\begin{subfigure}{.23\textwidth}
  \centering
  % include fourth image
  \includegraphics[width=4.6cm]{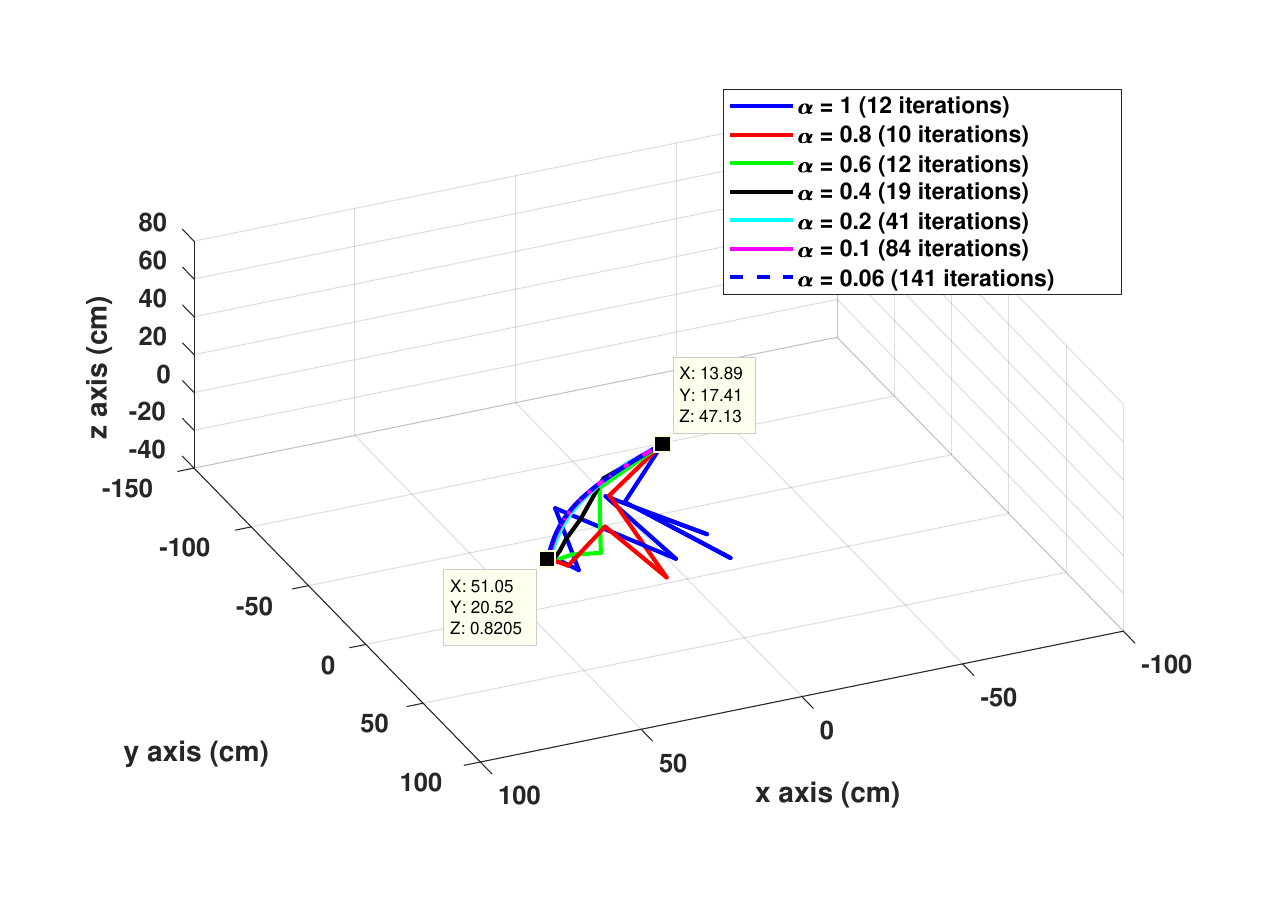}  
  \caption{\footnotesize MP / cm / multiple $\alpha$}
  \label{fig:5dof-mp-alphas-cm}
\end{subfigure}

\begin{subfigure}{.23\textwidth}
  \centering
  % include second image
  \includegraphics[width=4.6cm]{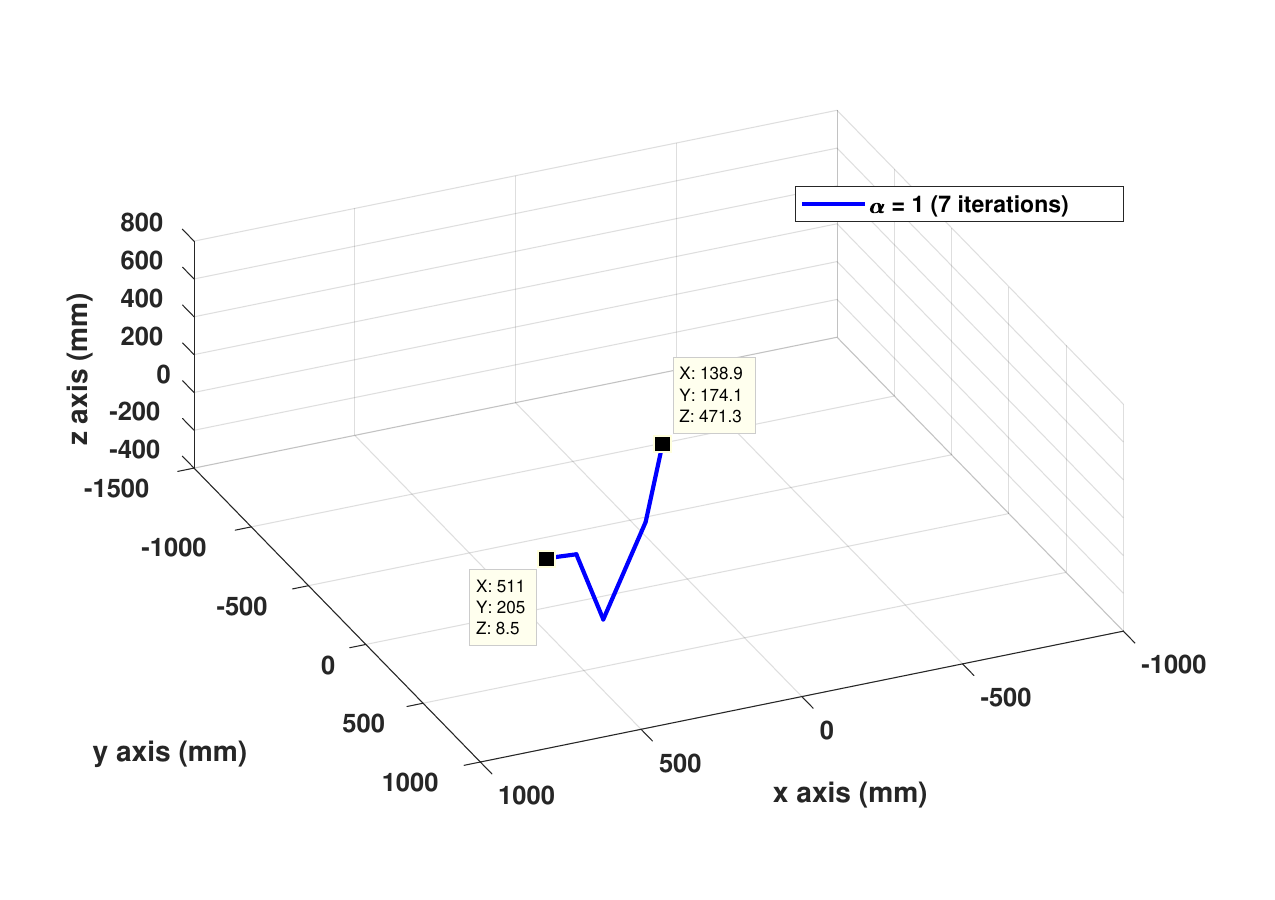}  
  \caption{\footnotesize MP / mm /  $\alpha=1$}
  \label{fig:5dof-mp-alpha-mm}
\end{subfigure}
\begin{subfigure}{.23\textwidth}
  \centering
  % include fourth image
  \includegraphics[width=4.6cm]{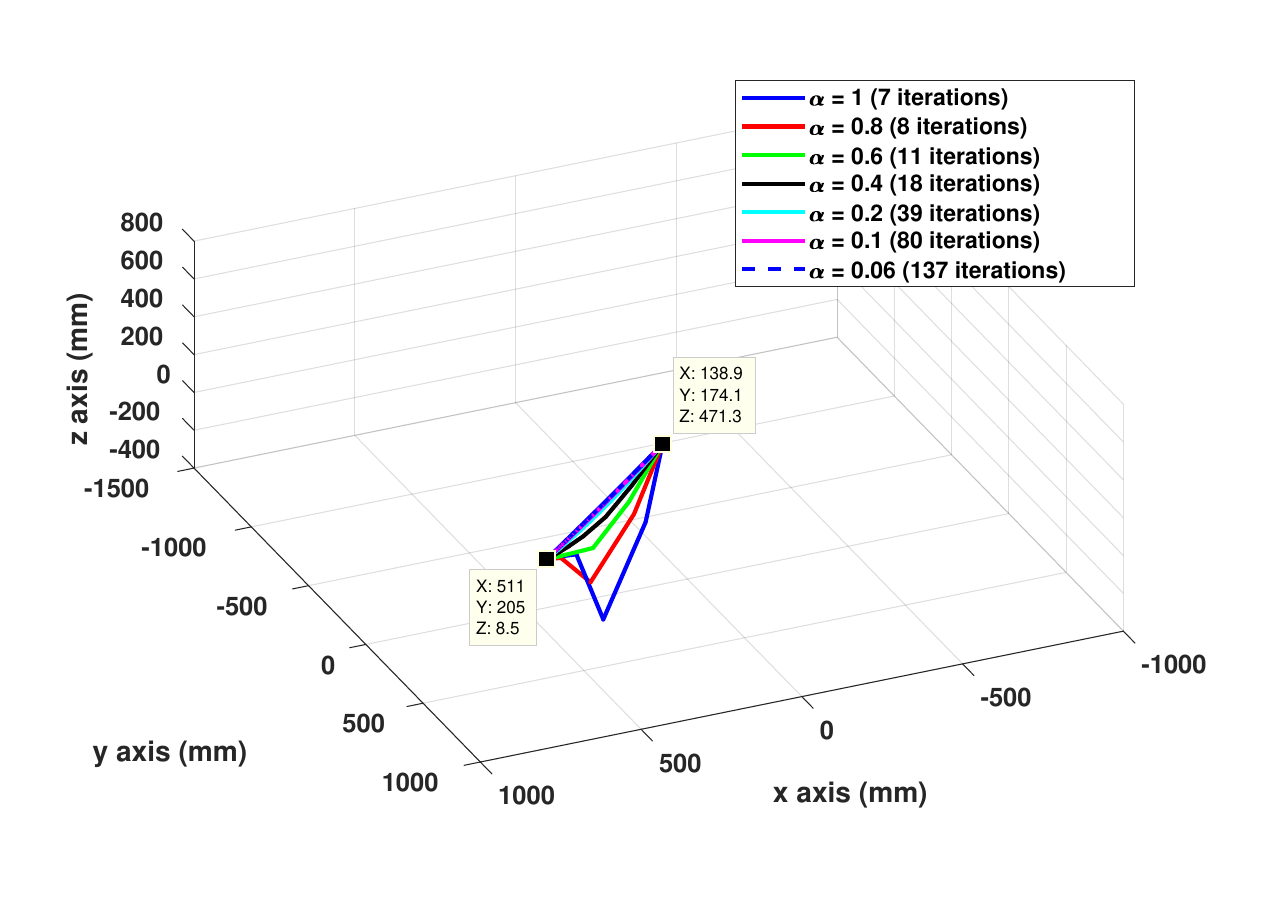}  
  \caption{\footnotesize MP / mm / multiple $\alpha$}
  \label{fig:5dof-mp-alphas-mm}
\end{subfigure}

\caption{\footnotesize Behavior of the trajectories of the end-effector of the 5DoF robot when varying the units while using the MP inverse with (\ref{fig:5dof-mp-alpha-m}), (\ref{fig:5dof-mp-alpha-dm}), (\ref{fig:5dof-mp-alpha-cm}), and  (\ref{fig:5dof-mp-alpha-mm}) the attenuation parameter $\alpha = 1$ and (\ref{fig:5dof-mp-alphas-m}), (\ref{fig:5dof-mp-alphas-dm}), (\ref{fig:5dof-mp-alphas-cm}), and (\ref{fig:5dof-mp-alphas-mm}) multiple values of $\alpha$.}
\label{fig:trajectories-MP-5DoF-alpha(s)}
\vspace{-7mm}
\end{figure}

\begin{figure}[t!] %[!htb]
\centering

\begin{subfigure}{.23\textwidth}
  \centering
  % include first image
  \includegraphics[width=4.6cm]{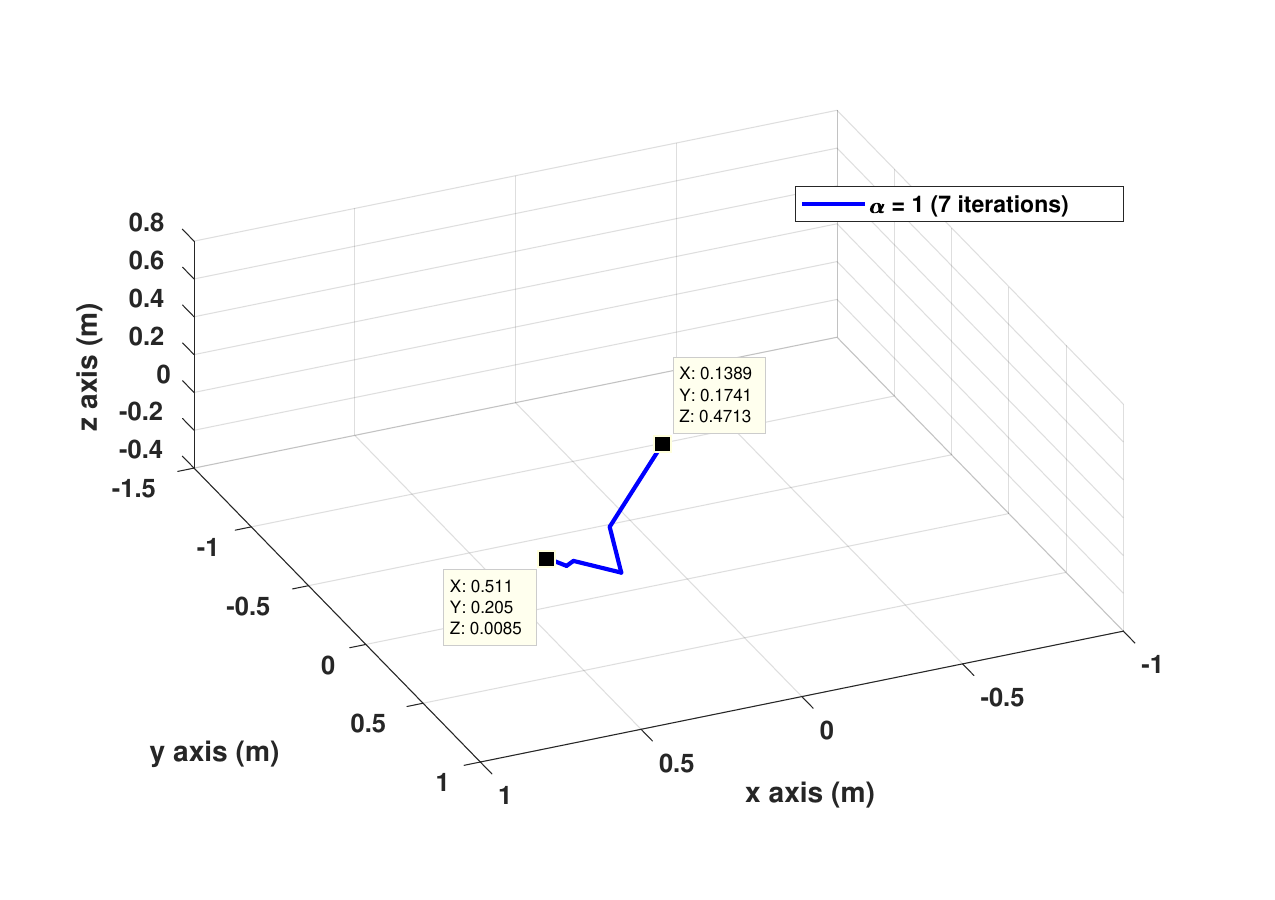}  
  \caption{\footnotesize UC / m / $\alpha=1$}
  \label{fig:5dof-uc-alpha-m}
\end{subfigure}
\begin{subfigure}{.23\textwidth}
  \centering
  % include third image
  \includegraphics[width=4.6cm]{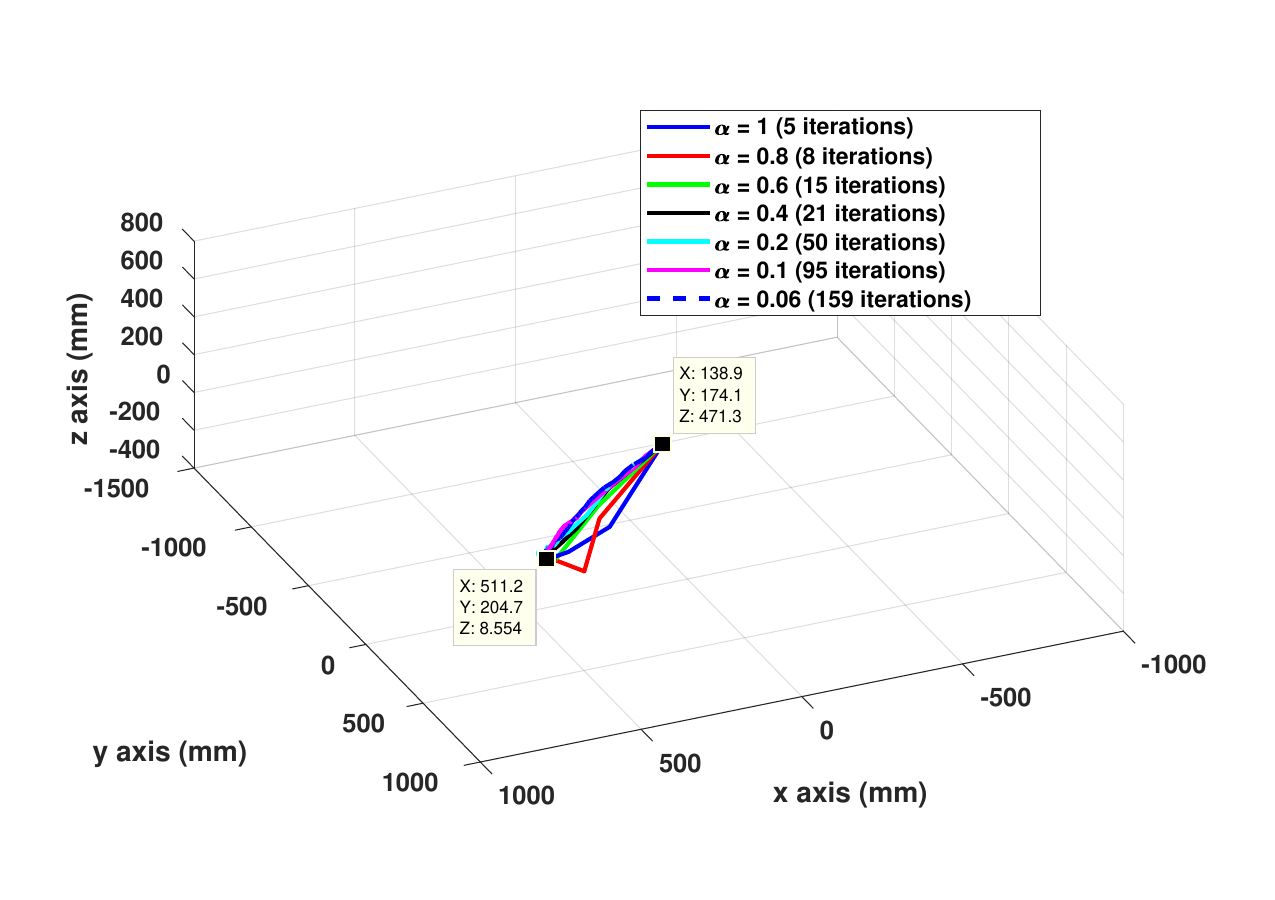}  
  \caption{\footnotesize UC / m / multiple $\alpha$}
  \label{fig:5dof-uc-alphas-m}
\end{subfigure}

\begin{subfigure}{.23\textwidth}
  \centering
  % include second image
  \includegraphics[width=4.6cm]{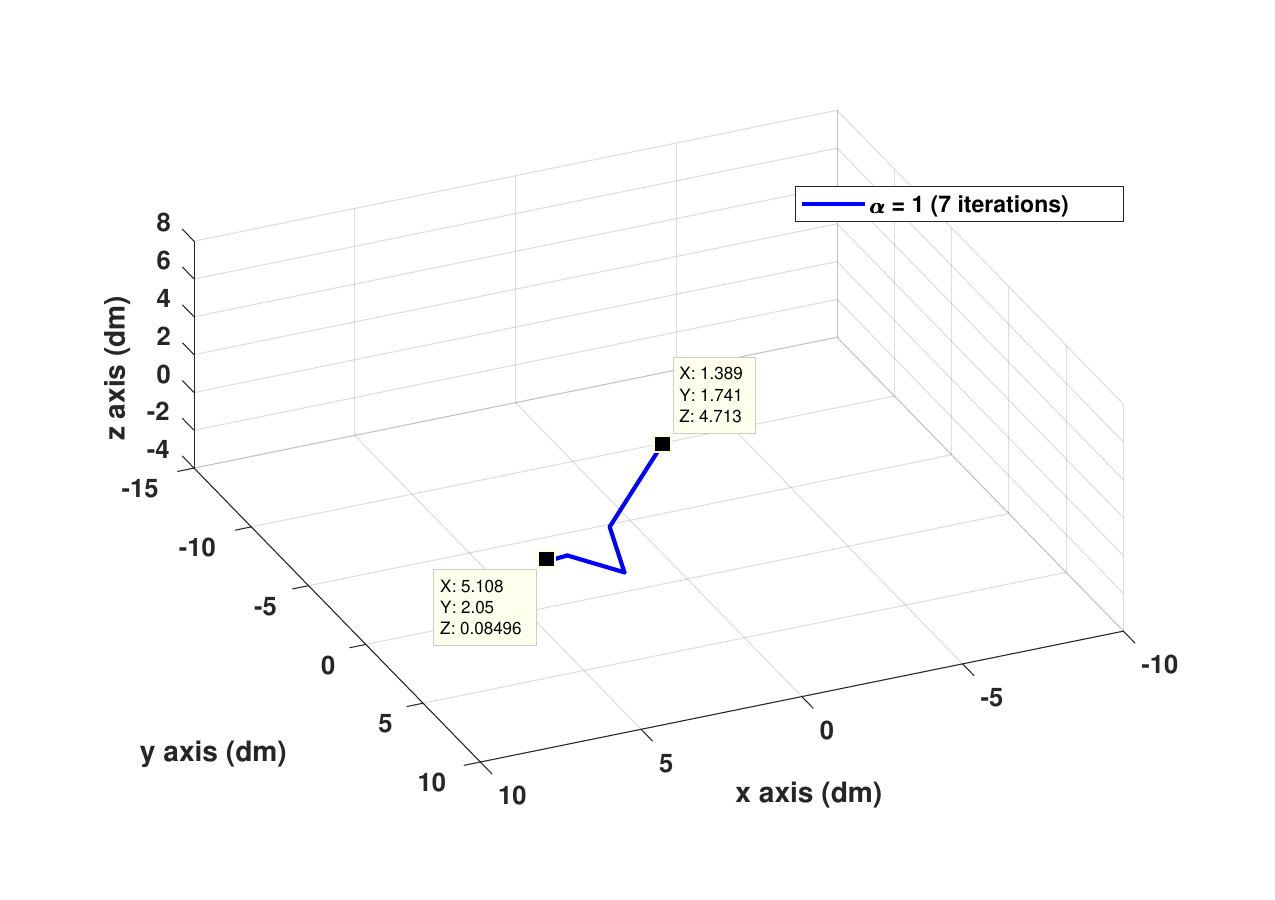}  
  \caption{\footnotesize UC / dm / $\alpha=1$}
  \label{fig:5dof-uc-alpha-dm}
\end{subfigure}
\begin{subfigure}{.23\textwidth}
  \centering
  % include fourth image
  \includegraphics[width=4.6cm]{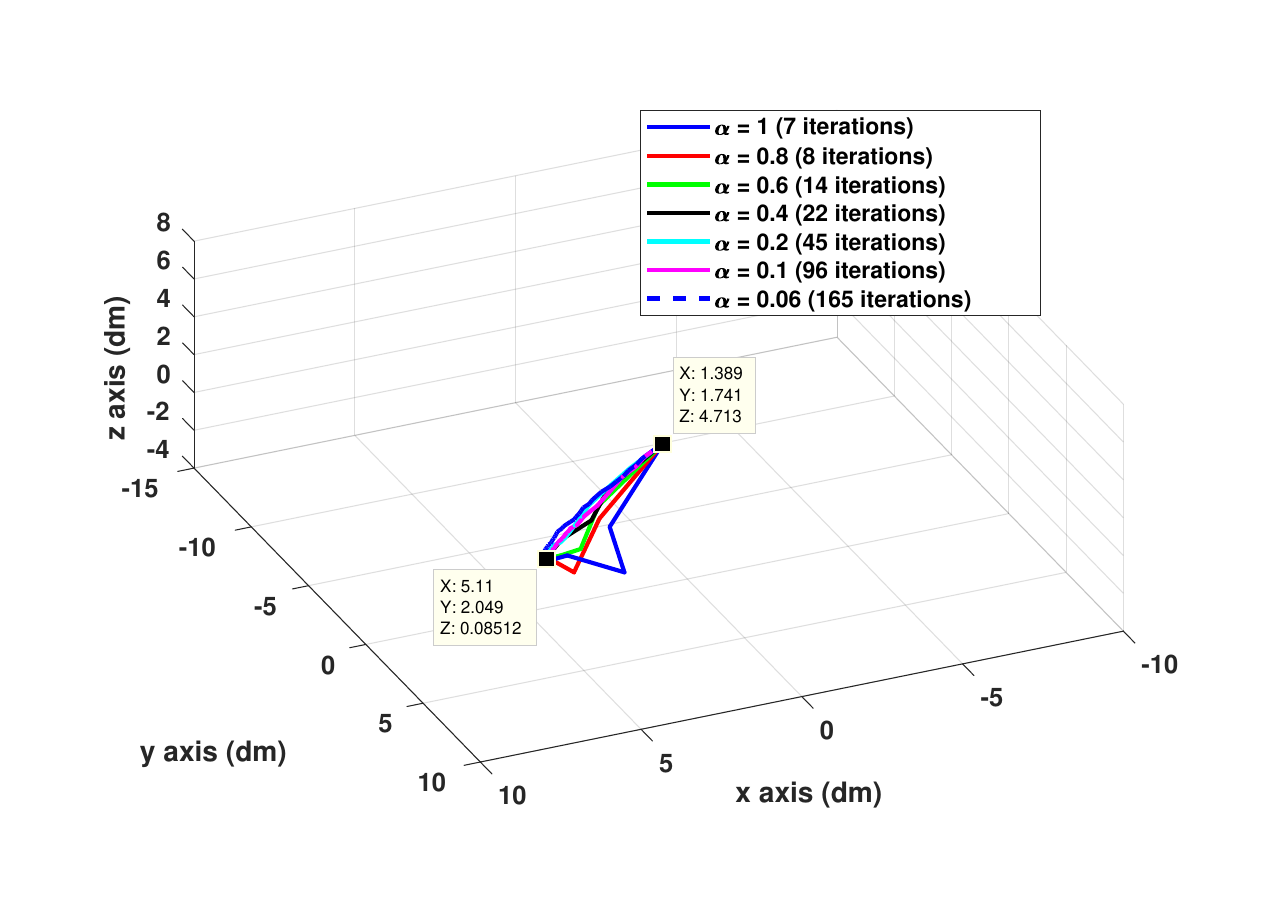}  
  \caption{\footnotesize UC / dm / multiple $\alpha$}
  \label{fig:5dof-uc-alphas-dm}
\end{subfigure}

\begin{subfigure}{.23\textwidth}
  \centering
  % include second image
  \includegraphics[width=4.6cm]{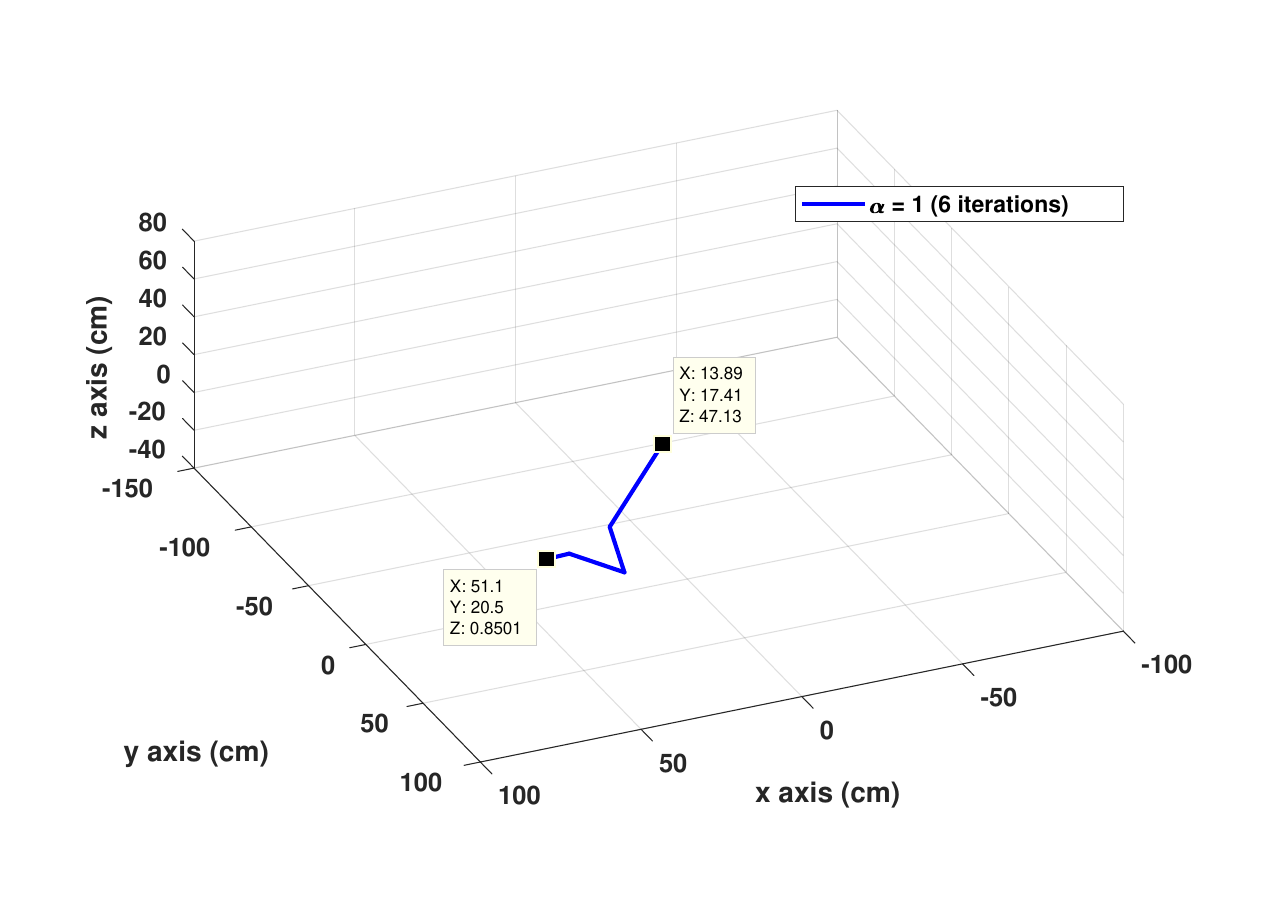}  
  \caption{\footnotesize UC / cm / $\alpha=1$}
  \label{fig:5dof-uc-alpha-cm}
\end{subfigure}
\begin{subfigure}{.23\textwidth}
  \centering
  % include fourth image
  \includegraphics[width=4.6cm]{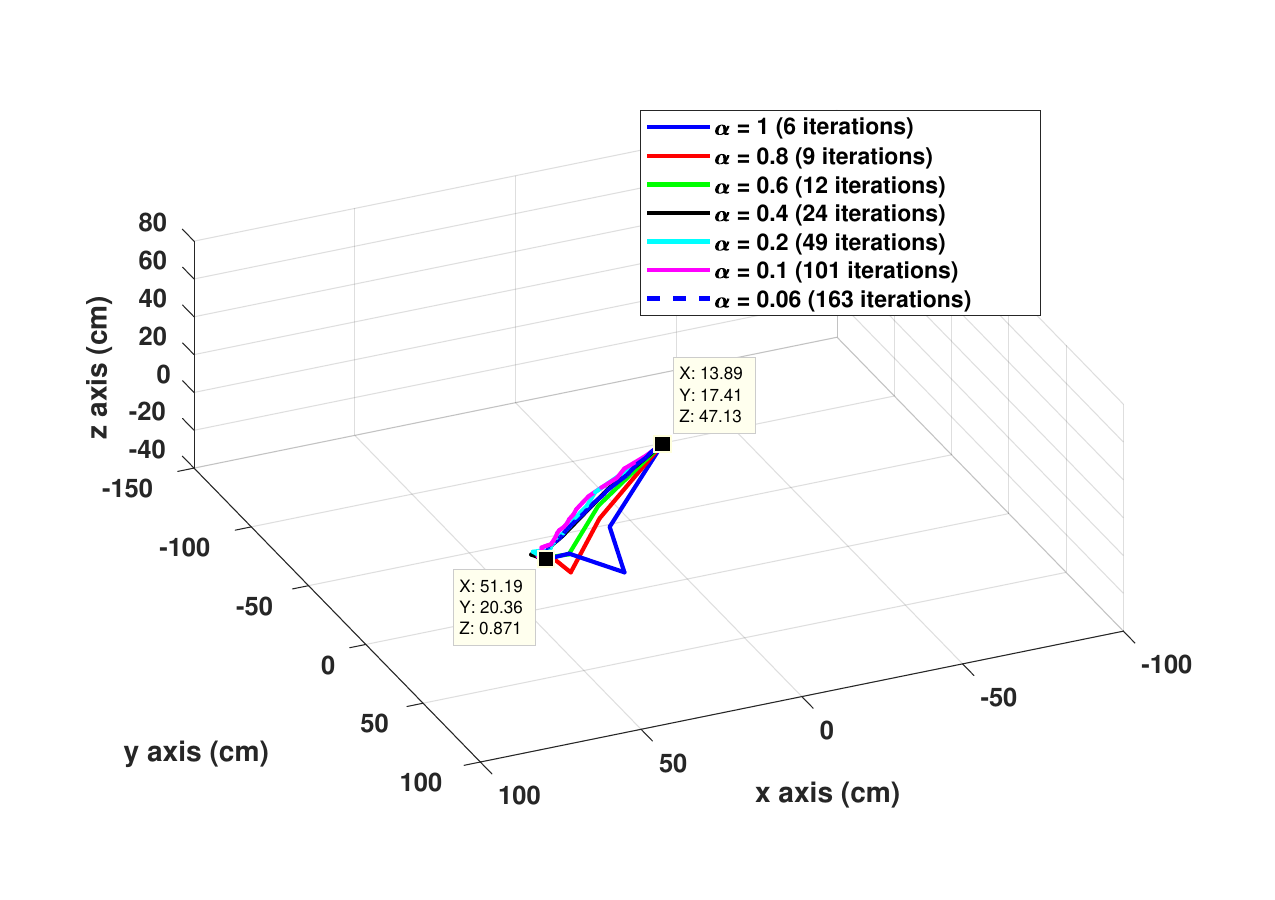}  
  \caption{\footnotesize UC / cm / multiple $\alpha$}
  \label{fig:5dof-uc-alphas-cm}
\end{subfigure}

\begin{subfigure}{.23\textwidth}
  \centering
  % include second image
  \includegraphics[width=4.6cm]{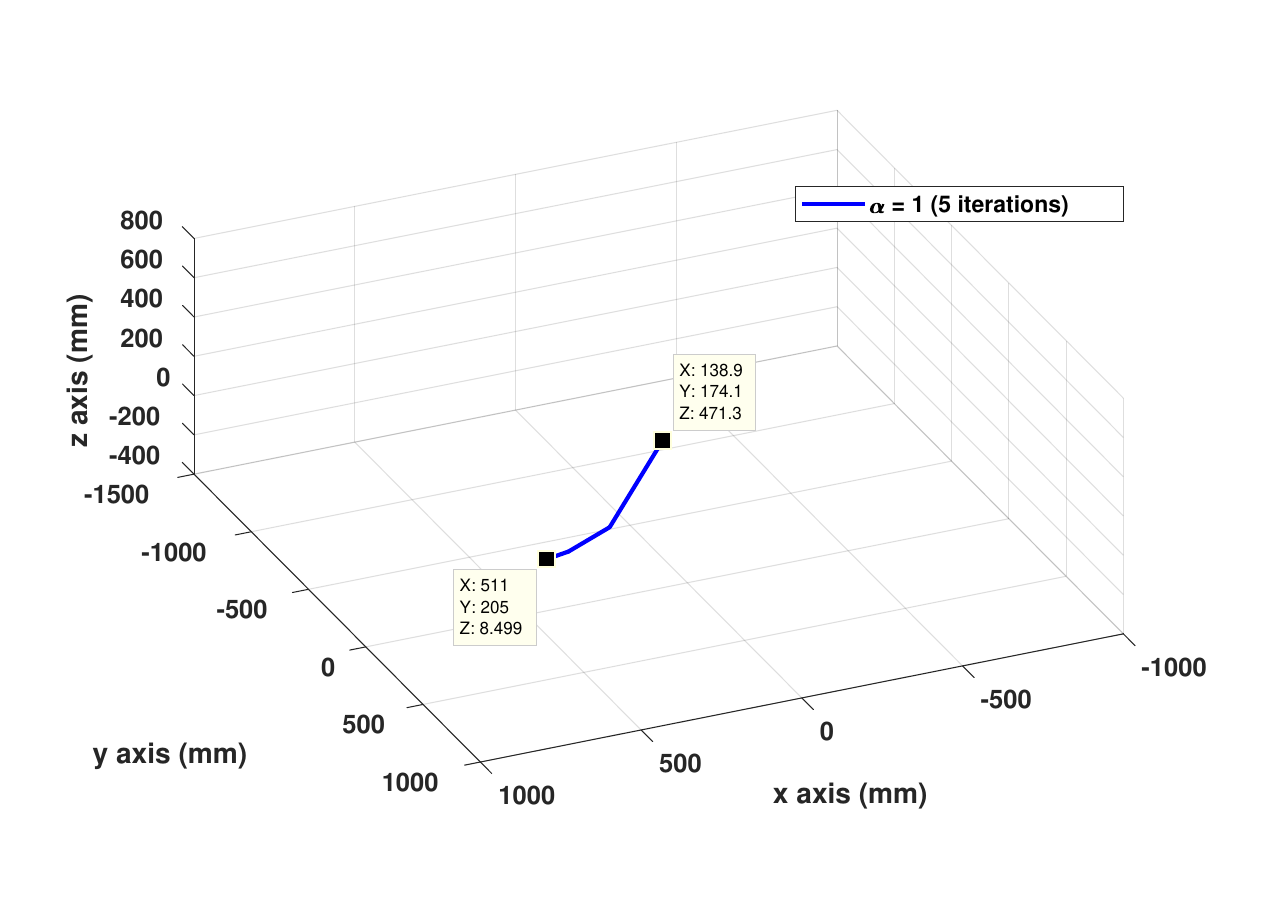}  
  \caption{\footnotesize UC / mm / $\alpha=1$}
  \label{fig:5dof-uc-alpha-mm}
\end{subfigure}
\begin{subfigure}{.23\textwidth}
  \centering
  % include fourth image
  \includegraphics[width=4.6cm]{images/p2_5dof_mm_uc_a.pdf}  
  \caption{\footnotesize UC / mm / multiple $\alpha$}
  \label{fig:5dof-uc-alphas-mm}
\end{subfigure}

\caption{\footnotesize Behavior of the trajectories of the end-effector of the 5DoF robot when varying the units while using the UC inverse with (\ref{fig:5dof-uc-alpha-m}), (\ref{fig:5dof-uc-alpha-dm}), (\ref{fig:5dof-uc-alpha-cm}), and (\ref{fig:5dof-uc-alpha-mm}) the attenuation parameter $\alpha = 1$ and (\ref{fig:5dof-uc-alphas-m}), (\ref{fig:5dof-uc-alphas-dm}), (\ref{fig:5dof-uc-alphas-cm}), and  (\ref{fig:5dof-uc-alphas-mm}) multiple values of $\alpha$.}
\label{fig:trajectories-UC-5DoF-alpha(s)}
\vspace{-6mm}
\end{figure}

\begin{figure}[t!] %[!htb]
%%%%%%%%%%%%%%%%% MX Inverse - 5DoF alpha = 1
\begin{subfigure}{.23\textwidth}
  \centering
  % include first image
  \includegraphics[width=4.6cm]{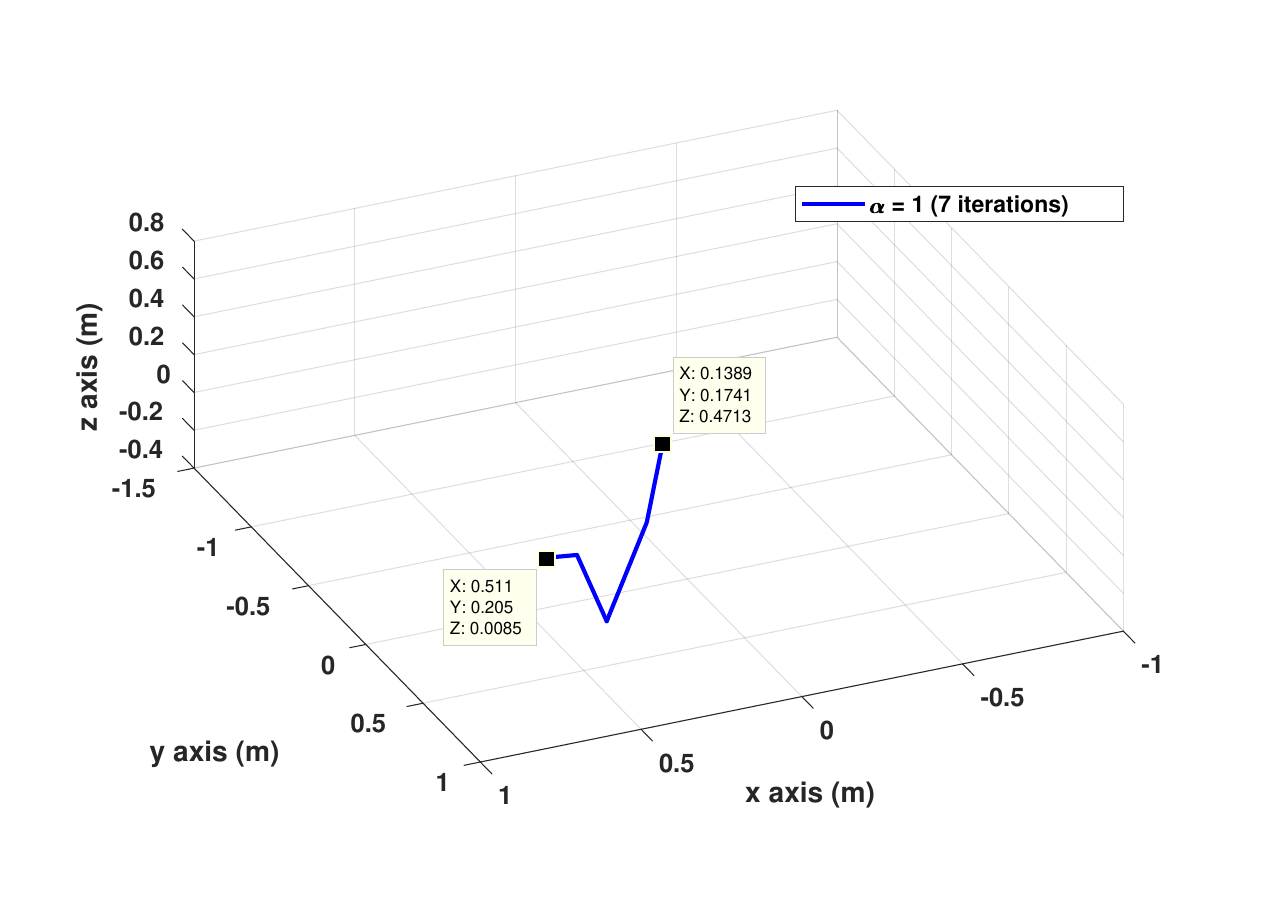}  
  \caption{\footnotesize MX / m / $\alpha=1$}
  \label{fig:5dof-mx-alpha-m}
\end{subfigure}
\begin{subfigure}{.23\textwidth}
  \centering
  % include third image
  \includegraphics[width=4.6cm]{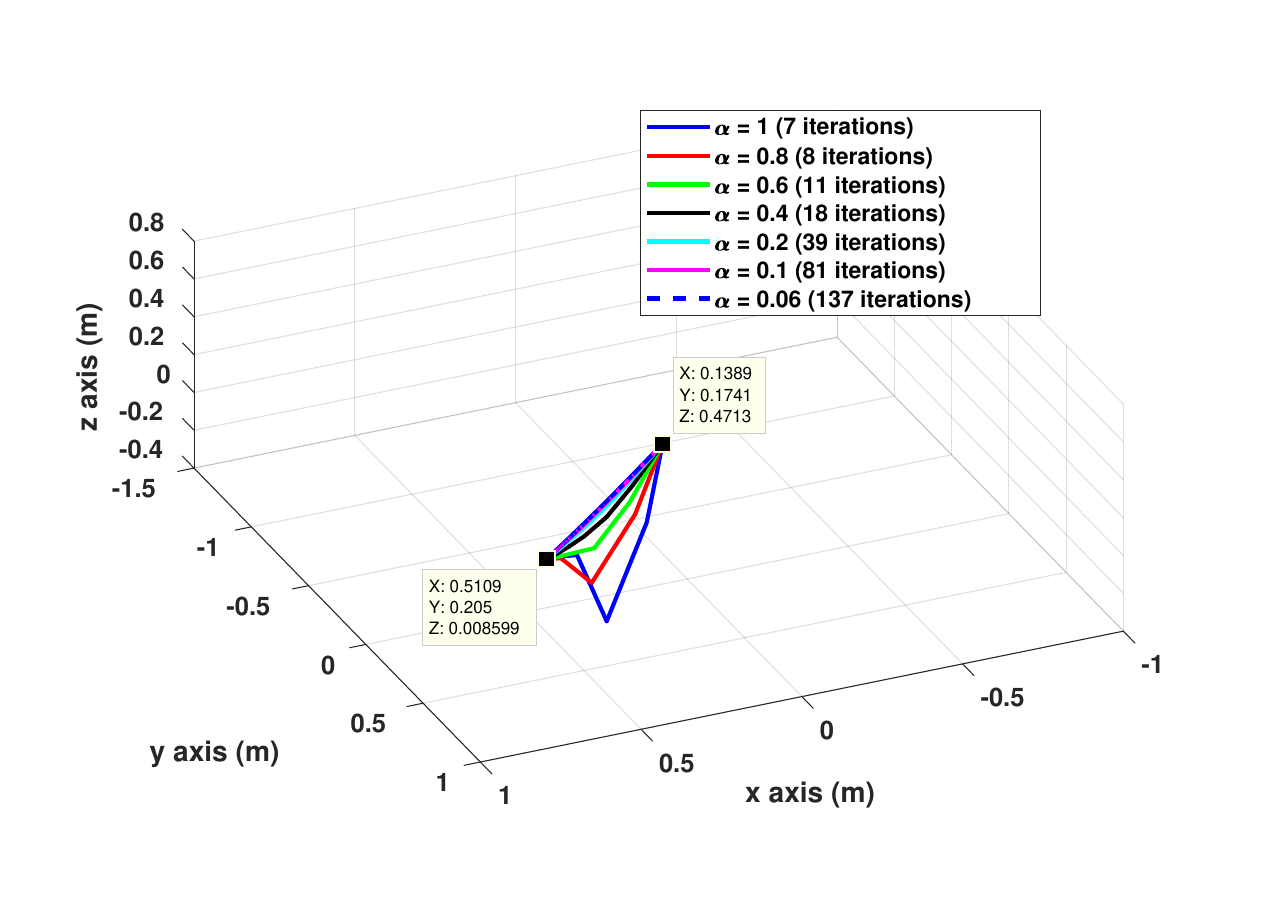}  
  \caption{\footnotesize MX / m / multiple $\alpha$}
  \label{fig:5dof-mx-alphas-m}
\end{subfigure}

\begin{subfigure}{.23\textwidth}
  \centering
  % include second image
  \includegraphics[width=4.6cm]{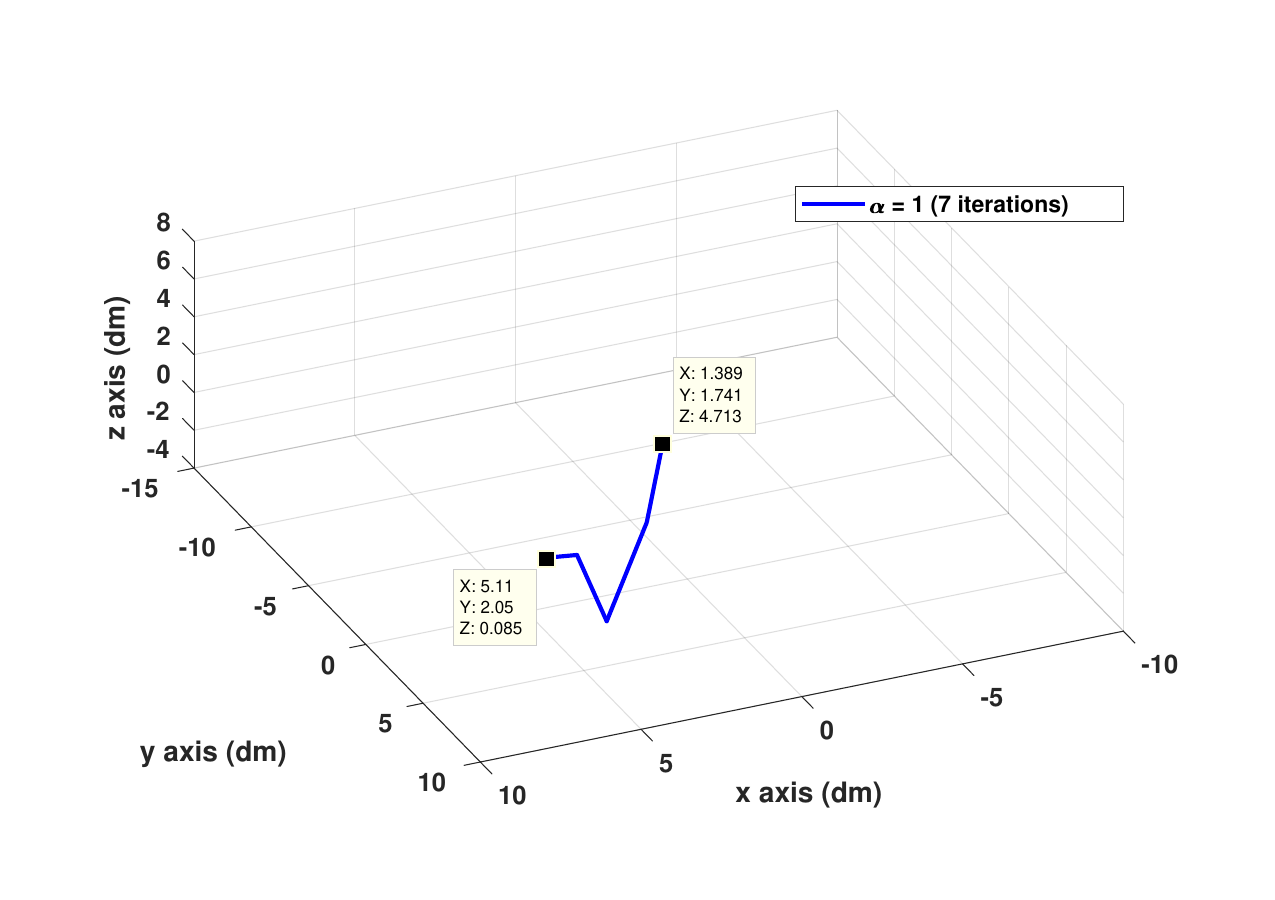}  
  \caption{\footnotesize MX / dm / $\alpha=1$}
  \label{fig:5dof-mx-alpha-dm}
\end{subfigure}
\begin{subfigure}{.23\textwidth}
  \centering
  % include fourth image
  \includegraphics[width=4.6cm]{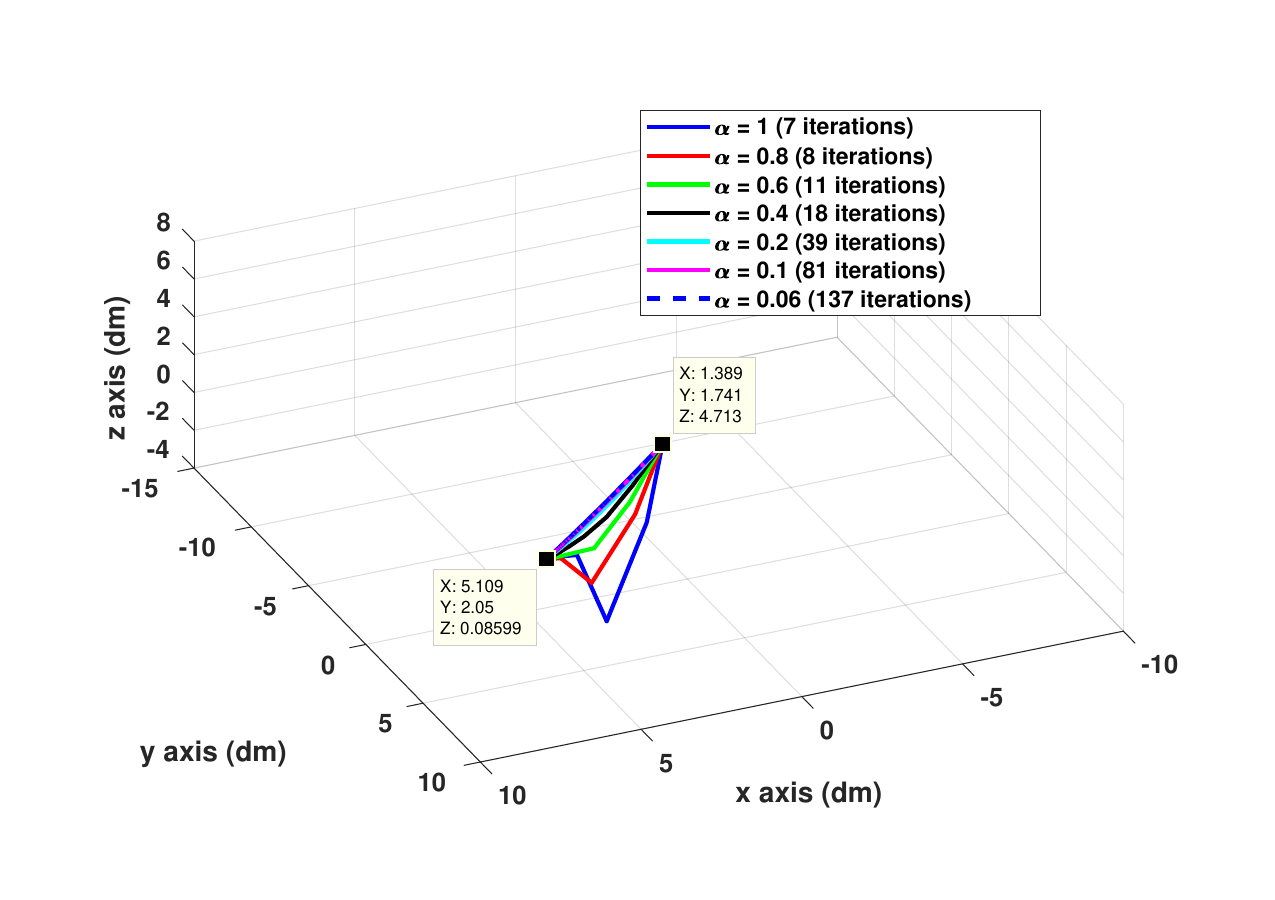}  
  \caption{\footnotesize MX / dm / multiple $\alpha$}
  \label{fig:5dof-mx-alphas-dm}
\end{subfigure}

\begin{subfigure}{.23\textwidth}
  \centering
  % include second image
  \includegraphics[width=4.6cm]{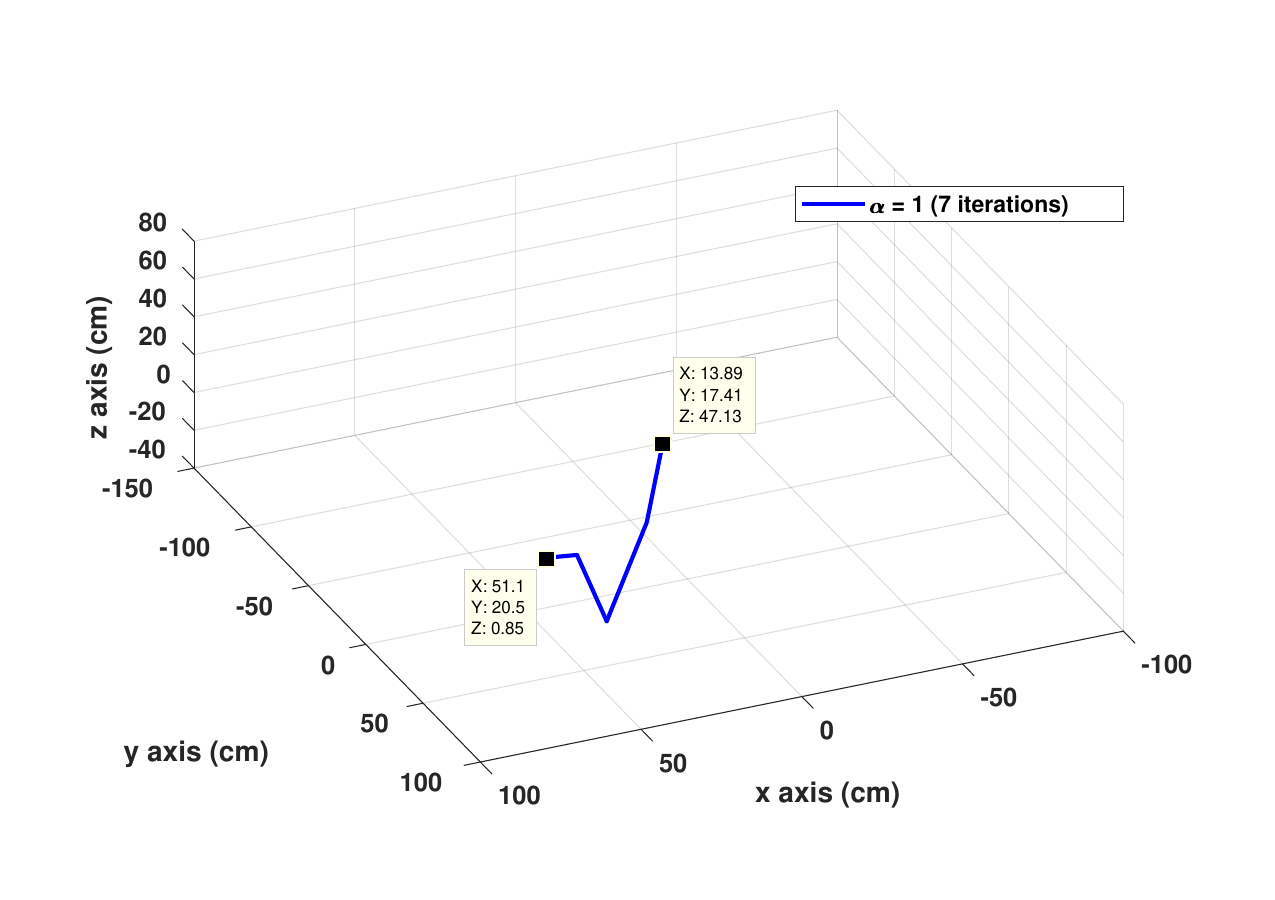}  
  \caption{\footnotesize MX / cm / $\alpha=1$}
  \label{fig:5dof-mx-alpha-cm}
\end{subfigure}
\begin{subfigure}{.23\textwidth}
  \centering
  % include fourth image
  \includegraphics[width=4.6cm]{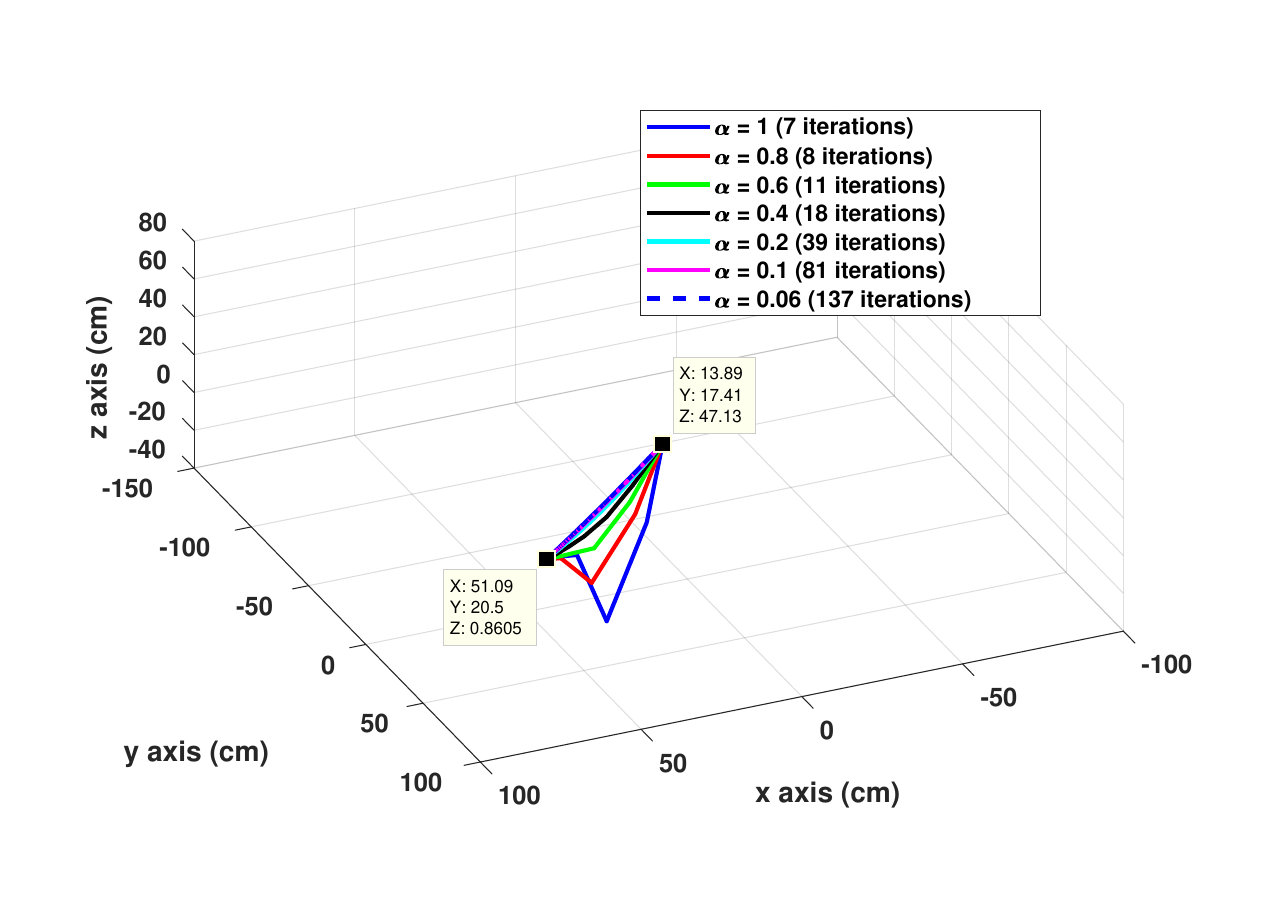}  
  \caption{\footnotesize MX / cm / multiple $\alpha$}
  \label{fig:5dof-mx-alphas-cm}
\end{subfigure}

\begin{subfigure}{.23\textwidth}
  \centering
  % include second image
  \includegraphics[width=4.6cm]{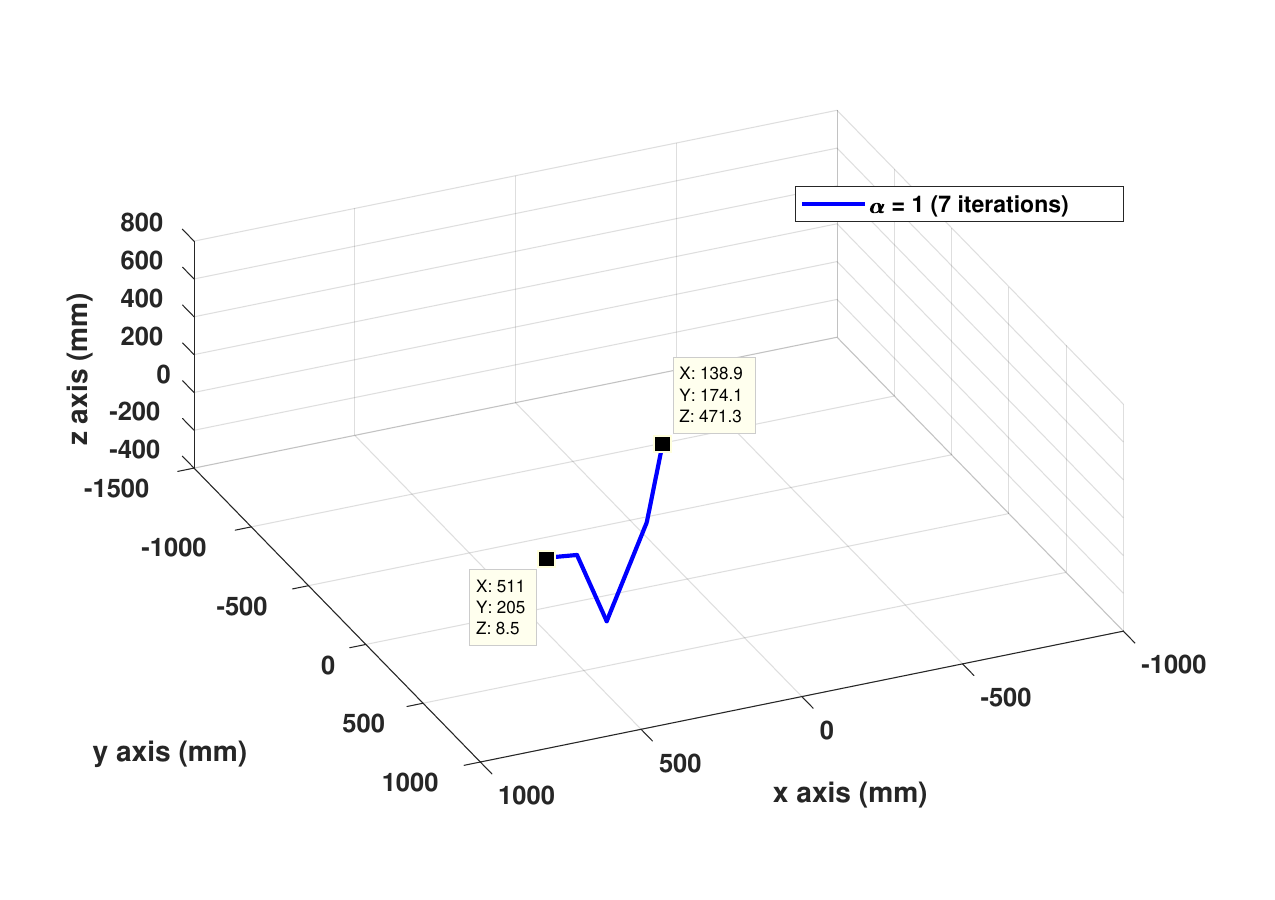}  
  \caption{\footnotesize MX / mm / $\alpha=1$}
  \label{fig:5dof-mx-alpha-mm}
\end{subfigure}
\begin{subfigure}{.23\textwidth}
  \centering
  % include fourth image
  \includegraphics[width=4.6cm]{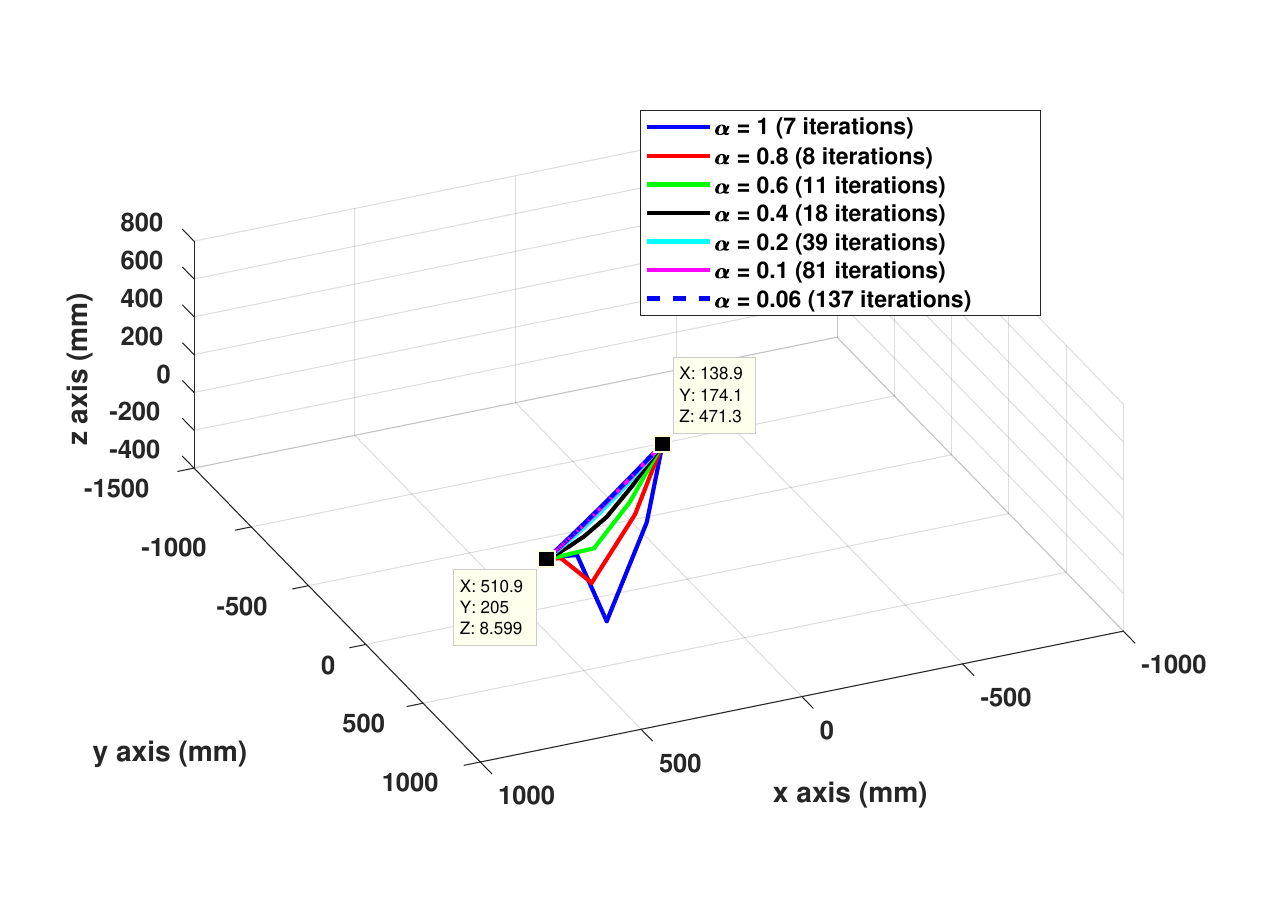}  
  \caption{\footnotesize MX / mm / multiple $\alpha$}
  \label{fig:5dof-mx-alphas-mm}
\end{subfigure}

\caption{\footnotesize Behavior of the trajectories of the end-effector of the 5DoF robot when varying the units while using the MX inverse with (\ref{fig:5dof-mx-alpha-m}), (\ref{fig:5dof-mx-alpha-dm}), (\ref{fig:5dof-mx-alpha-cm}), and (\ref{fig:5dof-mx-alpha-mm}) the attenuation parameter $\alpha = 1$ and (\ref{fig:5dof-mx-alphas-m}), (\ref{fig:5dof-mx-alphas-dm}), (\ref{fig:5dof-mx-alphas-cm}), and (\ref{fig:5dof-mx-alphas-mm}) multiple values of $\alpha$.}
\label{fig:trajectories-MX-5DoF-alpha(s)}
\vspace{-6mm}
\end{figure}

 In this case, we explore the 5DoF RRPRR (modified) Stanford manipulator with a non-full ranked Jacobian to evaluate the proposed rule-of-thumb  when the MX inverse does not reduce to either the MP or UC inverses. Here, since the prismatic joint appears in the middle between two sets of two revolute joints, and since the $Z-axis$ of the prior revolute joints are not aligned with the $Z-axis$ of the prismatic one; a combination of MP and UC in the MX inverse is required -- i.e. the MX does not reduce to either the MP nor the UC. Therefore, the use of either one exclusively leads to inconsistencies and uncontrollable paths, and only the MX can provide stable motion under any choice of the attenuation parameter. More specifically, an inspection of the two early rotations of this RRPRR manipulator shows that both $\theta_1$ and $\theta_2$ affect the linear (prismatic) joint $d_3$, which in turn affects $X$, $Y$, and $Z$ coordinates. So, the block partitioning of the MX inverse applied to the Jacobian of this 5DoF robot becomes as follows, given the Jacobian matrix $J$: 
\begin{equation} \label{eq14}
J= \begin{bmatrix}
A_W & A_X\\
A_Y & A_Z
\end{bmatrix}
= \begin{bmatrix}
\frac{\partial X}{\partial \theta_1} & \frac{\partial X}{\partial \theta_2} & \frac{\partial X}{\partial d_3} & \frac{\partial X}{\partial \theta_4} & \frac{\partial X}{\partial \theta_5}\\
\frac{\partial Y}{\partial \theta_1} & \frac{\partial Y}{\partial \theta_2} & \frac{\partial Y}{\partial d_3} & \frac{\partial Y}{\partial \theta_4} & \frac{\partial Y}{\partial \theta_5}\\
\frac{\partial Z}{\partial \theta_1} & \frac{\partial Z}{\partial \theta_2} & \frac{\partial Z}{\partial d_3} & \frac{\partial Z}{\partial \theta_4} & \frac{\partial Z}{\partial \theta_5}\\
\frac{\partial Ro}{\partial \theta_1} & \frac{\partial Ro}{\partial \theta_2} & \frac{\partial Ro}{\partial d_3} & \frac{\partial Ro}{\partial \theta_4} & \frac{\partial Ro}{\partial \theta_5}\\
\frac{\partial Pi}{\partial \theta_1} & \frac{\partial Pi}{\partial \theta_2} & \frac{\partial Pi}{\partial d_3} & \frac{\partial Pi}{\partial \theta_4} & \frac{\partial Pi}{\partial \theta_5}\\
\frac{\partial Ya}{\partial \theta_1} & \frac{\partial Ya}{\partial \theta_2} & \frac{\partial Ya}{\partial d_3} & \frac{\partial Ya}{\partial \theta_4} & \frac{\partial Ya}{\partial \theta_5}
\end{bmatrix}
\end{equation}

\noindent and the fact that the variables requiring unit consistency should be in $A_W$, so it is handled by the UC inverse, we have: 

\noindent$A_W = \left[\begin{array}{ccc}
\frac{\partial X}{\partial \theta_1} & \frac{\partial X}{\partial \theta_2} & \frac{\partial X}{\partial d_3}\\
\frac{\partial Y}{\partial \theta_1} & \frac{\partial Y}{\partial \theta_2} & \frac{\partial Y}{\partial d_3}\\
\frac{\partial Z}{\partial \theta_1} & \frac{\partial Z}{\partial \theta_2} & \frac{\partial Z}{\partial d_3}
\end{array}\right], \qquad \qquad A_X = \left[\begin{array}{cc}
\frac{\partial X}{\partial \theta_4} & \frac{\partial X}{\partial \theta_5}\\
\frac{\partial Y}{\partial \theta_4} & \frac{\partial Y}{\partial \theta_5}\\
\frac{\partial Z}{\partial \theta_4} & \frac{\partial Z}{\partial \theta_5}
\end{array}\right],$

\noindent$A_Y = \left[\begin{array}{ccc}
\frac{\partial Ro}{\partial \theta_1} & \frac{\partial Ro}{\partial \theta_2} & \frac{\partial Ro}{\partial d_3}\\
\frac{\partial Pi}{\partial \theta_1} & \frac{\partial Pi}{\partial \theta_2} & \frac{\partial Pi}{\partial d_3}\\
\frac{\partial Ya}{\partial \theta_1} & \frac{\partial Ya}{\partial \theta_2} & \frac{\partial Ya}{\partial d_3}
\end{array}\right], \qquad \qquad A_Z = \left[\begin{array}{ccc}
\frac{\partial Ro}{\partial \theta_4} & \frac{\partial Ro}{\partial \theta_5}\\
\frac{\partial Pi}{\partial \theta_4} & \frac{\partial Pi}{\partial \theta_5}\\
\frac{\partial Ya}{\partial \theta_4} & \frac{\partial Ya}{\partial \theta_5}
\end{array}\right].$

Based on this partitioning, the MX inverse $J^{-M}$ can be computed by replacing each block in equation (\ref{eq7}). In this case, the MX is not reduced to either MP or UC but needs the combination of both generalized inverses to produce unit invariant IK solutions. The resulting $J^{-M}$ is a 5x6 matrix. Once again, we provide the results obtained from applying the MP and UC inverses, before applying the MX inverse. 

The results presented in Figure \ref{fig:trajectories-MP-5DoF-alpha(s)} show that the behavior (path) of the robot end-effector is different when the unit of the linear joints in the 5DoF are varied from $m$, to $mm$. That is, for a robotic system that requires a mix of units and rotation consistency, the MP inverse alone cannot guarantee a reliable behavior of the system. The same observation is made when the attenuation parameter $\alpha$ is varied as depicted in Sub-Figures \ref{fig:5dof-mp-alphas-m}, \ref{fig:5dof-mp-alphas-dm}, \ref{fig:5dof-mp-alphas-cm} and \ref{fig:5dof-mp-alphas-mm}; it just cannot remedy the effect of the change of units.

We also applied the UC inverse to the same motion of the 5DoF robot. As depicted in Figure \ref{fig:trajectories-UC-5DoF-alpha(s)}, the behavior of the robot is still quite different when the units are varied from $m$, to $mm$ but also when the attenuation $\alpha$ is varied. That is, the UC inverse is unable to provide unit-consistency in this case and $\alpha$ cannot remedy the effects of the units change as depicted in Sub-Figures  \ref{fig:5dof-uc-alphas-m},  \ref{fig:5dof-uc-alphas-dm},  \ref{fig:5dof-uc-alphas-cm} and \ref{fig:5dof-uc-alphas-mm}.

Finally, we applied the MX inverse to this same motion of the 5DoF robot. As we explained at the beginning of this section, the MX inverse does not reduce to either the MP inverse or the UC inverse. Here, it combines both inverses using the block-matrix inverse as shown in equation (\ref{eq7}) to provide reliable results. In Figure \ref{fig:trajectories-MX-5DoF-alpha(s)}, we observe that the behavior of the robot end-effector is exactly the same as the units are varied from $m$ to $mm$. The results also show that the attenuation parameter $\alpha$ is able to smooth the end-effector path and remedy the effects of the unit-changes as depicted in Sub-Figures  \ref{fig:5dof-mx-alphas-m}, \ref{fig:5dof-mx-alphas-dm}, \ref{fig:5dof-mx-alphas-cm} and \ref{fig:5dof-mx-alphas-mm}.

\begin{table*}[!htb] %!htb
\notsotinyone 
\caption{\footnotesize Experimental IK results obtained using 1000 random motions of the 3DoF (2RP) manipulator generated using the D-H parameters and the forward kinematics. \textcolor{black}{($\alpha$ = 1)}}

\label{tab:results-3DoF}
\centering
    \begin{threeparttable}
         \begin{tabular}{|c|c|c|c|c|c|c|c|c|c|c|c|c|c|c|c|c|c|} 
         %\begin{tblr}{
         %       colspec = {|c|c|c|c|c|c|c|c|c|c|c|c|c|c|c|c|c|c|},
         %       row{11} = {gray9},
         %       row{12} = {gray9},
         %       row{21} = {gray9},
         %       row{22} = {gray9},
         %       row{31} = {gray9},
         %       row{32} = {gray9},
                %column{3} = {teal7},
                %cell{2}{3} = {yellow7},
         %    }
         \hline
         \multirow{2}{1em}{\textbf{$J^{\widetilde{-1}}$}} & \multicolumn{4}{|c|}{\textbf{m}} & \multicolumn{4}{|c|}{\textbf{dm}} & \multicolumn{4}{|c|}{\textbf{cm}} & \multicolumn{4}{|c|}{\textbf{mm}} & \multirow{2}{2.5em}{$\%$\textbf{IPs}}\\
         \cline{2-17} 
          & $\%$\textbf{Sol} & $\overline{t}$  & $\overline{\epsilon_{P}}$  & $\overline{iter}$ & $\%$\textbf{Sol} & $\overline{t}$ & $\overline{\epsilon_{P}}$ & $\overline{iter}$ & $\%$\textbf{Sol} & $\overline{t}$  & $\overline{\epsilon_{P}}$ & $\overline{iter}$ & $\%$\textbf{Sol} & $\overline{t}$  & $\overline{\epsilon_{P}}$  & $\overline{iter}$ &\\   
         \hline\hline
         \multicolumn{18}{|c|}{\textbf{Numerical Jacobian}}\\  
         \hline
        MP
        &
        100
        &
        7.9
        &
        0.2
        &
        5
        &
        95.9
        &
        20.6
        &
        0.2
        &
        14
        &
        87.9
        &
        48.9
        &
        0.3
        &
        34
        &
        78.1
        &
        46.8
        &
        0.3
        &
        35
        &
        10.2
          \\ 
         \hline
          ED
        &
        100
        &
        7.8
        &
        0.2
        &
        6
        &
        100
        &
        13.7
        &
        0.2
        &
        9
        &
        89.5
        &
        73.8
        &
        0.3
        &
        57
        &
        76.7
        &
        13.1
        &
        0.2
        &
        8
        &
        26.4
          \\
         \hline
          JF
        &
        100
        &
        7.7
        &
        0.2
        &
        5
        &
        95.8
        &
        16.7
        &
        0.2
        &
        13
        &
        87.9
        &
        44.8
        &
        0.3
        &
        34
        &
        78.0
        &
        37.5
        &
        0.3
        &
        34
        &
        10.3
          \\
         \hline
          JD
        &
        100
        &
        7.9
        &
        0.2
        &
        5
        &
        95.9
        &
        17.2
        &
        0.2
        &
        13
        &
        87.1
        &
        39.5
        &
        0.3
        &
        30
        &
        77.6
        &
        50.7
        &
        0.3
        &
        38
        &
        10.5
          \\  
         \hline
          SD
        &
        100
        &
        9.2
        &
        0.3
        &
        9
        &
        100
        &
        49.1
        &
        0.3
        &
        48
        &
        84.1
        &
        29.3
        &
        0.2
        &
        28
        &
        78.6
        &
        24.1
        &
        0.3
        &
        25
        &
        7.7
          \\  
         \hline
          IED
        &
        100
        &
        5.1
        &
        0.2
        &
        6
        &
        100
        &
        8.5
        &
        0.2
        &
        9
        &
        89.7
        &
        51.8
        &
        0.3
        &
        59
        &
        76.7
        &
        7.1
        &
        0.2
        &
        8
        &
        26.2
          \\ 
         \hline 
          SVF
        &
        100
        &
        6.6
        &
        0.2
        &
        5
        &
        96.4
        &
        16.9
        &
        0.2
        &
        13
        &
        87.6
        &
        45.3
        &
        0.3
        &
        38
        &
        77.6
        &
        43.9
        &
        0.3
        &
        34
        &
        10.2
          \\  
         \hline
        \cellcolor{gray9}UC
        &
        \cellcolor{gray9}100
        &
        \cellcolor{gray9}8.4
        &
        \cellcolor{gray9}0.2
        &
        \cellcolor{gray9}6
        &
        \cellcolor{gray9}100
        &
        \cellcolor{gray9}8.3
        &
        \cellcolor{gray9}0.2
        &
        \cellcolor{gray9}6
        &
        \cellcolor{gray9}100
        &
        \cellcolor{gray9}8.6
        &
        \cellcolor{gray9}0.2
        &
        \cellcolor{gray9}6
        &
        \cellcolor{gray9}100
        &
        \cellcolor{gray9}8.6
        &
        \cellcolor{gray9}0.2
        &
        \cellcolor{gray9}6
        &
        \cellcolor{gray9}100
          \\   
         \hline
        \cellcolor{gray9}MX
        &
        \cellcolor{gray9}100
        &
        \cellcolor{gray9}8.6
        &
        \cellcolor{gray9}0.2
        &
        \cellcolor{gray9}6
        &
        \cellcolor{gray9}100
        &
        \cellcolor{gray9}8.4
        &
        \cellcolor{gray9}0.2
        &
        \cellcolor{gray9}6
        &
        \cellcolor{gray9}100
        &
        \cellcolor{gray9}8.7
        &
        \cellcolor{gray9}0.2
        &
        \cellcolor{gray9}6
        &
        \cellcolor{gray9}100
        &
        \cellcolor{gray9}8.8
        &
        \cellcolor{gray9}0.2
        &
        \cellcolor{gray9}6
        &
        \cellcolor{gray9}100
          \\   
         \hline
         \multicolumn{18}{|c|}{\textbf{Geometric Jacobian}}\\  
         \hline
         MP
        &
        100
        &
        4.4
        &
        0.2
        &
        5
        &
        95.6
        &
        11.4
        &
        0.2
        &
        12
        &
        88.5
        &
        28.4
        &
        0.3
        &
        31
        &
        78.3
        &
        38.7
        &
        0.3
        &
        37
        &
        10.5
          \\ 
         \hline
        ED
        &
        100
        &
        7.0
        &
        0.2
        &
        6
        &
        100
        &
        10.6
        &
        0.2
        &
        10
        &
        89.7
        &
        63.6
        &
        0.3
        &
        60
        &
        76.7
        &
        8.3
        &
        0.2
        &
        8
        &
        26.7
          \\
         \hline
       JF
        &
        100
        &
        5.2
        &
        0.2
        &
        5
        &
        95.5
        &
        13.8
        &
        0.2
        &
        12
        &
        88.6
        &
        33.9
        &
        0.3
        &
        32
        &
        78.3
        &
        37.5
        &
        0.3
        &
        37
        &
        10.6
          \\
         \hline
        JD
        &
        100
        &
        5.4
        &
        0.2
        &
        5
        &
        95.5
        &
        13.0
        &
        0.2
        &
        12
        &
        88.2
        &
        27.1
        &
        0.3
        &
        25
        &
        79.0
        &
        29.8
        &
        0.2
        &
        38
        &
        10.7
          \\  
         \hline
        SD
        &
        100
        &
        4.6
        &
        0.2
        &
        8
        &
        100
        &
        27.0
        &
        0.2
        &
        49
        &
        84.2
        &
        15.8
        &
        0.2
        &
        27
        &
        78.2
        &
        21.1
        &
        0.2
        &
        24
        &
        7.6
          \\  
         \hline
        IED
        &
        100
        &
        6.5
        &
        0.2
        &
        6
        &
        100
        &
        10.0
        &
        0.2
        &
        10
        &
        89.6
        &
        58.2
        &
        0.3
        &
        58
        &
        76.8
        &
        11.1
        &
        0.2
        &
        10
        &
        26.4
          \\ 
         \hline 
        SVF
        &
        100
        &
        5.1
        &
        0.2
        &
        5
        &
        95.4
        &
        14.6
        &
        0.2
        &
        12
        &
        88.9
        &
        28.8
        &
        0.3
        &
        29
        &
        77.6
        &
        41.3
        &
        0.2
        &
        39
        &
        10.7
          \\  
         \hline
        \cellcolor{gray9}UC
        &
        \cellcolor{gray9}100
        &
        \cellcolor{gray9}7.1
        &
        \cellcolor{gray9}0.2
        &
        \cellcolor{gray9}6
        &
        \cellcolor{gray9}100
        &
        \cellcolor{gray9}6.7
        &
        \cellcolor{gray9}0.2
        &
        \cellcolor{gray9}6
        &
        \cellcolor{gray9}100
        &
        \cellcolor{gray9}7.3
        &
        \cellcolor{gray9}0.2
        &
        \cellcolor{gray9}6
        &
        \cellcolor{gray9}100
        &
        \cellcolor{gray9}7.7
        &
        \cellcolor{gray9}0.2
        &
        \cellcolor{gray9}6
        &
        \cellcolor{gray9}100
          \\   
         \hline
        \cellcolor{gray9}MX
        &
        \cellcolor{gray9}100
        &
        \cellcolor{gray9}7.3
        &
        \cellcolor{gray9}0.2
        &
        \cellcolor{gray9}6
        &
        \cellcolor{gray9}100
        &
        \cellcolor{gray9}6.9
        &
        \cellcolor{gray9}0.2
        &
        \cellcolor{gray9}6
        &
        \cellcolor{gray9}100
        &
        \cellcolor{gray9}7.7
        &
        \cellcolor{gray9}0.2
        &
        \cellcolor{gray9}6
        &
        \cellcolor{gray9}100
        &
        \cellcolor{gray9}7.9
        &
        \cellcolor{gray9}0.2
        &
        \cellcolor{gray9}6
        &
        \cellcolor{gray9}100
          \\ 
         \hline
         \multicolumn{18}{|c|}{\textbf{Analytical Jacobian}}\\  
         \hline
         MP
        &
        100
        &
        2.4
        &
        0.2
        &
        5
        &
        96.7
        &
        7.0
        &
        0.2
        &
        12
        &
        86.8
        &
        16.4
        &
        0.3
        &
        31
        &
        76.3
        &
        15.7
        &
        0.3
        &
        27
        &
        9.4
          \\ 
         \hline
          ED
        &
        100
        &
        3.3
        &
        0.2
        &
        6
        &
        100
        &
        5.3
        &
        0.2
        &
        9
        &
        89.7
        &
        33.6
        &
        0.3
        &
        59
        &
        76.8
        &
        5.5
        &
        0.2
        &
        9
        &
        27.2
          \\
         \hline
          JF
        &
        100
        &
        3.1
        &
        0.2
        &
        5
        &
        96.7
        &
        7.3
        &
        0.2
        &
        13
        &
        86.9
        &
        18.9
        &
        0.3
        &
        31
        &
        76.3
        &
        15.1
        &
        0.3
        &
        27
        &
        9.3
          \\
        \hline
        JD
        &
        100
        &
        2.5
        &
        0.2
        &
        5
        &
        96.4
        &
        6.8
        &
        0.2
        &
        12
        &
        88.0
        &
        23.8
        &
        0.3
        &
        31
        &
        77.4
        &
        40.5
        &
        0.3
        &
        36
        &
        9.5
          \\  
         \hline
          SD
        &
        100
        &
        8.4
        &
        0.2
        &
        9
        &
        100
        &
        38.0
        &
        0.3
        &
        48
        &
        84.2
        &
        21.3
        &
        0.2
        &
        27
        &
        78.3
        &
        19.3
        &
        0.2
        &
        24
        &
        8.2
          \\  
         \hline
         IED
        &
        100
        &
        5.8
        &
        0.2
        &
        6
        &
        100
        &
        9.8
        &
        0.2
        &
        9
        &
        89.6
        &
        56.1
        &
        0.3
        &
        58
        &
        76.8
        &
        9.6
        &
        0.2
        &
        9
        &
        27.0
          \\ 
         \hline 
        SVF
        &
        100
        &
        5.1
        &
        0.2
        &
        5
        &
        96.3
        &
        12.9
        &
        0.3
        &
        13
        &
        86.6
        &
        29.9
        &
        0.3
        &
        27
        &
        78.8
        &
        43.6
        &
        0.3
        &
        42
        &
        9.4
          \\  
         \hline
         \cellcolor{gray9}UC
        &
        \cellcolor{gray9}100
        &
        \cellcolor{gray9}4.0
        &
        \cellcolor{gray9}0.2
        &
        \cellcolor{gray9}6
        &
        \cellcolor{gray9}100
        &
        \cellcolor{gray9}3.6
        &
        \cellcolor{gray9}0.2
        &
        \cellcolor{gray9}6
        &
        \cellcolor{gray9}100
        &
        \cellcolor{gray9}3.7
        &
        \cellcolor{gray9}0.2
        &
        \cellcolor{gray9}6
        &
        \cellcolor{gray9}100
        &
        \cellcolor{gray9}3.8
        &
        \cellcolor{gray9}0.2
        &
        \cellcolor{gray9}6
        &
        \cellcolor{gray9}100
          \\   
         \hline
         \cellcolor{gray9}MX
        &
        \cellcolor{gray9}100
        &
        \cellcolor{gray9}4.3
        &
        \cellcolor{gray9}0.2
        &
        \cellcolor{gray9}6
        &
        \cellcolor{gray9}100
        &
        \cellcolor{gray9}4.6
        &
        \cellcolor{gray9}0.2
        &
        \cellcolor{gray9}6
        &
        \cellcolor{gray9}100
        &
        \cellcolor{gray9}4.8
        &
        \cellcolor{gray9}0.2
        &
        \cellcolor{gray9}6
        &
        \cellcolor{gray9}100
        &
        \cellcolor{gray9}4.7
        &
        \cellcolor{gray9}0.2
        &
        \cellcolor{gray9}6
        &
        \cellcolor{gray9}100
          \\
          \hline
        %\end{tblr}
        \end{tabular}
        % Note under the table
        %\begin{tablenotes}
        %\centering
        %\small
        %\item\textbf{$\%$Sol}: percentage of solution found, $\overline{t}$: average computation time ($ms$), $\overline{\epsilon_{P}}$: average position error ($mm$), $\overline{iter}$: average number of iterations when the solution is found, $\%$\textbf{IPs}: percentage of identical paths found across the investigated units.
        %\end{tablenotes}
    \end{threeparttable}
\end{table*}

\smallskip
\subsubsection{Other Case Studies} 
Many other motions were experimented for three additional case studies and are included \href{http://vigir.missouri.edu/~dembysj/publications/GI2023/index.html}{in our website}, as for example: 1) the  6DoF serial manipulator with a RRPRRR configuration involving non-singular configurations, and therefore, with a full-rank Jacobian. Here too, an inspection of the two early revolute joints also shows that both $\theta_{1}$ and $\theta_{2}$ affect the prismatic joint $d_{3}$, which in turn could potentially affect $X$, $Y$, and $Z$ coordinates. So, the block partitioning of the MX inverse applied to the Jacobian of this 6DoF robot becomes a combination of all four, non-zero blocks $A_W$, $A_X$, $A_Y$, and $A_Z$. However, since this Jacobian is non-singular in this case, we will have $J^{-1}=J^{-P}=J^{-U}=J^{-M}$ which makes the use of any of the generalized inverses equivalent (and apparently unnecessary) unless the 6DoF robotic arm reaches a singular configuration in its joint space (i.e. the $det(J)=0$). 2) the same 6DoF while starting at a singular configuration. Based on the dynamic rule of thumb, the MX inverse is reduced to the MP inverse at the singular position and is computed with the formula given by equation (\ref{eq7}) at any other position. The results show that the MX inverse is equivalent to the MP inverse in this case. 3) the 7DoF serial manipulator with a RRPRRRR configuration with a non-full rank Jacobian matrix. In this case also, only the MX inverse with a combination of the MP and UC inverses is able to provide unit and rotation consistency when the units are varied from $m$ to $mm$. Due to space issue, complete example motions for these additional cases are provided \href{http://vigir.missouri.edu/~dembysj/publications/GI2023/index.html}{in our website}.

\begin{table*}[!htb] %!htb
\notsotinyone
\caption{\footnotesize Experimental IK results obtained using 1000 random motions of the 7DoF (7R) manipulator generated using the D-H parameters and the forward kinematics. \textcolor{black}{($\alpha$ = 1)}}
\label{tab:results-7DoF-R}
\centering
\begin{threeparttable}
         \begin{tabular}{|c|c|c|c|c|c|c|c|c|c|c|c|c|c|c|c|c|c|c|c|c|c|} 
         %\begin{tblr}{
         %       colspec = {|c|c|c|c|c|c|c|c|c|c|c|c|c|c|c|c|c|c|},
         %       row{4} = {gray9},
         %       row{12} = {gray9},
         %       %row{12} = {gray9},
         %       row{14} = {gray9},
         %       row{22} = {gray9},
                %column{3} = {teal7},
                %cell{2}{3} = {yellow7},
         %     }
         \hline
         \multirow{2}{1em}{\textbf{$J^{\widetilde{-1}}$}} & \multicolumn{5}{c|}{\textbf{m}} & \multicolumn{5}{c|}{\textbf{dm}} & \multicolumn{5}{c|}{\textbf{cm}} & \multicolumn{5}{c|}{\textbf{mm}}  & \multirow{2}{2.5em}{$\%$\textbf{IPs}}\\
         \cline{2-21} 
          & $\%$\textbf{Sol} & $\overline{t}$  & $\overline{\epsilon_{P}}$  & $\overline{\epsilon_{O}}$  & $\overline{iter}$ & $\%$\textbf{Sol} & $\overline{t}$ & $\overline{\epsilon_{P}}$ & $\overline{\epsilon_{O}}$  &  $\overline{iter}$ & $\%$\textbf{Sol} & $\overline{t}$  & $\overline{\epsilon_{P}}$ & $\overline{\epsilon_{O}}$ & $\overline{iter}$ & $\%$\textbf{Sol} & $\overline{t}$  & $\overline{\epsilon_{P}}$  & $\overline{\epsilon_{O}}$ & $\overline{iter}$ &\\   
         \hline\hline
         \multicolumn{22}{|c|}{\textbf{Numerical Jacobian}}\\  
         \hline
        \cellcolor{gray9}MP
        &
        \cellcolor{gray9}99.2
        &
        \cellcolor{gray9}90.3
        &
        \cellcolor{gray9}0.3
        &
        \cellcolor{gray9}0.1
        &
        \cellcolor{gray9}37
        &
        \cellcolor{gray9}99.2
        &
        \cellcolor{gray9}83.2
        &
        \cellcolor{gray9}0.3
        &
        \cellcolor{gray9}0.1
        &
        \cellcolor{gray9}37
        &
        \cellcolor{gray9}99.2
        &
        \cellcolor{gray9}83.3
        &
        \cellcolor{gray9}0.3
        &
        \cellcolor{gray9}0.1
        &
        \cellcolor{gray9}37
        &
        \cellcolor{gray9}99.2
        &
        \cellcolor{gray9}84.1
        &
        \cellcolor{gray9}0.3
        &
        \cellcolor{gray9}0.1
        &
        \cellcolor{gray9}37
        &
        \cellcolor{gray9}100
          \\ 
         \hline
        ED
        &
        78.0
        &
        35.3
        &
        0.3
        &
        0.0
        &
        16
        &
        81.4
        &
        23.7
        &
        0.3
        &
        0.1
        &
        11
        &
        85.4
        &
        28.5
        &
        0.3
        &
        0.2
        &
        13
        &
        85.4
        &
        42.7
        &
        0.3
        &
        0.2
        &
        20
        &
        7.4
          \\
         \hline
        JF
        &
        99.7
        &
        81.5
        &
        0.3
        &
        0.1
        &
        37
        &
        99.5
        &
        81.1
        &
        0.3
        &
        0.1
        &
        37
        &
        99.2
        &
        81.2
        &
        0.3
        &
        0.1
        &
        37
        &
        99.2
        &
        80.7
        &
        0.3
        &
        0.1
        &
        37
        &
        33.9
          \\
         \hline
        JD
        &
        93.8
        &
        68.2
        &
        0.3
        &
        0.1
        &
        33
        &
        99.7
        &
        84.8
        &
        0.3
        &
        0.1
        &
        40
        &
        99.5
        &
        78.6
        &
        0.3
        &
        0.1
        &
        38
        &
        99.4
        &
        74.7
        &
        0.3
        &
        0.1
        &
        36
        &
        14.6
          \\  
         \hline
        SD
        &
        76.7
        &
        126.7
        &
        0.4
        &
        0.0
        &
        63
        &
        79.6
        &
        34.3
        &
        0.4
        &
        0.1
        &
        17
        &
        84.0
        &
        107.2
        &
        0.1
        &
        0.3
        &
        53
        &
        76.4
        &
        725.6
        &
        0.0
        &
        0.6
        &
        363
        &
        7.4
          \\  
         \hline
        IED
        &
        77.4
        &
        29.2
        &
        0.5
        &
        0.0
        &
        14
        &
        81.3
        &
        23.4
        &
        0.3
        &
        0.1
        &
        11
        &
        85.4
        &
        30.8
        &
        0.3
        &
        0.2
        &
        13
        &
        85.6
        &
        42.8
        &
        0.3
        &
        0.2
        &
        19
        &
        7.5
          \\ 
         \hline 
        SVF
        &
        93.6
        &
        79.9
        &
        0.4
        &
        0.1
        &
        35
        &
        99.2
        &
        92.7
        &
        0.3
        &
        0.1
        &
        40
        &
        99.2
        &
        84.5
        &
        0.3
        &
        0.1
        &
        36
        &
        99.5
        &
        91.4
        &
        0.3
        &
        0.1
        &
        39
        &
        7.7
          \\  
         \hline
        UC
        &
        99.1
        &
        126.2
        &
        0.3
        &
        0.1
        &
        54
        &
        99.1
        &
        127.2
        &
        0.3
        &
        0.1
        &
        53
        &
        99.0
        &
        128.2
        &
        0.3
        &
        0.1
        &
        53
        &
        99.1
        &
        128.2
        &
        0.3
        &
        0.1
        &
        53
        &
        99.2
          \\   
         \hline
        \cellcolor{gray9}MX
        &
        \cellcolor{gray9}99.2
        &
        \cellcolor{gray9}90.8
        &
        \cellcolor{gray9}0.3
        &
        \cellcolor{gray9}0.1
        &
        \cellcolor{gray9}37
        &
        \cellcolor{gray9}99.2
        &
        \cellcolor{gray9}84.1
        &
        \cellcolor{gray9}0.3
        &
        \cellcolor{gray9}0.1
        &
        \cellcolor{gray9}37
        &
        \cellcolor{gray9}99.2
        &
        \cellcolor{gray9}84.7
        &
        \cellcolor{gray9}0.3
        &
        \cellcolor{gray9}0.1
        &
        \cellcolor{gray9}37
        &
        \cellcolor{gray9}99.2
        &
        \cellcolor{gray9}84.5
        &
        \cellcolor{gray9}0.3
        &
        \cellcolor{gray9}0.1
        &
        \cellcolor{gray9}37
        &
        \cellcolor{gray9}100
          \\ 
         \hline
         \multicolumn{22}{|c|}{\textbf{Geometric Jacobian}}\\  
         \hline
         
        \cellcolor{gray9}MP
        &
        \cellcolor{gray9}100
        &
        \cellcolor{gray9}38.4
        &
        \cellcolor{gray9}0.3
        &
        \cellcolor{gray9}0.1
        &
        \cellcolor{gray9}13
        &
        \cellcolor{gray9}100
        &
        \cellcolor{gray9}30.4
        &
        \cellcolor{gray9}0.3
        &
        \cellcolor{gray9}0.1
        &
        \cellcolor{gray9}13
        &
        \cellcolor{gray9}100
        &
        \cellcolor{gray9}31.1
        &
        \cellcolor{gray9}0.3
        &
        \cellcolor{gray9}0.1
        &
        \cellcolor{gray9}13
        &
        \cellcolor{gray9}100
        &
        \cellcolor{gray9}30.9
        &
        \cellcolor{gray9}0.3
        &
        \cellcolor{gray9}0.1
        &
        \cellcolor{gray9}13
        &
        \cellcolor{gray9}100
          \\ 
         \hline
          ED
        &
        100
        &
        28.4
        &
        0.3
        &
        0.0
        &
        11
        &
        100
        &
        25.1
        &
        0.3
        &
        0.1
        &
        10
        &
        100
        &
        31.2
        &
        0.3
        &
        0.2
        &
        12
        &
        100
        &
        40.3
        &
        0.3
        &
        0.2
        &
        16
        &
        0.0
          \\
         \hline
          JF
        &
        100
        &
        32.4
        &
        0.3
        &
        0.1
        &
        12
        &
        100
        &
        32.1
        &
        0.3
        &
        0.1
        &
        12
        &
        100
        &
        33.0
        &
        0.3
        &
        0.1
        &
        13
        &
        100
        &
        18.5
        &
        0.3
        &
        0.1
        &
        13
        &
        61.7
          \\
         \hline
        JD
        &
        100
        &
        15.8
        &
        0.3
        &
        0.1
        &
        12
        &
        100
        &
        16.0
        &
        0.3
        &
        0.1
        &
        12
        &
        100
        &
        16.3
        &
        0.3
        &
        0.1
        &
        13
        &
        100
        &
        16.5
        &
        0.3
        &
        0.1
        &
        13
        &
        27.5
          \\  
         \hline
          SD
        &
        99.4
        &
        58.3
        &
        0.4
        &
        0.0
        &
        49
        &
        100
        &
        17.1
        &
        0.4
        &
        0.1
        &
        15
        &
        100
        &
        67.3
        &
        0.1
        &
        0.3
        &
        57
        &
        92.2
        &
        512.6
        &
        0.0
        &
        0.6
        &
        423
        &
        0.0
          \\  
         \hline
        IED
        &
        100
        &
        18.8
        &
        0.5
        &
        0.0
        &
        14
        &
        100
        &
        13.2
        &
        0.3
        &
        0.1
        &
        10
        &
        100
        &
        16.9
        &
        0.3
        &
        0.2
        &
        12
        &
        100
        &
        21.9
        &
        0.3
        &
        0.2
        &
        16
        &
        0.0
          \\ 
         \hline 
        SVF
        &
        100
        &
        16.0
        &
        0.4
        &
        0.0
        &
        11
        &
        100
        &
        17.2
        &
        0.3
        &
        0.1
        &
        12
        &
        100
        &
        17.1
        &
        0.3
        &
        0.1
        &
        13
        &
        100
        &
        17.5
        &
        0.3
        &
        0.1
        &
        13
        &
        16.1
          \\  
         \hline
        UC
        &
        100
        &
        64.3
        &
        0.3
        &
        0.0
        &
        22
        &
        100
        &
        64.6
        &
        0.3
        &
        0.0
        &
        22
        &
        100
        &
        64.4
        &
        0.3
        &
        0.0
        &
        21
        &
        100
        &
        66.3
        &
        0.3
        &
        0.0
        &
        22
        &
        4.9
          \\   
         \hline
        \cellcolor{gray9}MX
        &
        \cellcolor{gray9}100
        &
        \cellcolor{gray9}38.6
        &
        \cellcolor{gray9}0.3
        &
        \cellcolor{gray9}0.1
        &
        \cellcolor{gray9}13
        &
        \cellcolor{gray9}100
        &
        \cellcolor{gray9}30.8
        &
        \cellcolor{gray9}0.3
        &
        \cellcolor{gray9}0.1
        &
        \cellcolor{gray9}13
        &
        \cellcolor{gray9}100
        &
        \cellcolor{gray9}31.5
        &
        \cellcolor{gray9}0.3
        &
        \cellcolor{gray9}0.1
        &
        \cellcolor{gray9}13
        &
        \cellcolor{gray9}100
        &
        \cellcolor{gray9}31.2
        &
        \cellcolor{gray9}0.3
        &
        \cellcolor{gray9}0.1
        &
        \cellcolor{gray9}13
        &
        \cellcolor{gray9}100
          \\ 
         \hline
        %\end{tblr}
        \end{tabular}
        % Note under the table
        %\begin{tablenotes}
        %\centering
        %\small
        %\item\textbf{$\%$Sol}: percentage of solution found, $\overline{t}$: average computation time ($ms$), $\overline{\epsilon_{P}}$: average position error ($mm$), $\overline{iter}$: average number of iterations when the solution is found, $\%$\textbf{IPs}: percentage of identical paths found across the investigated units.
        %\end{tablenotes}
    \end{threeparttable}
    %\vspace{-6mm}
\end{table*}

%\smallskip
\vspace{-4mm}
\subsection{Comparison with Other Inverse Jacobian Methods}

We also compared the MX GI's with commonly used inverse Jacobian methods as global Jacobian-based IK solvers. That is, all the Jacobian types presented in section \ref{sec:jacobians} and inverse Jacobian methods presented in section \ref{sec:generalized-inverses} were implemented for three redundant robotic manipulators: a 3DoF (2RP), a 7DoF (7R), and a 7DoF (2RP4R) with their D-H parameters found in Table \ref{tab:DH-parameters}. For each manipulator, 1000 random initial joint configurations and final poses were generated based on a uniform probability distribution between each joint's limits. Comparisons were performed based on the percentage of solutions found (\textbf{$\%$Sol}), the average computation time ($\overline{t}$) in milliseconds ($ms$), the average position error ($\overline{\epsilon_{P}}$) in millimeter ($mm$), the average orientation error ($\overline{\epsilon_{O}}$) in degrees ($deg$), and the average number of iterations ($\overline{iter}$) for each unit when a solution was found. To evaluate the consistency of the system behaviors when the units are varied; we also compared the percentage of identical paths ($\%$\textbf{IPs}) found across all the investigated units. For the results presented in this section, the maximum number of iterations in the search of an IK solution is set to 500, the accepted position and orientation errors are respectively set to 1$mm$ and 1$degree$.

For the 3DoF (2RP); as it can be seen from the results presented in Table \ref{tab:results-3DoF}, the MX inverse, which reduces to the UC inverse in this case, guarantees all the solutions found within the maximum number of iterations to be unit-consistent across all the investigated units and Jacobian types. However, for all the other inverse Jacobian methods, not only the \textbf{$\%$Sol} is different from one unit to another, but also the $\%$\textbf{IPs} is not $100\%$ which clearly shows that they do not satisfy unit-consistency requirements in this case.

\begin{table*}[!htb] %!htb
\notsotinyone
\caption{\footnotesize Experimental IK results obtained using 1000 random motions of the 7DoF (2RP4R) manipulator generated using the D-H parameters and the forward kinematics. \textcolor{black}{($\alpha$ = 1)}}
\label{tab:results-7DoF-P}
\centering
\begin{threeparttable}
         \begin{tabular}{|c|c|c|c|c|c|c|c|c|c|c|c|c|c|c|c|c|c|c|c|c|c|} 
         %\begin{tblr}{
         %       colspec = {|c|c|c|c|c|c|c|c|c|c|c|c|c|c|c|c|c|c|},
         %       row{4} = {gray9},
         %       row{12} = {gray9},
         %       %row{12} = {gray9},
         %       row{14} = {gray9},
         %       row{22} = {gray9},
                %column{3} = {teal7},
                %cell{2}{3} = {yellow7},
         %     }
         \hline
         \multirow{2}{1em}{\textbf{$J^{\widetilde{-1}}$}} & \multicolumn{5}{c|}{\textbf{m}} & \multicolumn{5}{c|}{\textbf{dm}} & \multicolumn{5}{c|}{\textbf{cm}} & \multicolumn{5}{c|}{\textbf{mm}}  & \multirow{2}{2.5em}{$\%$\textbf{IPs}}\\
         \cline{2-21} 
          & $\%$\textbf{Sol} & $\overline{t}$  & $\overline{\epsilon_{P}}$  & $\overline{\epsilon_{O}}$  & $\overline{iter}$ & $\%$\textbf{Sol} & $\overline{t}$ & $\overline{\epsilon_{P}}$ & $\overline{\epsilon_{O}}$  &  $\overline{iter}$ & $\%$\textbf{Sol} & $\overline{t}$  & $\overline{\epsilon_{P}}$ & $\overline{\epsilon_{O}}$ & $\overline{iter}$ & $\%$\textbf{Sol} & $\overline{t}$  & $\overline{\epsilon_{P}}$  & $\overline{\epsilon_{O}}$ & $\overline{iter}$ &\\   
         \hline\hline
         \multicolumn{22}{|c|}{\textbf{Numerical Jacobian}}\\  
         \hline
        MP
        &
        99.8
        &
        32.8
        &
        0.2
        &
        0.1
        &
        12
        &
        100
        &
        35.3
        &
        0.2
        &
        0.1
        &
        13
        &
        99.9
        &
        154.9
        &
        0.2
        &
        0.1
        &
        58
        &
        83.4
        &
        484.4
        &
        0.1
        &
        0.2
        &
        179
        &
        0.1
          \\ 
         \hline
        ED
        &
        99.5
        &
        23.7
        &
        0.3
        &
        0.0
        &
        11
        &
        99.7
        &
        35.2
        &
        0.2
        &
        0.1
        &
        16
        &
        88.3
        &
        495.8
        &
        0.2
        &
        0.1
        &
        223
        &
        21.6
        &
        309.8
        &
        0.1
        &
        0.2
        &
        138
        &
        0.1
          \\
         \hline
        JF
        &
        99.8
        &
        27.8
        &
        0.2
        &
        0.1
        &
        12
        &
        99.8
        &
        29.8
        &
        0.2
        &
        0.1
        &
        13
        &
        99.5
        &
        123.2
        &
        0.2
        &
        0.1
        &
        55
        &
        83.1
        &
        400.6
        &
        0.1
        &
        0.2
        &
        178
        &
        0.1
          \\
         \hline
        JD
        &
        99.9
        &
        29.7
        &
        0.2
        &
        0.1
        &
        13
        &
        99.9
        &
        28.5
        &
        0.2
        &
        0.1
        &
        13
        &
        99.9
        &
        122.5
        &
        0.2
        &
        0.1
        &
        55
        &
        83.8
        &
        384.2
        &
        0.1
        &
        0.2
        &
        169
        &
        0.1
          \\  
         \hline
        SD
        &
        99.0
        &
        96.0
        &
        0.3
        &
        0.0
        &
        44
        &
        99.9
        &
        43.5
        &
        0.2
        &
        0.1
        &
        20
        &
        99.6
        &
        181.3
        &
        0.1
        &
        0.2
        &
        82
        &
        53.5
        &
        693.4
        &
        0.1
        &
        0.4
        &
        313
        &
        0.0
          \\  
         \hline
        IED
        &
        99.4
        &
        28.2
        &
        0.3
        &
        0.0
        &
        12
        &
        99.7
        &
        37.0
        &
        0.2
        &
        0.1
        &
        16
        &
        88.3
        &
        499.6
        &
        0.2
        &
        0.1
        &
        223
        &
        21.6
        &
        298.9
        &
        0.1
        &
        0.2
        &
        138
        &
        0.1
          \\ 
         \hline 
        SVF
        &
        100
        &
        25.4
        &
        0.3
        &
        0.1
        &
        12
        &
        99.9
        &
        29.5
        &
        0.2
        &
        0.1
        &
        14
        &
        99.8
        &
        114.7
        &
        0.2
        &
        0.1
        &
        55
        &
        81.6
        &
        383.8
        &
        0.1
        &
        0.2
        &
        183
        &
        0.0
          \\  
         \hline
        UC
        &
        99.8
        &
        52.3
        &
        0.2
        &
        0.1
        &
        17
        &
        99.6
        &
        48.3
        &
        0.2
        &
        0.1
        &
        14
        &
        99.8
        &
        39.7
        &
        0.2
        &
        0.1
        &
        15
        &
        99.9
        &
        40.2
        &
        0.2
        &
        0.1
        &
        16
        &
        31.4
          \\   
         \hline
        \cellcolor{gray9}MX
        &
        \cellcolor{gray9}99.9
        &
        \cellcolor{gray9}61.3
        &
        \cellcolor{gray9}0.2
        &
        \cellcolor{gray9}0.1
        &
        \cellcolor{gray9}18
        &
        \cellcolor{gray9}99.9
        &
        \cellcolor{gray9}61.1
        &
        \cellcolor{gray9}0.2
        &
        \cellcolor{gray9}0.1
        &
        \cellcolor{gray9}18
        &
        \cellcolor{gray9}99.9
        &
        \cellcolor{gray9}62.4
        &
        \cellcolor{gray9}0.2
        &
        \cellcolor{gray9}0.1
        &
        \cellcolor{gray9}18
        &
        \cellcolor{gray9}99.9
        &
        \cellcolor{gray9}62.7
        &
        \cellcolor{gray9}0.2
        &
        \cellcolor{gray9}0.1
        &
        \cellcolor{gray9}18
        &
        \cellcolor{gray9}100
          \\
         \hline
         \multicolumn{18}{|c|}{\textbf{Geometric Jacobian}}\\  
         \hline
        MP
        &
        100
        &
        20.5
        &
        0.2
        &
        0.1
        &
        9
        &
        100
        &
        21.8
        &
        0.2
        &
        0.1
        &
        9
        &
        99.8
        &
        104.6
        &
        0.1
        &
        0.1
        &
        46
        &
        81.6
        &
        382.0
        &
        0.1
        &
        0.2
        &
        165
        &
        0.0
          \\ 
         \hline
        ED
        &
        99.8
        &
        23.0
        &
        0.3
        &
        0.0
        &
        10
        &
        99.9
        &
        33.8
        &
        0.2
        &
        0.1
        &
        14
        &
        88.6
        &
        410.3
        &
        0.2
        &
        0.2
        &
        223
        &
        21.6
        &
        233.6
        &
        0.1
        &
        0.2
        &
        139
        &
        0.0
          \\
         \hline
        JF
        &
        100
        &
        14.2
        &
        0.2
        &
        0.1
        &
        9
        &
        100
        &
        15.8
        &
        0.2
        &
        0.1
        &
        9
        &
        100
        &
        77.8
        &
        0.1
        &
        0.1
        &
        47
        &
        81.5
        &
        271.2
        &
        0.1
        &
        0.2
        &
        164
        &
        0.0
          \\
         \hline
        JD
        &
        100
        &
        14.4
        &
        0.2
        &
        0.1
        &
        9
        &
        100
        &
        14.5
        &
        0.2
        &
        0.1
        &
        9
        &
        99.9
        &
        71.1
        &
        0.1
        &
        0.1
        &
        45
        &
        81.2
        &
        290.3
        &
        0.2
        &
        0.1
        &
        184
        &
        0.0
          \\  
         \hline
        SD
        &
        99.8
        &
        43.5
        &
        0.3
        &
        0.0
        &
        30
        &
        99.9
        &
        22.7
        &
        0.2
        &
        0.1
        &
        15
        &
        99.9
        &
        95.7
        &
        0.1
        &
        0.1
        &
        66
        &
        58.0
        &
        489.3
        &
        0.1
        &
        0.4
        &
        345
        &
        0.0
          \\  
         \hline
        IED
        &
        99.8
        &
        16.7
        &
        0.3
        &
        0.0
        &
        10
        &
        99.9
        &
        23.6
        &
        0.2
        &
        0.1
        &
        14
        &
        88.6
        &
        359.5
        &
        0.2
        &
        0.2
        &
        223
        &
        21.6
        &
        217.7
        &
        0.1
        &
        0.2
        &
        139
        &
        0.1
          \\ 
         \hline 
        SVF
        &
        99.9
        &
        13.6
        &
        0.3
        &
        0.1
        &
        8
        &
        100
        &
        14.5
        &
        0.2
        &
        0.1
        &
        9
        &
        100
        &
        69.4
        &
        0.1
        &
        0.1
        &
        45
        &
        78.3
        &
        275.1
        &
        0.1
        &
        0.1
        &
        173
        &
        0.0
          \\  
         \hline
        UC
        &
        99.4
        &
        26.7
        &
        0.2
        &
        0.1
        &
        9
        &
        99.5
        &
        26.9
        &
        0.2
        &
        0.1
        &
        9
        &
        99.3
        &
        27.0
        &
        0.2
        &
        0.1
        &
        9
        &
        99.5
        &
        26.6
        &
        0.2
        &
        0.1
        &
        9
        &
        47.5
          \\   
         \hline
        \cellcolor{gray9}MX
        &
        \cellcolor{gray9}100
        &
        \cellcolor{gray9}35.6
        &
        \cellcolor{gray9}0.2
        &
        \cellcolor{gray9}0.1
        &
        \cellcolor{gray9}13
        &
        \cellcolor{gray9}100
        &
        \cellcolor{gray9}35.4
        &
        \cellcolor{gray9}0.2
        &
        \cellcolor{gray9}0.1
        &
        \cellcolor{gray9}13
        &
        \cellcolor{gray9}100
        &
        \cellcolor{gray9}36.2
        &
        \cellcolor{gray9}0.2
        &
        \cellcolor{gray9}0.1
        &
        \cellcolor{gray9}13
        &
        \cellcolor{gray9}100
        &
        \cellcolor{gray9}35.8
        &
        \cellcolor{gray9}0.2
        &
        \cellcolor{gray9}0.1
        &
        \cellcolor{gray9}13
        &
        \cellcolor{gray9}100
          \\ 
         \hline
        %\end{tblr}
        \end{tabular}
        % Note under the table
        %\begin{tablenotes}
        %\centering
        %\small
        %\item\textbf{$\%$Sol}: percentage of solution found, $\overline{t}$: average computation time ($ms$), $\overline{\epsilon_{P}}$: average position error ($mm$), $\overline{iter}$: average number of iterations when the solution is found, $\%$\textbf{IPs}: percentage of identical paths found across the investigated units.
        %\end{tablenotes}
    \end{threeparttable}
    \vspace{-4mm}
\end{table*}

For the 7DoF (7R), the results can be seen in Table \ref{tab:results-7DoF-R}. Applying the proposed rule of thumb to this commensurate 7DoF manipulator where all the joints are revolute, reduces the MX inverse to the MP inverse. In this case, also, the MX inverse (reduced to the MP inverse) guarantees all the solutions found within the maximum number of iterations to be unit-consistent across all the investigated units with the $\%$\textbf{IPs} being $100\%$. Here, we also observed that all the other inverse Jacobian methods did not guarantee all solutions to be unit-consistent as the $\%$\textbf{IPs} was not $100\%$ even when the \textbf{$\%$Sol} was $100\%$ for each unit of a specific inverse method.

For the 7DoF (2RP4R), the use of the MX inverse requires a combination of the MP and UC inverses. As it can be seen in Table \ref{tab:results-7DoF-P}, only the MX inverse guarantees all the solutions found within the maximum number of iterations to be unit-consistent across all the investigated units with the $\%$\textbf{IPs} being $100\%$ while the other inverse methods do not.

\vspace{-2mm}
\section{Conclusion}
This research investigated the most commonly used Jacobian types and Generalized Inverses to compute the inverse Jacobian matrix for numerical IK solvers. It showed that  the most relevant Jacobian-based IK approaches fail to presercve the same system behavior when the units are varied for incommensurate robotic manipulators.  While the Moore-Penrose (MP) inverse is applicable to systems that require rotation consistency, it may fail in the presence of systems that require unit consistency where the Unit-Consistency (UC) inverse is applicable instead. With the goal of always applying the MX inverse that combines the MP and UC inverses to achieve unit and rotation invariant robotic systems, a new dynamic rule of thumb based on the Denhavit-Hartenberg (D-H) methodology was introduced to determine which specific variables to place in each block of Mixed (MX) inverse. This new rule was validated by conducting multiple experiments with six different manipulators and various configuration cases. Furthermore, we investigated the effects of the attenuation parameter (gain) $\alpha$ used in the numerical IK solvers, both on the accuracy of the final IK solution and the smoothness between the initial and final poses of the end-effector. 
For future work, we would like to further investigate the integration of the proposed approach to reliably provide unit-consistency in cases of kinematic singularities and multiple solutions for IK solvers. In addition, the proposed approach will be extended to Pseudoinverse-based Path Planning (PPP) and Repetitive Motion Planning (RMP) where commonly used GI's may also alter the system behavior when units are varied for incommensurate robotic manipulators. It is a common practice to reformulate PPP and PRMP schemes as Quadratic Programming (QP) problems, another research direction will investigate if these methods also suffer from the same problem. Finally, GI's are ubiquitous to many robotics and machine learning problems, another potential research direction will conduct a vast investigation to survey, identify, and address the issues related to the use of GI's in various subjects of these areas.

\ifCLASSOPTIONcaptionsoff
  \newpage
\fi

% trigger a \newpage just before the given reference
% number - used to balance the columns on the last page
% adjust value as needed - may need to be readjusted if
% the document is modified later
%\IEEEtriggeratref{8}
% The "triggered" command can be changed if desired:
%\IEEEtriggercmd{\enlargethispage{-5in}}

% references section

% can use a bibliography generated by BibTeX as a .bbl file
% BibTeX documentation can be easily obtained at:
% http://mirror.ctan.org/biblio/bibtex/contrib/doc/
% The IEEEtran BibTeX style support page is at:
% http://www.michaelshell.org/tex/ieeetran/bibtex/
%\bibliographystyle{IEEEtran}
% argument is your BibTeX string definitions and bibliography database(s)
%\bibliography{IEEEabrv,../bib/paper}
%
% <OR> manually copy in the resultant .bbl file
% set second argument of \begin to the number of references
% (used to reserve space for the reference number labels box)
%\begin{thebibliography}{1}

%\bibitem{IEEEhowto:kopka}
%H.~Kopka and P.~W. Daly, \emph{A Guide to \LaTeX}, 3rd~ed.\hskip %1em plus
%  0.5em minus 0.4em\relax Harlow, England: Addison-Wesley, 1999.

%\end{thebibliography}

%\printbibliography[]
\vspace{-2mm}
\bibliographystyle{IEEEtran}
\bibliography{IEEEabrv, main.bib}

\end{document}